\documentclass{article}

\usepackage{arxiv}

\usepackage[utf8]{inputenc} 
\usepackage[T1]{fontenc}    
\usepackage{hyperref}       
\usepackage{url}            
\usepackage{booktabs}       
\usepackage{amsmath,amssymb,amsfonts}       
\usepackage{nicefrac}       
\usepackage{microtype}      
\usepackage{lipsum}
\usepackage{graphicx}

\usepackage{algorithm}
\usepackage{algpseudocode}
\usepackage{textcomp}
\usepackage[dvipsnames]{xcolor}

\usepackage{booktabs}
\usepackage{tabularx}
\usepackage{array}
\usepackage{multirow}
\usepackage{arydshln} 
\usepackage{adjustbox} 
\usepackage{color, colortbl}
\usepackage{hhline}
\usepackage{multicol}

\usepackage{subcaption}
\usepackage{makecell}

\usepackage{arydshln}

\usepackage{stmaryrd}
\usepackage{makecell}
\usepackage{verbatim}

\usepackage[sorting=none]{biblatex}
\DeclareBibliographyDriver{article}{%
  \usebibmacro{bibindex}%
  \usebibmacro{begentry}%
  \printnames{author}%
  \newunit
  \printfield{title}%
  \newunit
  \printfield{journaltitle}%
  \setunit*{\addspace}%
  \printfield{volume}%
  \setunit*{\addcomma\space}%
  \printfield{number}%
  \setunit*{\addcomma\space}%
  \printfield{pages}%
  \setunit*{\addcomma\space}%
  \printfield{year}%
  \usebibmacro{finentry}%
}

\DeclareBibliographyDriver{inproceedings}{%
  \usebibmacro{bibindex}%
  \usebibmacro{begentry}%
  \printnames{author}%
  \newunit
  \printfield{year}
  \newunit
  \printfield{title}%
  \newunit
  \printtext{in \printfield{booktitle}}%
  \newunit
  \printfield{pages}
  \usebibmacro{finentry}%
}

\DeclareBibliographyDriver{misc}{%
  \usebibmacro{bibindex}%
  \usebibmacro{begentry}%
  \printnames{author}%
  \newunit
  \printfield{year}
  \newunit
  \printfield{title}%
  \newunit
  \printfield{number}
  \newunit\newblock
  \usebibmacro{finentry}%
}

\newcommand{\mr}[1]{\mathit{#1}}
\newcommand{\loss}{\mathcal{L}}
\newcommand{\xb}{\mathbf{x}}
\newcommand{\Zb}{\mathbf{Z}}
\newcommand{\Pb}{\mathbf{P}}
\newcommand{\pb}{\mathbf{p}}
\newcommand{\mb}{\mathbf{m}}
\newcommand{\bb}{\mathbf{b}}
\newcommand{\yb}{\mathbf{y}}
\newcommand{\ppm}{\,\tiny$\pm$}
\newcommand{\ccol}{\cellcolor{gray!15}}
\newcommand{\Real}{\mathbb{R}}
\newcommand{\RR}{\mathbb{R}}

\newcommand{\revision}[1]{{\color{black}#1}}
\newcommand{\revisiontwo}[1]{{\color{black}#1}}
\newcommand{\gray}[1]{{\color{gray}#1}}
\newcommand{\graytwo}[1]{{\color{gray}#1}}

\title{Prompt learning with bounding box constraints for medical image segmentation}

\author{Mélanie Gaillochet\thanks{École de Technologie Supérieure, Montréal, QC H3C 1K3, Canada.} \thanks{Mila - Quebec AI Institute, Montréal, QC H2S 3H1, Canada.} \thanks{ Polytechnique Montréal, QC H3T 1J4, Canada.} \\ \texttt{ melanie.gaillochet.1@ens.etsmtl.ca}
\And
Mehrdad Noori\footnotemark[1]\\
\texttt{mehrdad.noori.1@ens.etsmtl.ca}
\And
Sahar Dastani\footnotemark[1]\\
\texttt{sahar.dastani-oghani.1@ens.etsmtl.ca}
\And
Christian Desrosiers\footnotemark[1]\\
\texttt{christian.desrosiers@etsmtl.ca}
\And
Hervé Lombaert\footnotemark[1] \footnotemark[2] \footnotemark[3] \\ 
\texttt{herve.lombaert@polymtl.ca}
}

\begin{document}
\maketitle
\begin{abstract}
Pixel-wise annotations are notoriously \revision{labourious and costly} to obtain in the medical domain. \revision{To mitigate this burden,} weakly supervised approaches based on bounding box annotations\revision{—much easier to acquire—offer a practical alternative}. \revision{Vision foundation models have recently shown noteworthy segmentation} performance \revision{when provided with} prompts such as points or bounding boxes. Prompt learning \revision{exploits these models} by adapting them to downstream tasks and automating \revision{segmentation}, \revision{thereby reducing user intervention}. However, existing prompt learning approaches \revision{depend on fully annotated segmentation masks}.
This paper proposes a \revision{novel} framework that \revision{combines the representational power of foundation models with the annotation efficiency of weakly supervised segmentation}. \revision{More specifically, our approach} automates prompt generation for foundation models using only \revision{bounding box annotations}.
\revision{Our proposed optimization scheme integrates} multiple constraints derived from box annotations \revision{with} pseudo-labels generated by the prompted foundation model. \revision{Extensive experiments} across multi-modal datasets \revision{reveal that} our weakly supervised method \revision{achieves} an average Dice score of \revisiontwo{$84.90\%$} \revision{in a limited data setting, outperforming existing fully-supervised and weakly-supervised approaches}. 
The code is available at \href{https://github.com/Minimel/box-prompt-learning-VFM.git}{https://github.com/Minimel/box-prompt-learning-VFM.git}.
\end{abstract}


\section{Introduction}
\label{sec:introduction}

\begin{figure}[ht]
\centering
\setlength{\tabcolsep}{1pt}
\begin{tabular}{c}
    \includegraphics[width=0.5\linewidth]{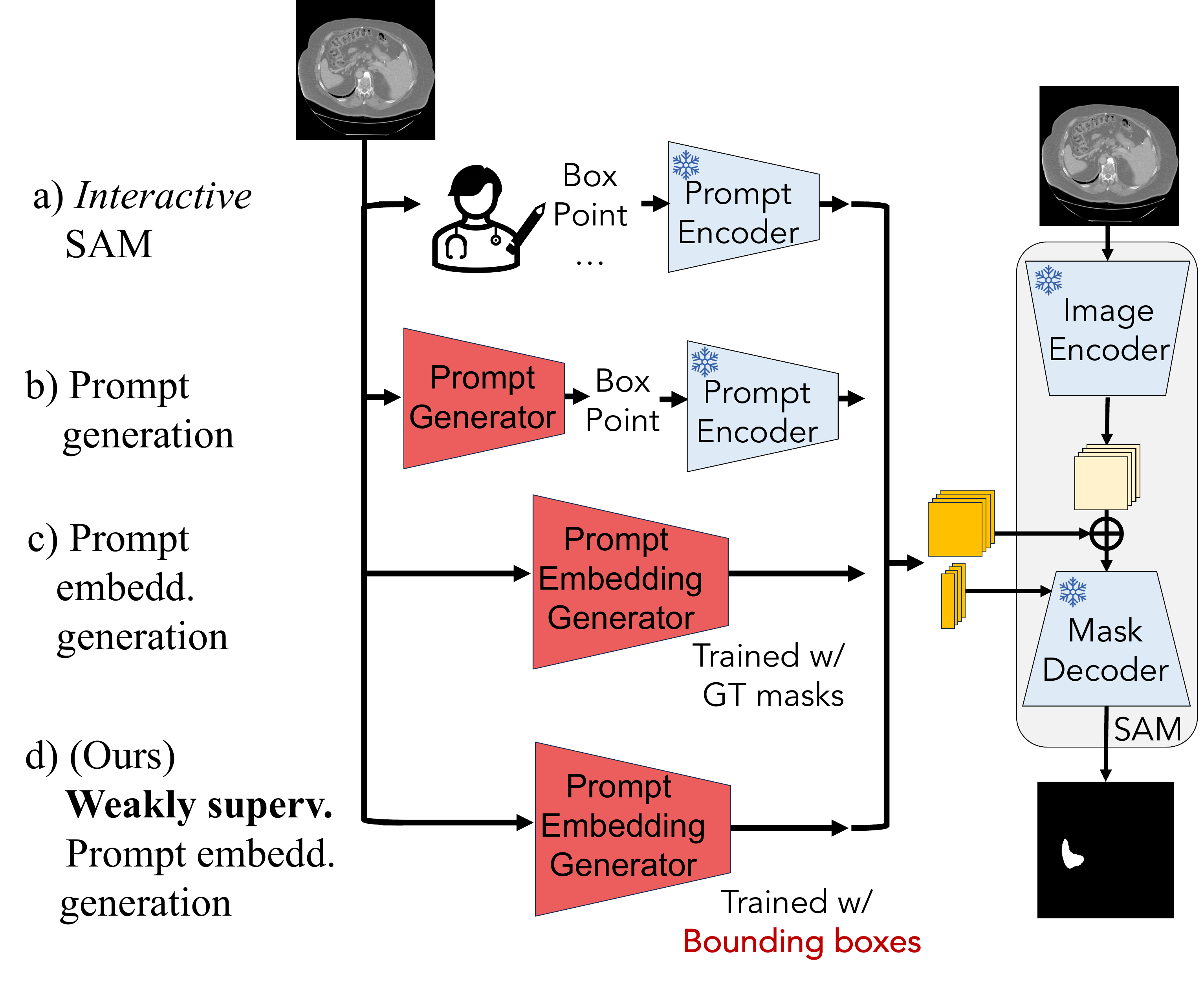} 
\end{tabular}
\caption{Differences between a) Interactive SAM and b-d) Various adaptive methods for prompt automation: 
b) Using SAM’s prompt encoder with physical prompts generated by a trained prompt module, 
c) Directly generating dense and/or sparse prompt embeddings (\textit{dark yellow}) with a prompt module trained with ground truth (GT) masks, and 
d) Our method, which generates prompt embeddings from the image using only bounding box annotations during training.}
\label{fig:SAM_AdaptedMethods}
\end{figure}

The precise delineation of regions of interest is a crucial step in clinical decision-making, affecting \revision{the} diagnosis, treatment planning and patient outcome. However, manual annotation of medical images \revision{remains} a labour-intensive and time-consuming process. \revision{With an ever-growing volume of medical imaging data, developing efficient and accurate segmentation methods} has become a major challenge. Automatic segmentation algorithms \cite{litjens_survey_2017} have been \revision{introduced} to alleviate the burden of manual annotation and reduce the \revision{impact of expert subjectivity and inadvertent errors}. In medical image analysis, segmentation methods based on variants of the UNet \cite{ronneberger_u-net_2015} and lately of the vision transformer architecture \cite{vaswani_attention_2017,dosovitskiy_image_2021}, have \revision{achieved} state-of-the-art performance on \revision{multiple} tasks and datasets \cite{isensee_nnu-net_2021,hatamizadeh_unetr_2022,chen_transunet_2024}. However, \revision{these models depend on large, fully-annotated datasets} for training, and their performance typically deteriorates when training data is scarce. \revision{Yet}, the prohibitive cost of manually labeling medical images makes the collection of such well-annotated datasets difficult, hampering the development of these data-hungry models. 

Recently, large promptable vision models \cite{kirillov_segment_2023,zou_segment_2023,ma_segment_2024} have gained considerable attention for their flexibility and generalization abilities on multiple natural imaging benchmarks. \revision{Unlike specialized models trained on domain-specific datasets, these foundation models can generate, for any new image, informative representations which can be leveraged to produce a segmentation mask}. Notably, the Segment Anything Model (SAM) \cite{kirillov_segment_2023} has showcased impressive zero-shot capabilities by generating \revision{accurate} segmentation masks for tasks and classes unseen during training. 

\begin{figure}[!t]
\centering
\setlength{\tabcolsep}{1pt}
\begin{tabular}{cccc}
    \includegraphics[width=.18\linewidth]{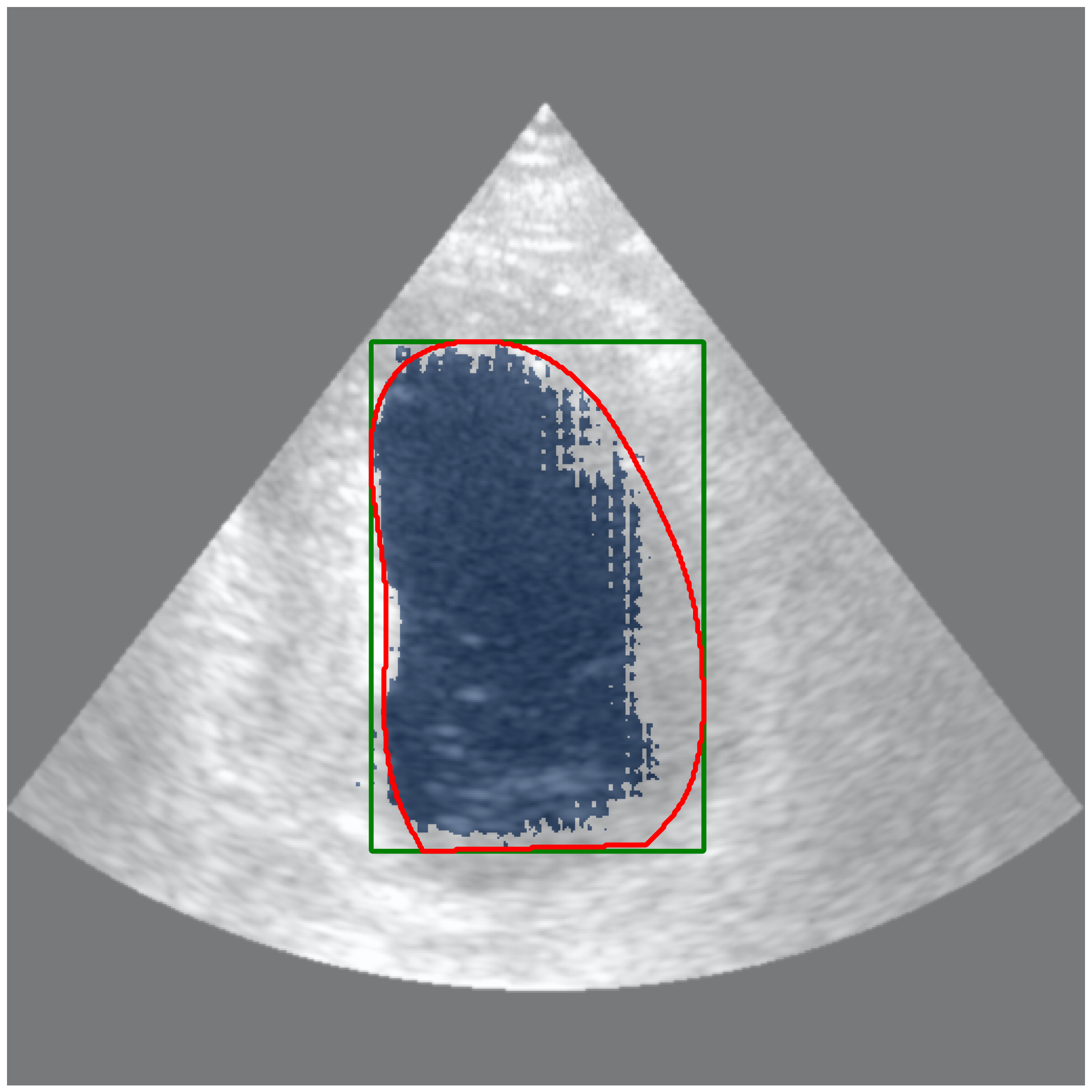} &
    \includegraphics[width=.18\linewidth]{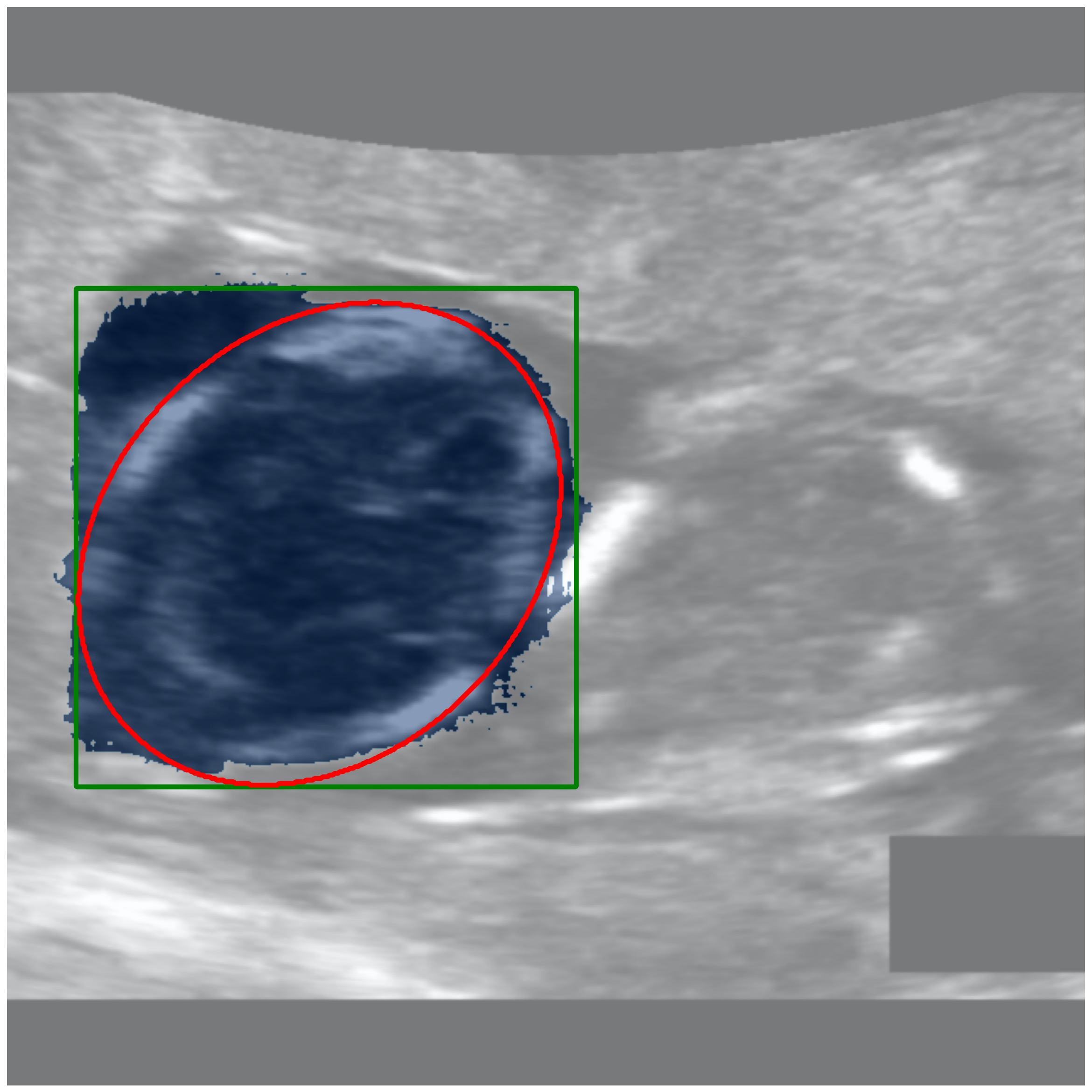} &
    \includegraphics[width=.18\linewidth]{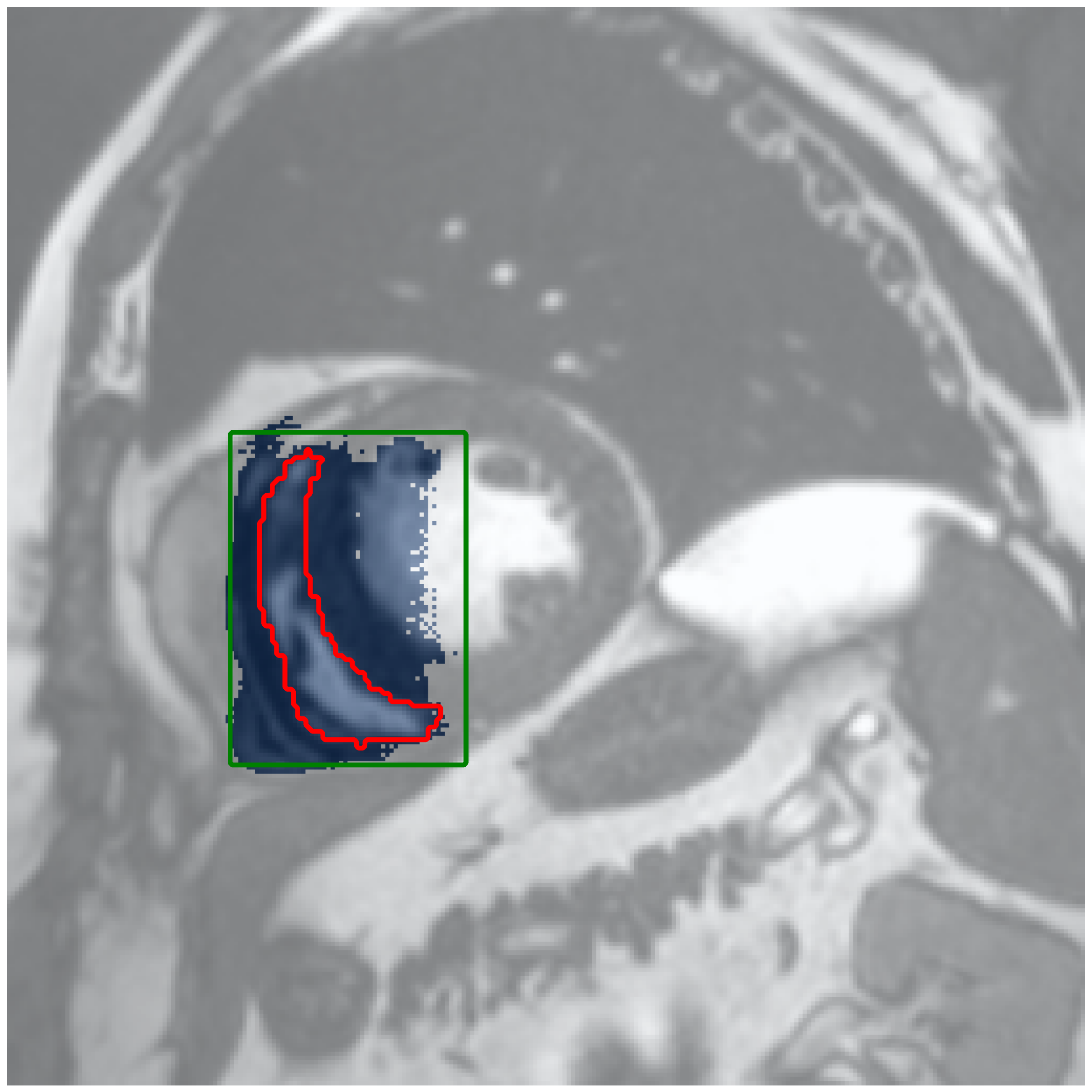} &
    \includegraphics[width=.18\linewidth]{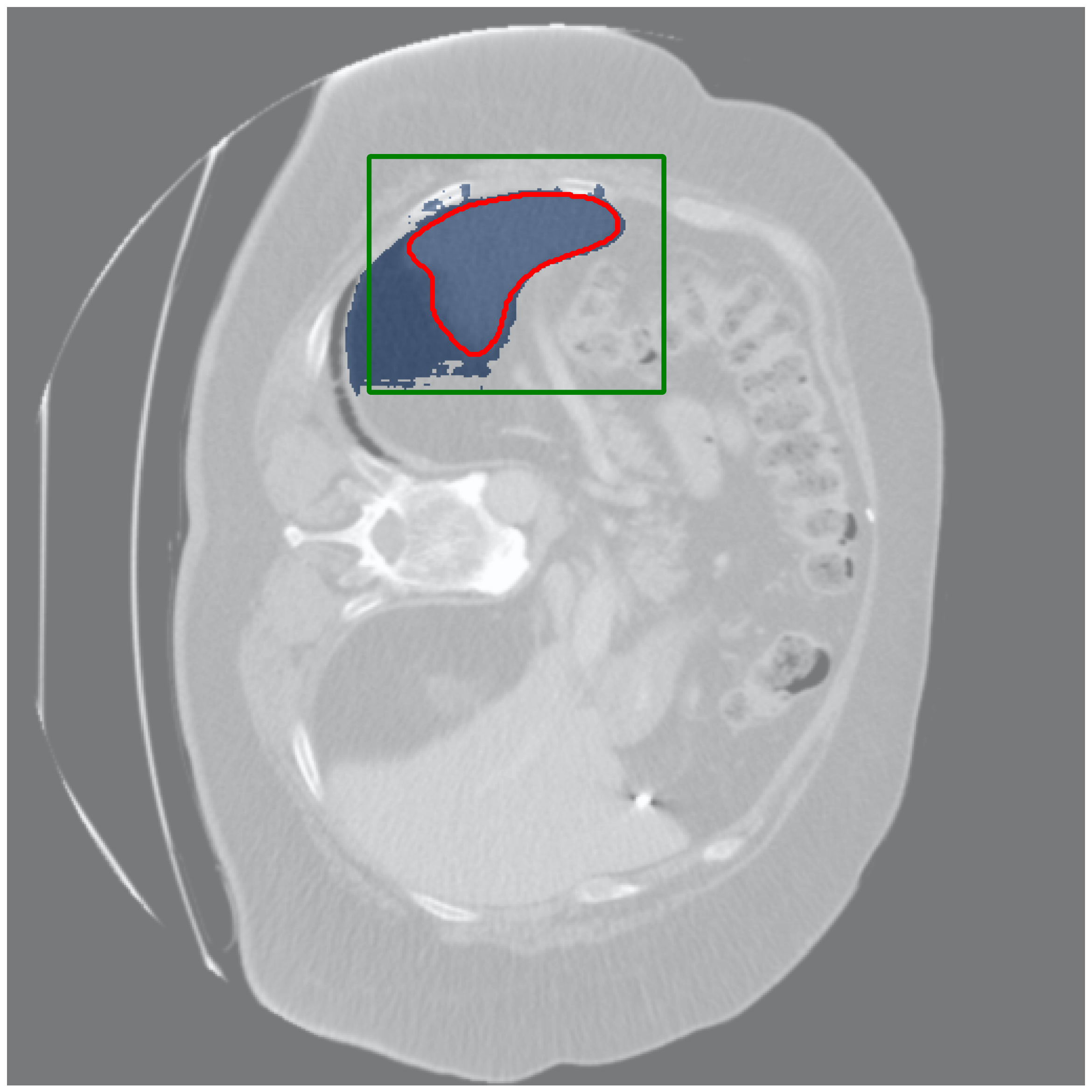} \\[8pt]
\end{tabular}
\caption{Examples of SAM predictions with varying noise levels in the box prompts. From left to right: no noise (tight box), 0-1.5\%, 1.5-3\%, and 3-5\% pixel displacement from the original tight box. Ground-truth annotations and associated box prompts are shown in red and green, respectively, with predicted segmentation masks overlaid in blue. Prompt ambiguity and noise, combined with applications to out-of-domain data, can cause vision foundation models to fail in effectively segmenting the target object.}
\label{fig:PromptAmbiguity}
\end{figure}

Current foundation models \revision{require user-provided prompts, typically manually crafted, to tailor the segmentation outputs to specific tasks}. The zero-shot performance of these models heavily depends on the quality of the user prompt \cite{mazurowski_segment_2023,kim_customizing_2024,yue_surgicalsam_2024,ma_segment_2024}. However, prompts can be ambiguous, particularly in medical images where object boundaries are often \revision{subtle}, leading to poor predictions (see Fig.\ref{fig:PromptAmbiguity}). 

\revision{To fully exploit the potential of foundation models and improve scalability, recent efforts have focused on automating prompt generation. In particular, prompt tuning \cite{shaharabany_autosam_2023,wu_self-prompting_2023,chen_rsprompter_2024,zhang_personalize_2024,kim_customizing_2024} adapts large models for downstream tasks by optimizing a small set of trainable prompt embeddings, rather than updating millions of model parameters}. Prompt tuning searches for the best input prompt to satisfy the target task. \revision{Visual prompt tuning \cite{avidan_visual_2022,wang_review_2023} has demonstrated remarkable adaptation performance with minimal trainable parameters, and several studies have applied this approach to SAM \cite{shaharabany_autosam_2023,wu_self-prompting_2023,zhang_personalize_2024,ayzenberg_protosam_2024,yue_surgicalsam_2024,li_autoprosam_2024}}. Existing methods for medical image segmentation have focused on automatically generating a physical prompt—i.e., point or box \cite{wu_self-prompting_2023,ayzenberg_protosam_2024} or a prompt embedding \cite{shaharabany_autosam_2023,yue_surgicalsam_2024,li_autoprosam_2024} (see Fig.\ref{fig:SAM_AdaptedMethods}). However, these techniques still \revision{heavily rely on fully annotated segmentation masks}, which are burdensome to obtain in the medical domain.

\revision{In parallel to advanced prompt tuning techniques, weakly supervised learning offers a pragmatic approach to reduce the reliance on exhaustive manual annotations by incorporating more accessible forms of labeling.} Weak labels can come in various shapes, such as scribbles \cite{lin_scribblesup_2016}, image tags \cite{pathak_constrained_2015}, points \cite{bearman_whats_2016} or bounding boxes \cite{dai_boxsup_2015,khoreva_simple_2017,hsu_weakly_2019,kervadec_bounding_2020}. Among \revision{these}, bounding boxes \revision{are particularly appealing} due to their simplicity and light storage—in practice, only two corner coordinates are needed to define a bounding box. Bounding boxes have been used as pseudo-labels \revision{to generate} initial segmentation proposals \cite{rother_grabcut_2004,rajchl_deepcut_2017}, which are \revision{refined iteratively} until a more precise segmentation is obtained. However, such approaches are subject to error propagation \revision{if the initial segmentation is inaccurate, ultimately impairing model performance. To mitigate this issue}, alternative methods have introduced attention mechanisms to \revision{improve gradient flow \cite{song_box-driven_2019,kulharia_box2seg_2020} or imposed constraints on the output probabilities during optimization} \cite{jia_constrained_2017,kervadec_bounding_2020}. Building on \revision{these} advancements, our work \revision{integrates constraint-based optimization with bounding box annotations into the framework of visual prompt tuning.} 

\revision{Our work introduces a novel framework that leverages the complementary} strengths of foundation models \revision{and weakly supervised learning through prompt tuning}. \revision{We} train an auxiliary prompt module \revision{using} only bounding box annotations—\revision{much} easier to obtain than ground truth masks—to automatically generate informative prompt \revision{embeddings from input images}. Relying on weak labels significantly reduces \revision{the annotation burden, making the adaptation} of foundation models more scalable for medical imaging applications. 
\revision{Building on our preliminary work \cite{gaillochet_automating_2024}, we address prior limitations by moving beyond} using weak-label constraints that only \revision{exploited} bounding box tightness. \revision{Our updated training strategy now employs a box-based multi-loss optimization framework that integrates predictions from the prompted foundation model with consistency-based regularization, leading to more robust performance.}
Furthermore, \revision{while our earlier work focused exclusively on MedSAM, we now demonstrate that SAM—despite being trained on out-of-domain data—can serve as a general backbone foundation model when coupled with a} deeper prompt module and our \revision{refined} optimization strategy.

\section{Related Work}
\subsection{Vision Foundation Models for medical image segmentation}
Recent years have witnessed the emergence of foundation models with impressive zero-shot capabilities like the Segment Anything Model \cite{kirillov_segment_2023}, trained on a vast dataset of 1B masks and 11M images.
Based on vision transformers \cite{dosovitskiy_image_2021}, the promptable SAM has demonstrated notable success across various computer vision tasks due to its extensive pre-training. 

Because of its success on natural images, SAM's potential for medical image segmentation has been a growing field of study \cite{zhang_segment_2023,mazurowski_segment_2023,huang_segment_2024,}. Efforts have been undertaken to \revision{adapt} SAM for medical imaging and other specific segmentation tasks 
\cite{shaharabany_autosam_2023,gu_how_2024}. \cite{cheng_sam-med2d_2023,ma_segment_2024,gu_how_2024} explored fine-tuning strategies, whereas \cite{wu_self-prompting_2023,shaharabany_autosam_2023,wu_MedicalSAMAdapter_2025} investigated externally designed components. However, fine-tuning approaches incur practical challenges in terms of data collection, data annotation and computational time.
Moreover, they remain interactive \revision{models}, requiring user input during the inference stage. Hence, alternative approaches have explored the idea of prompt tuning for SAM in the context of medical imaging.

\subsection{Prompt Learning for Vision Foundation Models}
Prompt-tuning has recently been applied to large vision models \cite{avidan_visual_2022} to adapt them to medical image segmentation or to automate prompt generation. In the medical domain, one approach has been to automatically generate point and box prompts from the image by incorporating a detection network into SAM's architecture \cite{wu_self-prompting_2023,ayzenberg_protosam_2024}. These prompts are then converted to embeddings via SAM's prompt encoder (see Fig.\ref{fig:SAM_AdaptedMethods}.b). An alternative has been to use a self-prompting module to yield prompt embeddings directly, hence replacing the original prompt encoder \cite{shaharabany_autosam_2023,yue_surgicalsam_2024,li_autoprosam_2024} (see Fig.\ref{fig:SAM_AdaptedMethods}.c).  
PerSAM \cite{zhang_personalize_2024} and ProtoSAM \cite{ayzenberg_protosam_2024} compare the embeddings of the image, from SAM or an external network, with those of a single image-mask pair and generated appropriate point and box prompts.
Unlike our approach, these prompt learning methods require samples with ground-truth annotation masks, which remains a burden for medical datasets. Recently, \cite{gaillochet_automating_2024} showed that weak labels could be exploited to automate prompt learning but used a restrictive tight box constraint, limiting the approach to in-domain performance. 

\subsection{Weakly Supervised Learning with Bounding Boxes}
Weakly supervised segmentation methods have been developed to reduce the cost of manual annotation. Popular among such methods are approaches based on bounding box annotations. These bounding boxes are typically used as initial pseudo-labels for identifying target regions. For instance, the classic GrabCut algorithm \cite{rother_grabcut_2004}, based on graph cuts \cite{boykov2001fast}, iteratively separates foregrounds from backgrounds using bounding boxes. DeepCut \cite{rajchl_deepcut_2017} extends GrabCut to neural networks.  
However, a well-known challenge with these iterative methods involves errors appearing in the initial segmentation and being propagated during training. Attention maps were proposed during training to reduce the propagation of incorrect gradients by masking out irrelevant regions \cite{song_box-driven_2019} or by handling the label noise assumed to be contained in the bounding boxes \cite{kulharia_box2seg_2020}.
More recently, constraints on tightness and size were applied to guide the segmentation during training, either alone \cite{kervadec_bounding_2020} or in combination with multiple instance learning \revision{(MIL) and smooth maximum approximation} \cite{hsu_weakly_2019,wang_bounding_2021}.

\section{Contributions}
\revision{This paper introduces a novel framework that leverages the strengths of foundation models and the cost-efficiency of weakly supervised segmentation. }
\revision{More specifically, we \textit{automate and adapt foundation models} by training with only \textit{bounding box annotations} an auxiliary \textit{prompt module} that automatically \textit{generates a relevant prompt embedding} from the input image}.
Our novel training approach for weakly labeled data exploits multiple pieces of information provided by bounding box annotations. Training our auxiliary prompt module optimizes a loss based on the prediction of the foundation model when prompted by a bounding box, as well as box-based spatial constraints and a consistency-based regularization. The constraints and regularization refine the segmentation of the prompted foundation model, which may contain inaccuracies, and allow for the application of the foundation model to out-of-domain data. 

At its core, our method fundamentally \revision{replaces} the traditional prompt encoder \revision{of the foundation model} with a new auxiliary prompt embedding generator, which:
\begin{enumerate}\setlength\itemsep{.25em}
    \item Requires only \textbf{bounding boxes} to train by employing a novel box-based multi-loss optimization strategy,
    \item Generalizes effectively to full and \textbf{limited training data}, 
    \item Performs well \textbf{across modalities}, as shown by our experiments on MRI, CT and ultrasound images, and
    \item Successfully functions on \textbf{out-of-domain data}, as shown by our experiments with SAM as a backbone foundation model.
\end{enumerate}

\begin{figure*}[htb!]
\centering
    \begin{subfigure}[b]{0.7\textwidth} 
        \centering
        \includegraphics[width=\textwidth]{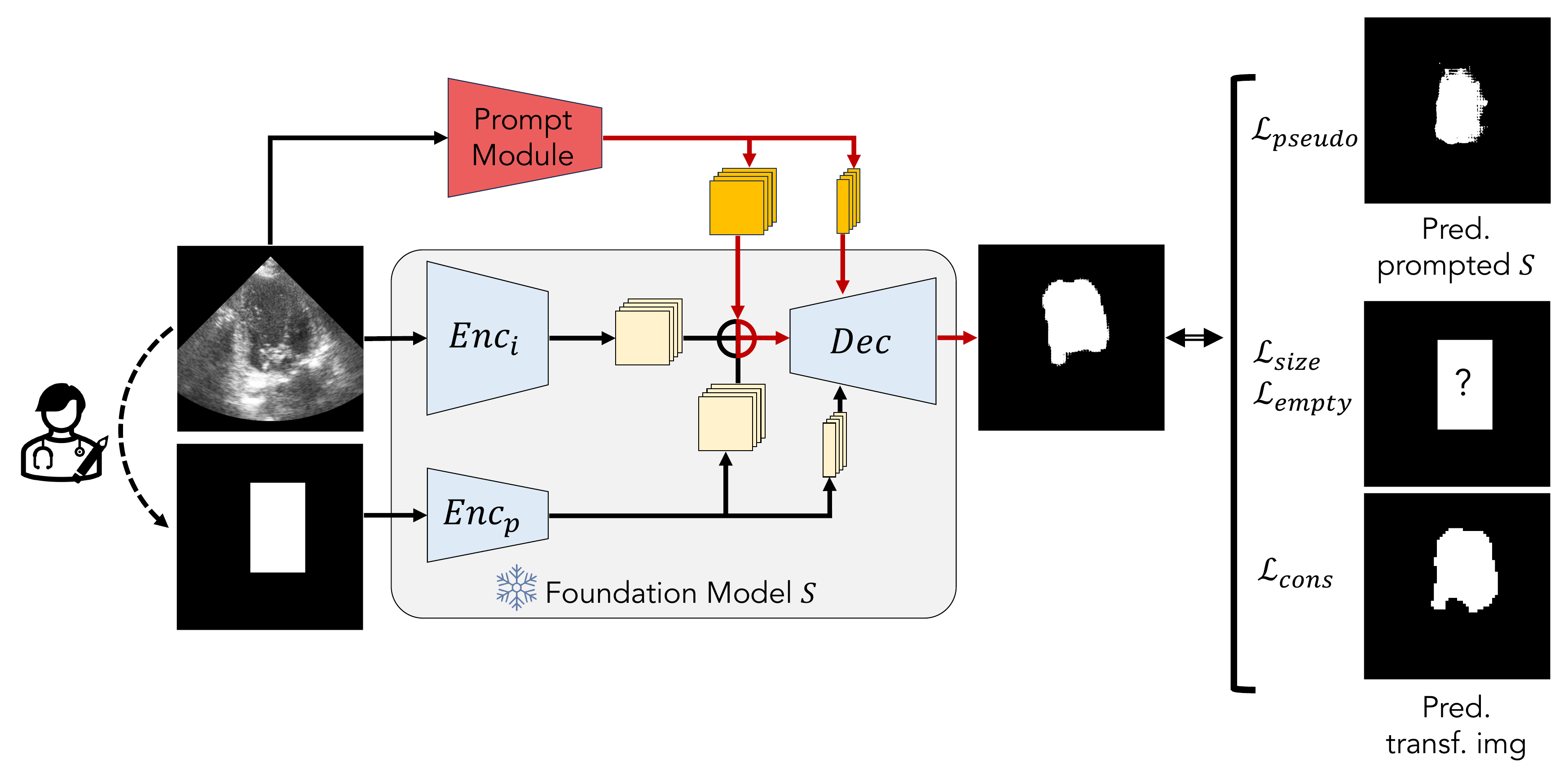}
        \caption{Training of our add-on prompt embedding generator module with bounding box annotations}
        \label{subfig:framework_train}
    \end{subfigure}
    \hfill 
    \begin{subfigure}[b]{0.7\textwidth} 
        \centering
        \includegraphics[width=\textwidth]{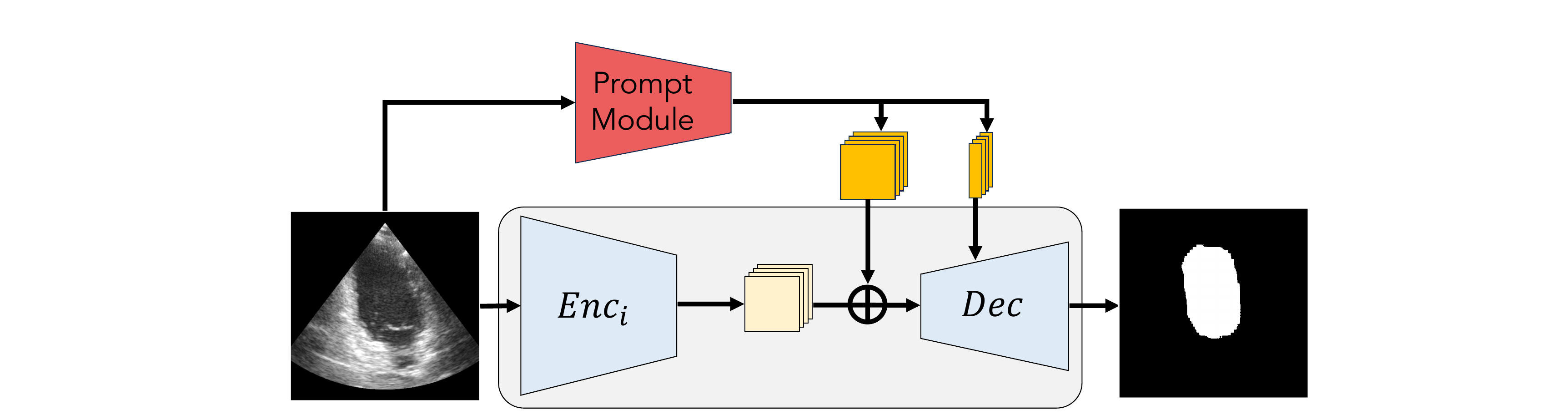}
        \caption{Inference with our prompt module}
        \label{subfig:framework_test}
    \end{subfigure}
\caption{\revision{Overview of our framework. Our prompt module fits on top of promptable foundation models. (\subref{subfig:framework_train}) During training, our module (\emph{red}) learns to generate relevant prompt embeddings (\emph{dark yellow}) from the image. Training requires only bounding box annotations and optimizes multiple losses. The components of the backbone foundation model (\emph{blue}) remain frozen during training. Red arrows indicate active gradient propagation. (\subref{subfig:framework_test}) At inference, our trained prompt module replaces the original prompt encoder of the foundation model, eliminating user interaction.}}
\label{fig:framework}
\end{figure*}

\section{Method}
Let $\xb \in \Real^{3 \times H \times W}$ be a 3-channel input image of height $H$ and width $W$. Suppose we only have access to a bounding box $\mb$ enclosing the object of interest, rather than a full ground-truth segmentation mask. Our approach automates and adapts promptable foundation models for new tasks using such weakly labeled data. We train a prompt module to generate a relevant prompt embedding from the input image using an objective function comprising three main components, detailed in Sections \ref{subsec:pseudolabel} to \ref{subsec:regularization}. The overall framework is illustrated in Fig.\ref{fig:framework}, and the training procedure is outlined in Alg.\ref{alg:training_strategy}.


\subsection{Add-on Prompt Module to Vision Foundation Model}
Although promptable vision foundation models are universal models that can generalize to varied tasks, they lack the ability to automatically segment a specific target object without user interaction. Our end-to-end method eliminates the reliance on user-defined prompts to automatically segment specified objects in medical images.

\subsubsection{Vision Foundation Model Architecture} 

The Segment Anything Model (SAM) \cite{kirillov_segment_2023} is used as our prototypical promptable vision foundation model architecture. SAM consists of three distinct components: an image encoder, a prompt encoder and a mask decoder, which we denote as $\mr{Enc}_{i}$, $\mr{Enc}_{p}$ and $\mr{Dec}$. As a promptable model, SAM takes as input a set of encoded prompts $\Pb = \{\Pb_1 \ldots \Pb_n \}$ where $\Pb_i \in \Real^D$, which can represent any combination of:
\begin{enumerate}\setlength\itemsep{.25em}
    \item Point coordinates: $\pb \in \Real^{2}$
    \item Bounding box boundary coordinates: $\bb = [\pb_1, \pb_2]$ with $\pb_1, \pb_2 \in \Real^{2}$, and
    \item A coarse mask: $\tilde \yb \in \{0, 1\}^{H \times W}$.
\end{enumerate}
Given $\xb$ and $\Pb$, SAM generates an image embedding  $\Zb_i \in \Real^{256 \times 64 \times 64}$, as well as a sparse and dense prompt embedding $\Zb_{ps}\in \Real^{256}$  and $\Zb_{pd}\in \Real^{256 \times 64 \times 64}$, following:
\begin{equation}
\begin{aligned}
    \Zb_i &= \mr{Enc}_{i}(\xb), \\
    \Zb_{p} &= \{\Zb_{ps}, \Zb_{pd}\} = \mr{Enc}_{p}(\Pb).
\end{aligned}
\end{equation}
The encoded image and prompts are then fed into the mask decoder to obtain a probability map.

\subsubsection{Prompt Module Architecture}
We automate the foundation model by training a prompt module $g_\theta$ to directly generate the prompt embedding $\Zb'_{p} = \{\Zb'_{pd}, \Zb'_{ps}\} = g_\theta(\xb)$ according to the target task. 
Following \cite{shaharabany_autosam_2023}, our prompt embedding generator module is composed of a Harmonic Dense Net \cite{chao_hardnet_2019} pre-trained on ImageNet as well as a decoder to produce an output of shape $256 \times 64 \times 64$. The Harmonic Dense Net takes as input the image $\xb$ and comprises six blocks with channel outputs of size 192, 256, 320, 480, 720, and 1280. The decoder consists of two upsampling blocks, each with two convolutional layers. Our experiments focus on $\Zb'_{pd}$ and use for $\Zb'_{ps}$ SAM's default sparse embedding given an empty input prompt.

\subsection{Pseudo-label loss from prompted foundation model} \label{subsec:pseudolabel}
Previous works have noted that bounding box prompts provide a less ambiguous spatial context for the object of interest \cite{mazurowski_segment_2023,huang_segment_2024,ma_segment_2024}. Hence, we take advantage of the fact that the provided tight bounding box $\mb$ can be converted into a box prompt $\bb$ by extracting the lower left and upper right coordinates of $\mb$. Given $\bb$, the promptable foundation model can generate a segmentation mask, which acts as a coarse pseudo-label to guide the learning process of the prompt module $g_\theta$. 
With the predicted output probabilities as $S_\theta(\xb)\!=\!\mr{Dec}(\Zb_i, g_\theta(\xb))$ and the pseudo-label obtained from the prompted foundation model as $S(\xb, \bb) =\llbracket \mr{Dec} \big(\Zb_i, \mr{Enc}_{p}(\bb) \big) \geq 0.5 \rrbracket$, we define our pseudo-label loss:
\begin{equation}
\label{eq:pseudo_loss}
\begin{split}
     \loss_{\mr{pseudo}}(\alpha, \beta, S(\xb, \bb)) \, =\, \alpha \cdot \mathrm{CE}\left( S_\theta (\xb), S(\xb, \bb) \right) 
     + \, \beta \cdot \mathrm{Dice} \left(S_\theta(\xb), S(\xb, \bb) \right).
\end{split}
\end{equation}

\subsection{Box-based constraints} \label{subsec:constraint}
Since the predictions of the foundation model are only a rough estimate of the true mask and may contain inaccuracies (see Fig.\ref{fig:PromptAmbiguity}), additional constraints should be used to guide the training process. 
Denoting as $\Omega_I$ and $\Omega_O$ the regions inside and outside the bounding box $\mb$ such that $\Omega_I \cup \Omega_O = \Omega \in \Real^{H \times W}$, we apply two box-based constraints on the output by exploiting the following facts:
\begin{itemize}\setlength\itemsep{.25em}
    \item The size of the bounding box sets a \textit{lower and upper limit} on the \textit{number of foreground pixels} 
    \item $\Omega_O$ contains \textit{only background pixels}
\end{itemize}

\subsubsection{Size Constraint} \label{subsubsec:loss_size}
The bounding box puts constraints on the size of the predicted mask. Thus, the sum of foreground predicted pixels should not be greater than the size of $\Omega_I$. Similarly, we can apply a prior on the size of the predicted foreground, given the size of $\Omega_I$. Setting the ratio $\epsilon_1, \epsilon_2 \in [0, 1]$ of pixels from $\Omega_I$ that should be classified as foreground, we obtain
\begin{equation}  \label{eq:size_constraint}
    \epsilon_1 |\Omega_I| \leq \sum_{(i, j) \in \Omega} S_\theta (\xb)_{ij} \leq \epsilon_2|\Omega_I|.
\end{equation}
Once again, $S_\theta(\xb) = \mr{Dec}(\Zb_i, g_\theta(\xb))$ and  $S_\theta(\xb)_{ij}$ refers to the pixel $(i, j$) of the output probabilities.

The condition can be translated into a constraint-based loss by applying a penalty function $\psi_t$, which becomes positive when the condition is not met, like a simple ReLU function. In this work, we employ a pseudo log-barrier function offering more stable optimization \cite{kervadec2022constrained}. The function $\psi_t(x)$ approximates a hard barrier as $t \to \infty$, where $\psi_t(x) = \infty$ if $x > 0$, and $\psi_t(x) = 0$ otherwise. The size-based loss can then be formulated as
\begin{equation}
\label{eq:size_loss}
\begin{split}
 \loss_{\mr{size}}(\epsilon_1, \epsilon_2, \Omega_I) = \psi_t \bigg(\epsilon_1 |\Omega_I| - \displaystyle\sum_{(i,j) \in \Omega} S_\theta (\xb)_{ij})\bigg) 
 +\  \psi_t \bigg(\displaystyle\sum_{(i, j) \in \Omega} S_\theta (\xb)_{ij} - \epsilon_2 |\Omega_I|\bigg).
\end{split}
\end{equation}

\subsubsection{Emptiness constraint} \label{subsubsec:loss_emptiness}
The bounding box also clearly defines a region that should contain only background pixels (i.e., $\Omega_O$, the regions outside the box). This emptiness constraint inside $\Omega_O$ can be expressed as a cross-entropy loss on the background pixels:
\begin{equation}
\label{eq:emptiness_loss}
    \loss_{\mr{empty}}(\Omega_O) \, = \, -\!\!\displaystyle\sum_{(i, j) \in \Omega_O} \!\!\log (1 - S_\theta (\xb)_{ij})
\end{equation}

\subsection{Consistency-based regularization} \label{subsec:regularization}

We propose a regularization strategy based on transformation consistency to alleviate the problem of over-fitting when training with few weakly-annotated samples. As the computational bottleneck of the foundation model is its image encoder, this strategy operates directly on features from the encoder instead of image pixels, as in traditional approaches. Following previous definitions, we denote $\Zb_i \in \RR^{256 \times 64 \times 64}$ the embedding of an image $\xb$ obtained with the original image encoder. When training our prompt module $g_\theta$, we randomly sample a geometric transformation $T$ (combination of random rotation/flip) and apply it on the image encoded features to obtain $T(\Zb_i)$. We then enforce the predictions of the segmentation decoder to be equivariant using an L2 consistency loss. Noting $S'_\theta \big( T(\xb) \big) = \mr{Dec} \big( T(\Zb_i), g_\theta \big( T(\xb) \big) \big)$:
\begin{equation} \label{eq:consistency_loss}
\loss_{cons}(T) \, = \, \left\|S'_\theta \big( T(\xb) \big) \, - \, T \left(S_\theta(\xb) \right) \right\| ^2_2.
\end{equation}

\subsection{Box-based Multi-loss Optimization Framework}  \label{subsubsec:loss_final}
In summary, training the prompt module involves optimizing four losses conditioned on a prompt-based pseudo-label, size and emptiness constraints and a consistency regularization.
Combining equations \eqref{eq:pseudo_loss}, \eqref{eq:size_loss}, \eqref{eq:emptiness_loss} and \eqref{eq:consistency_loss}, the final loss to optimize becomes:
\begin{equation}\label{eq:final_loss}
    \loss_{\mr{total}} = \lambda_1 \loss_{\mr{pseudo}} + \lambda_2 \loss_{\mr{size}} + \lambda_3 \loss_{\mr{empty}} + \lambda_4 \loss_{\mr{cons}},
\end{equation}
where $\lambda_i$'s are weights applied to each individual loss.

\begin{algorithm}
\caption{Prompt Module Training Procedure via our Box-based Multi-loss Optimization Scheme}
\label{alg:training_strategy}
\begin{algorithmic}[1]
\Require $g_\theta$: prompt module to train, $\mathcal{D}_{train} = \{\xb^{(k)}, \mb^{(k)}\}_{k = 1}^{N}$: training set, $\mr{Enc}_i, \mr{Enc}_p, \mr{Dec}$: components of foundation model
\Ensure $\alpha$, $\beta$, $\epsilon_1$, $\epsilon_2$, $\psi_t$, $T$, $\lambda_1$, $\lambda_2$, $\lambda_3$, $\lambda_4$

 \ForAll{$(\xb^{(k)}, \mb^{(k)}) \in \mathcal{D}_{train}$}
 
    \State $\Zb_i \gets \mr{Enc}_i(\xb^{(k)})$

    \vspace{1em}
    \State \texttt{// Pseudo-label loss}
    \State Convert $\mb^{(k)}$ to the box prompt $\bb$
    \State $S(\xb, \bb)\, \gets\, \llbracket \mr{Dec} \big(\Zb_i, \mr{Enc}_{p}(\bb) \big) \geq 0.5 \rrbracket$
    \State Compute $\loss_{\mr{pseudo}}(\alpha, \beta, S(\xb, \bb))$ using \eqref{eq:pseudo_loss}
    
    \vspace{1em}
    \State \texttt{// Constraint-based losses}
    \State $\Omega_I$: foreground region of $\mb^{(k)}$ (inside box)
    \State $\Omega_O$: background region of $\mb^{(k)}$ (outside box)
    \State Compute $\loss_{\mr{size}}(\epsilon_1, \epsilon_2, \Omega_I)$ using \eqref{eq:size_loss}

    \State Compute $\loss_{\mr{empty}}(\Omega_O)$ using \eqref{eq:emptiness_loss}
    
    \vspace{1em}
    \State \texttt{// Consistency-based regularization}
    \State Compute $\loss_{\mr{cons}}(T)$ using \eqref{eq:consistency_loss}

    \vspace{1em}
    \State $\loss_{\mr{total}} \gets \lambda_1 \loss_{\mr{pseudo}} + \lambda_2 \loss_{\mr{size}} + \lambda_3 \loss_{\mr{empty}}  + \lambda_4 \loss_{\mr{cons}}$
    
    \State Update weights $\theta$ of $g$
\EndFor
\end{algorithmic}
\end{algorithm}

\section{Experiments and Results}

\begin{figure*}[htb!]
\centering
\setlength{\tabcolsep}{.5pt}
\begin{tabular}{cccccccc}

    \rotatebox{90}{\hspace{2.5em}HC18} & 
     \includegraphics[width=.135\linewidth]{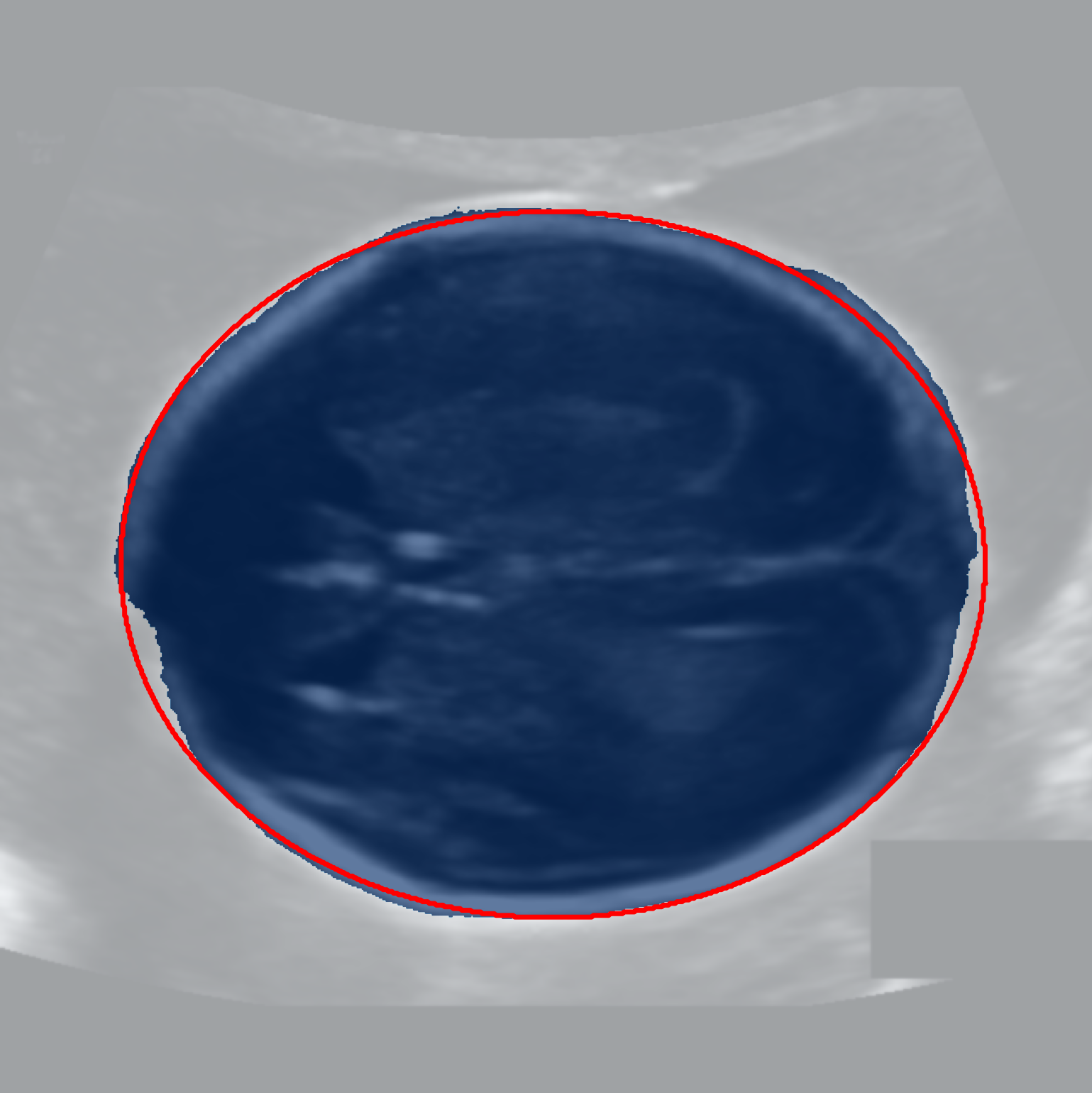}&
    \includegraphics[width=.135\linewidth]{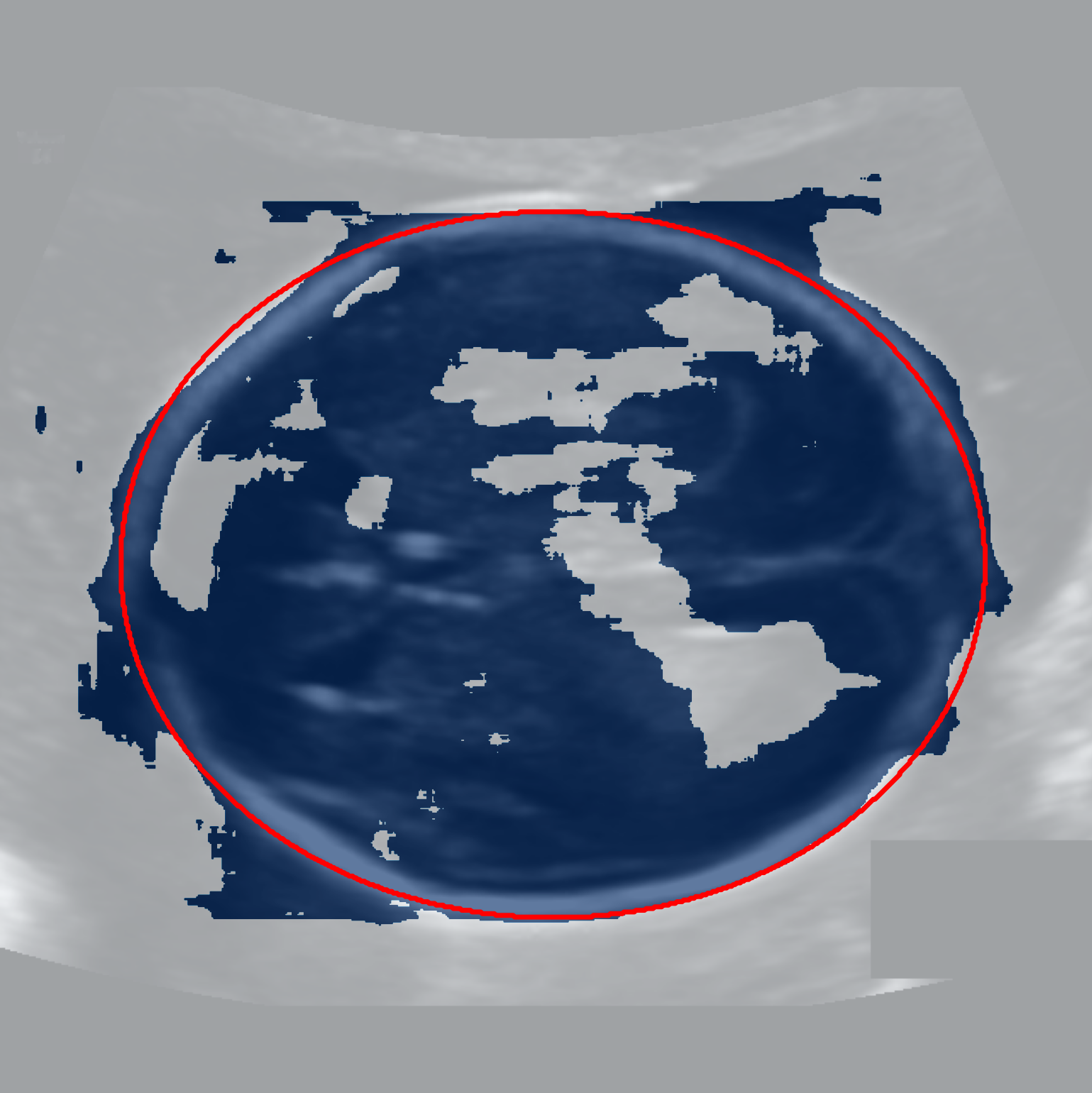} &
    \includegraphics[width=.135\linewidth]{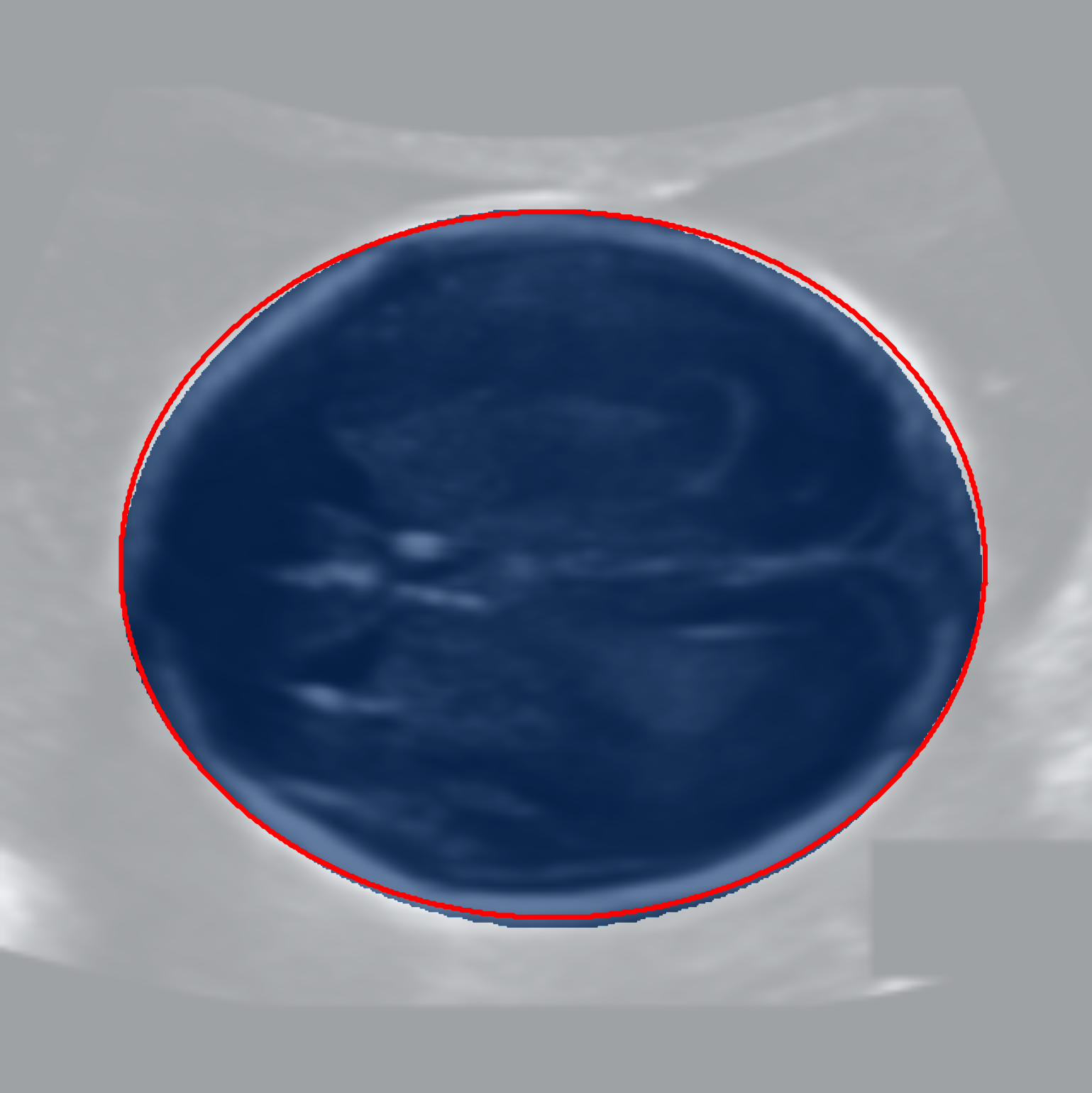} &
    \includegraphics[width=.135\linewidth]{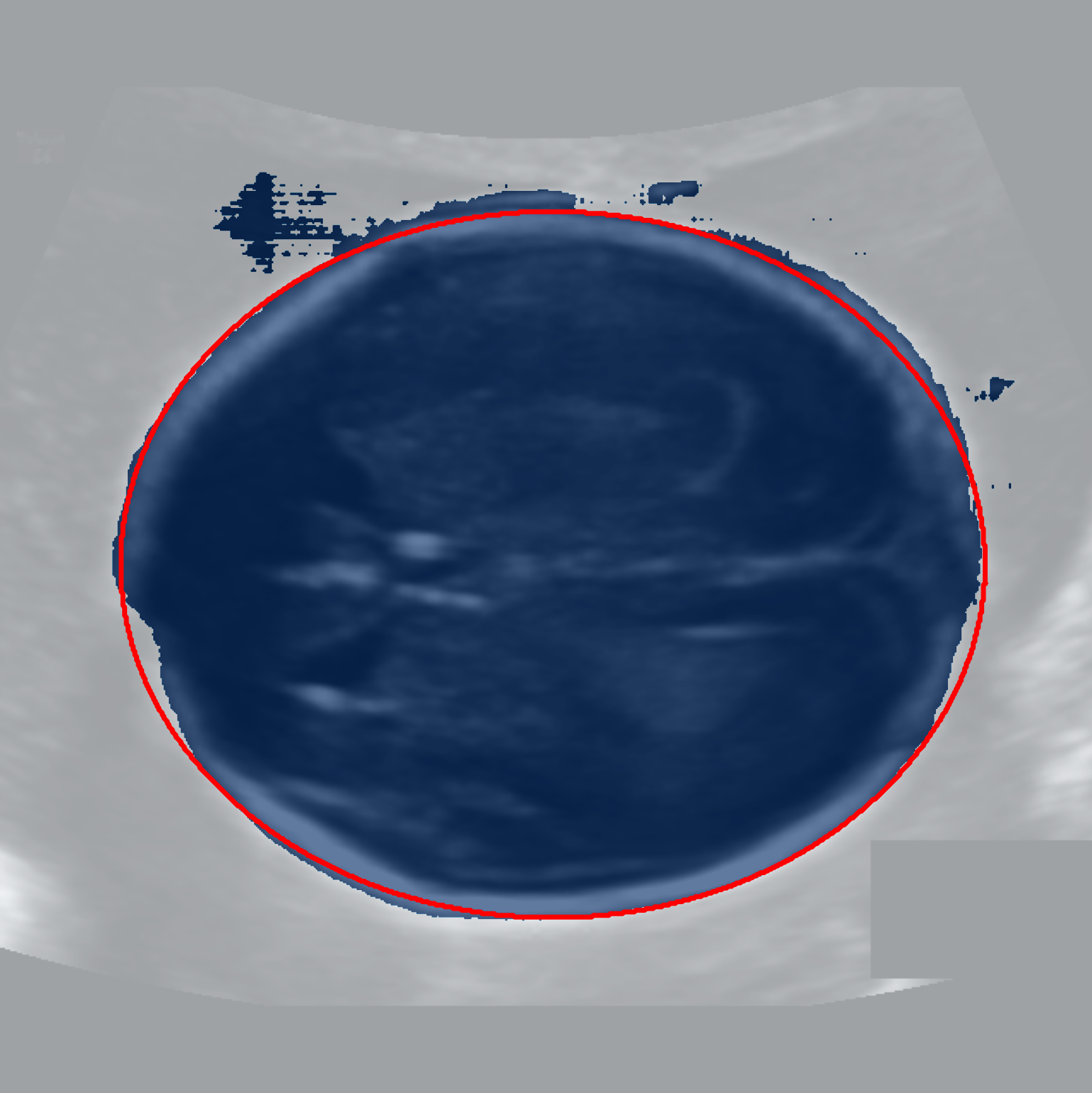} & 
    \includegraphics[width=.135\linewidth]{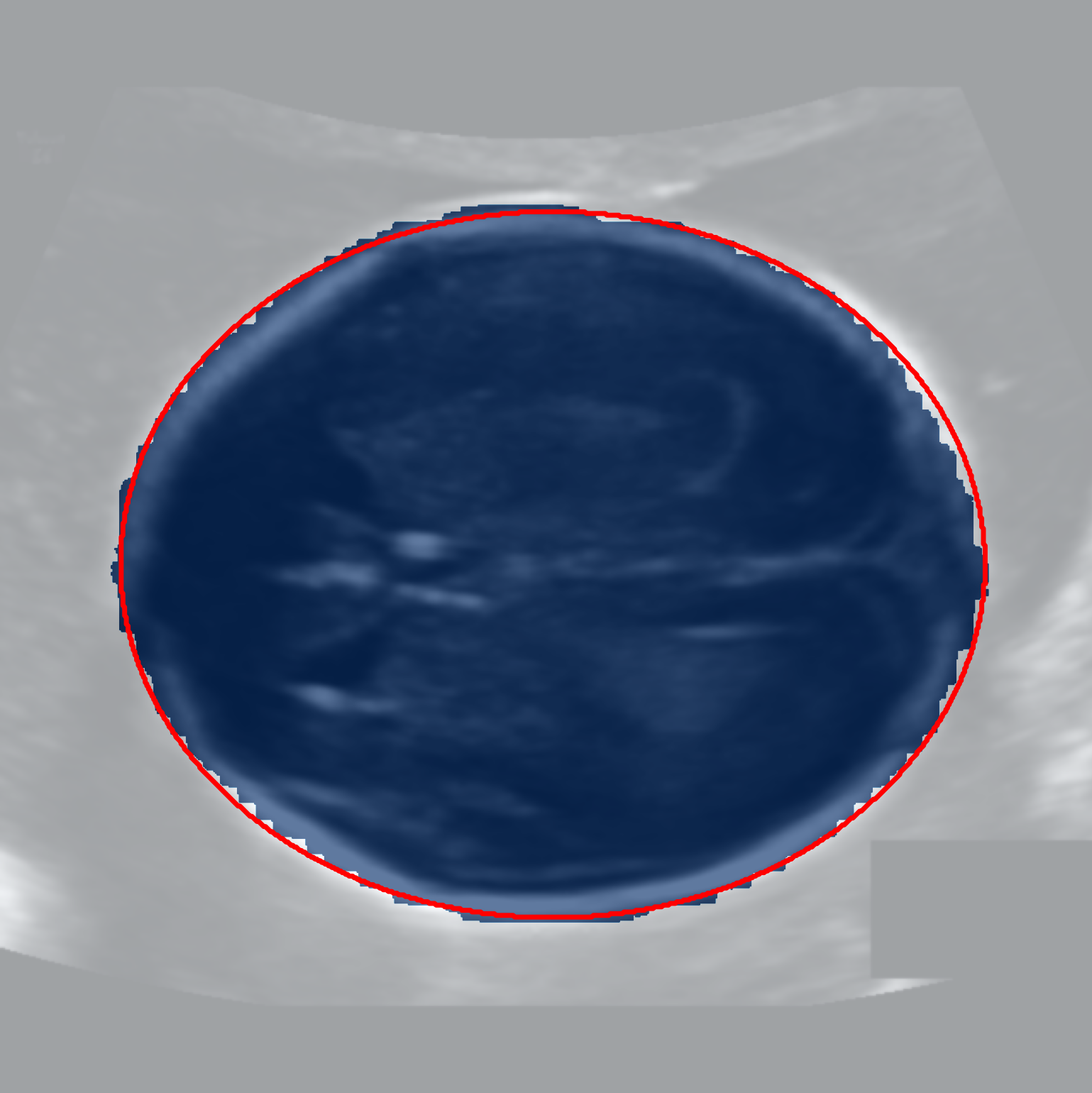} & 
    \includegraphics[width=.135\linewidth]{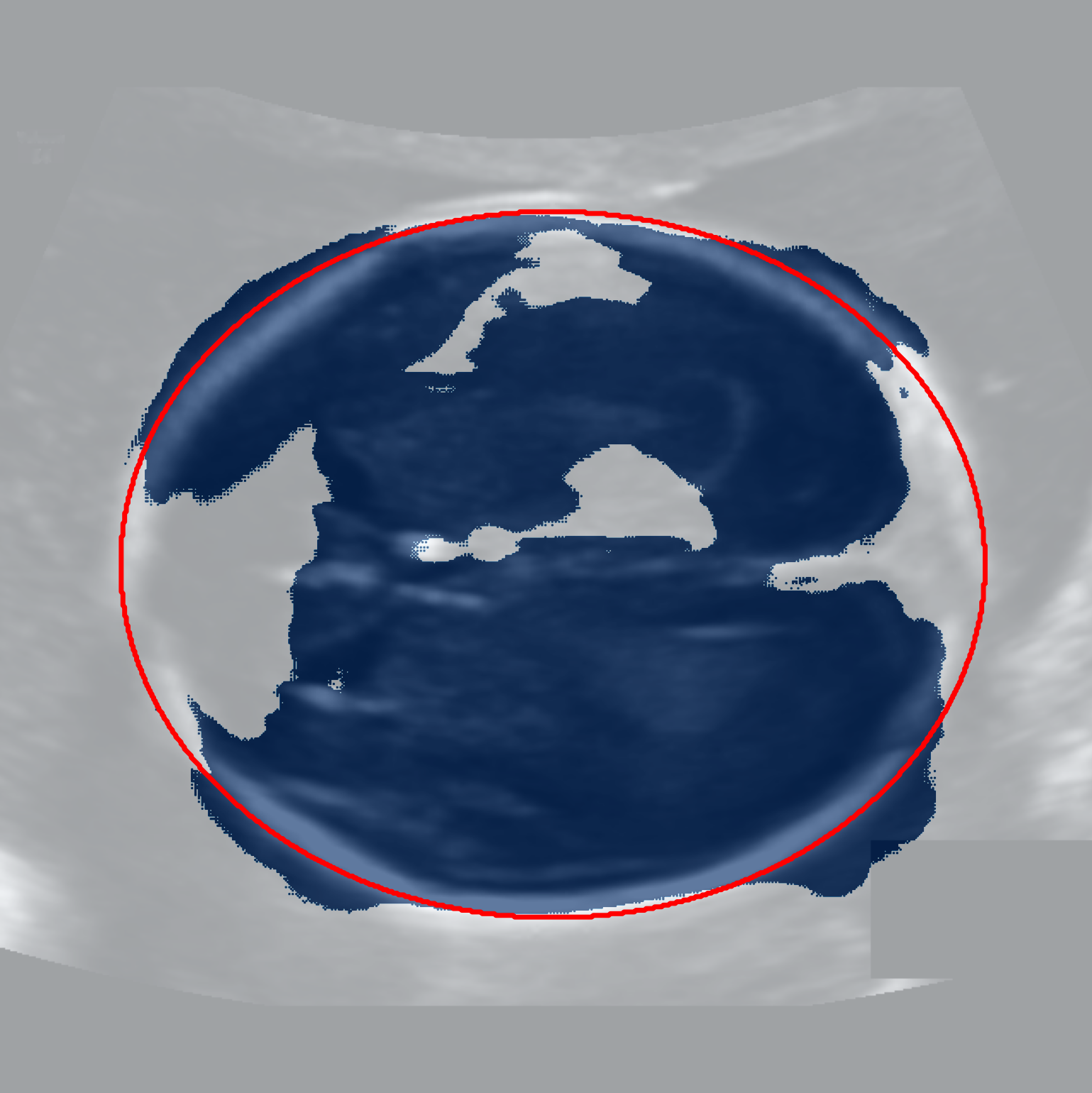} & 
    \includegraphics[width=.135\linewidth]{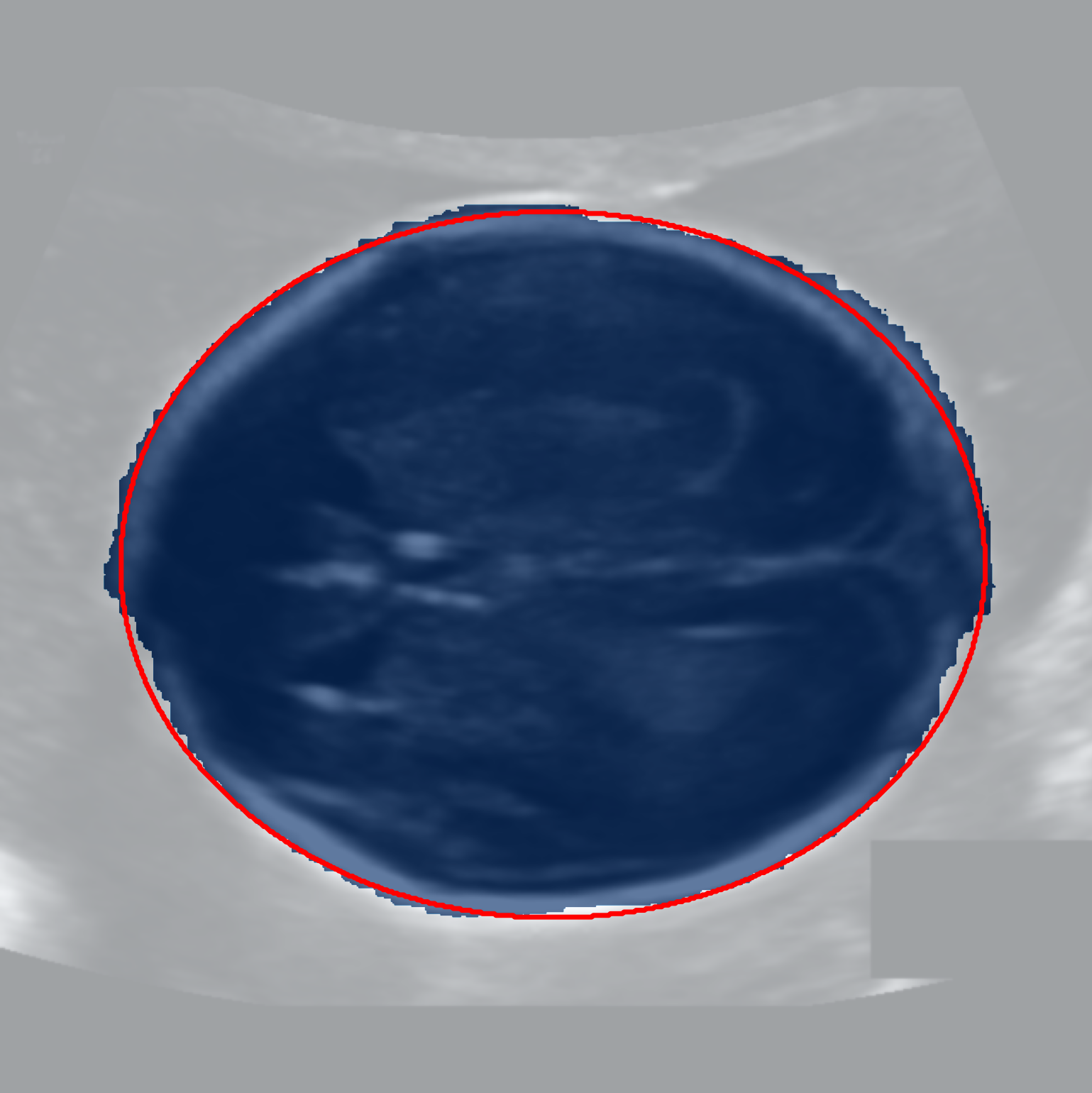} \\

    \rotatebox{90}{\hspace{1.5em}CAMUS} & 
     \includegraphics[width=.135\linewidth]{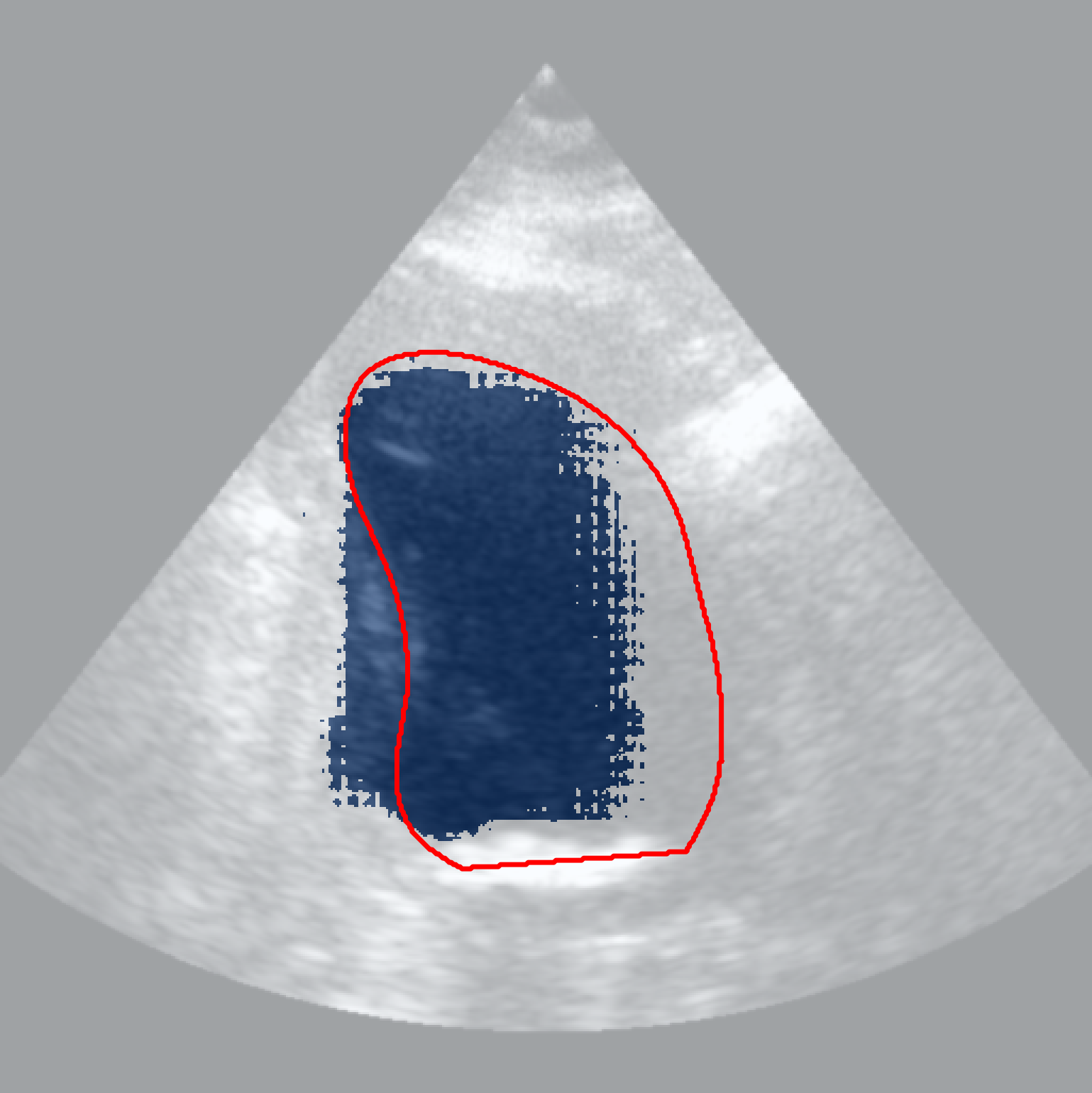}&
    \includegraphics[width=.135\linewidth]{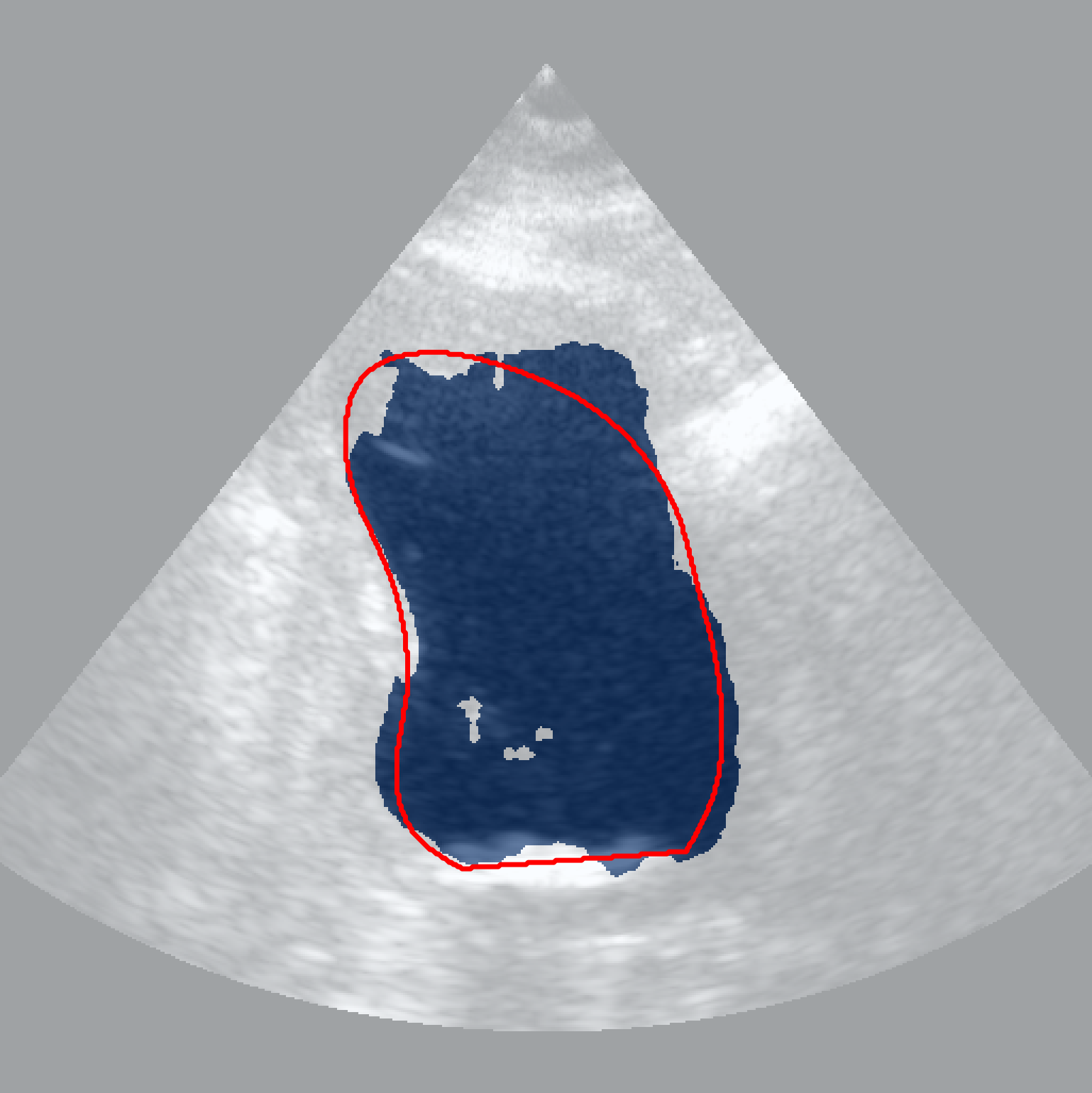} &
    \includegraphics[width=.135\linewidth]{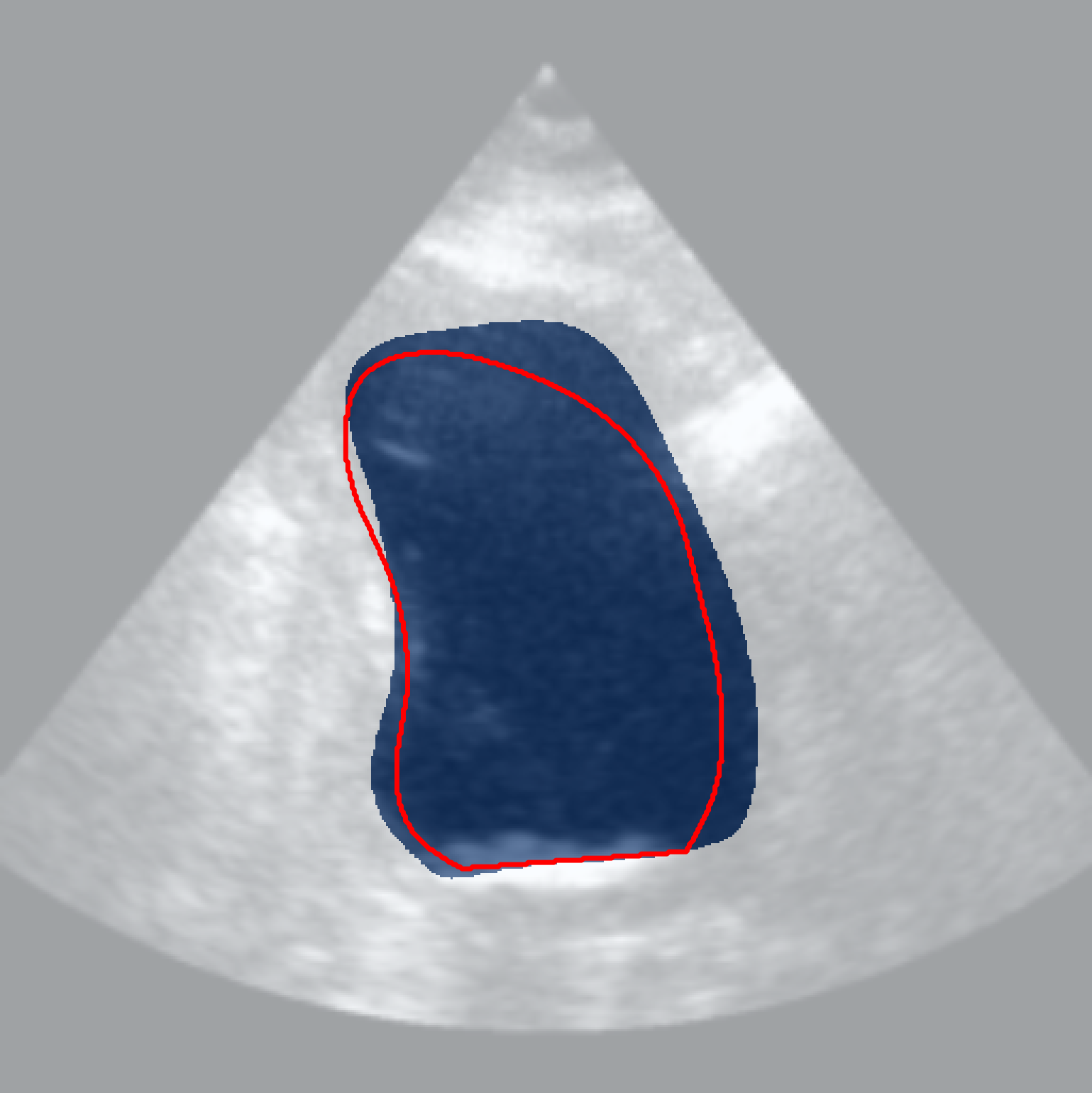} &
    \includegraphics[width=.135\linewidth]{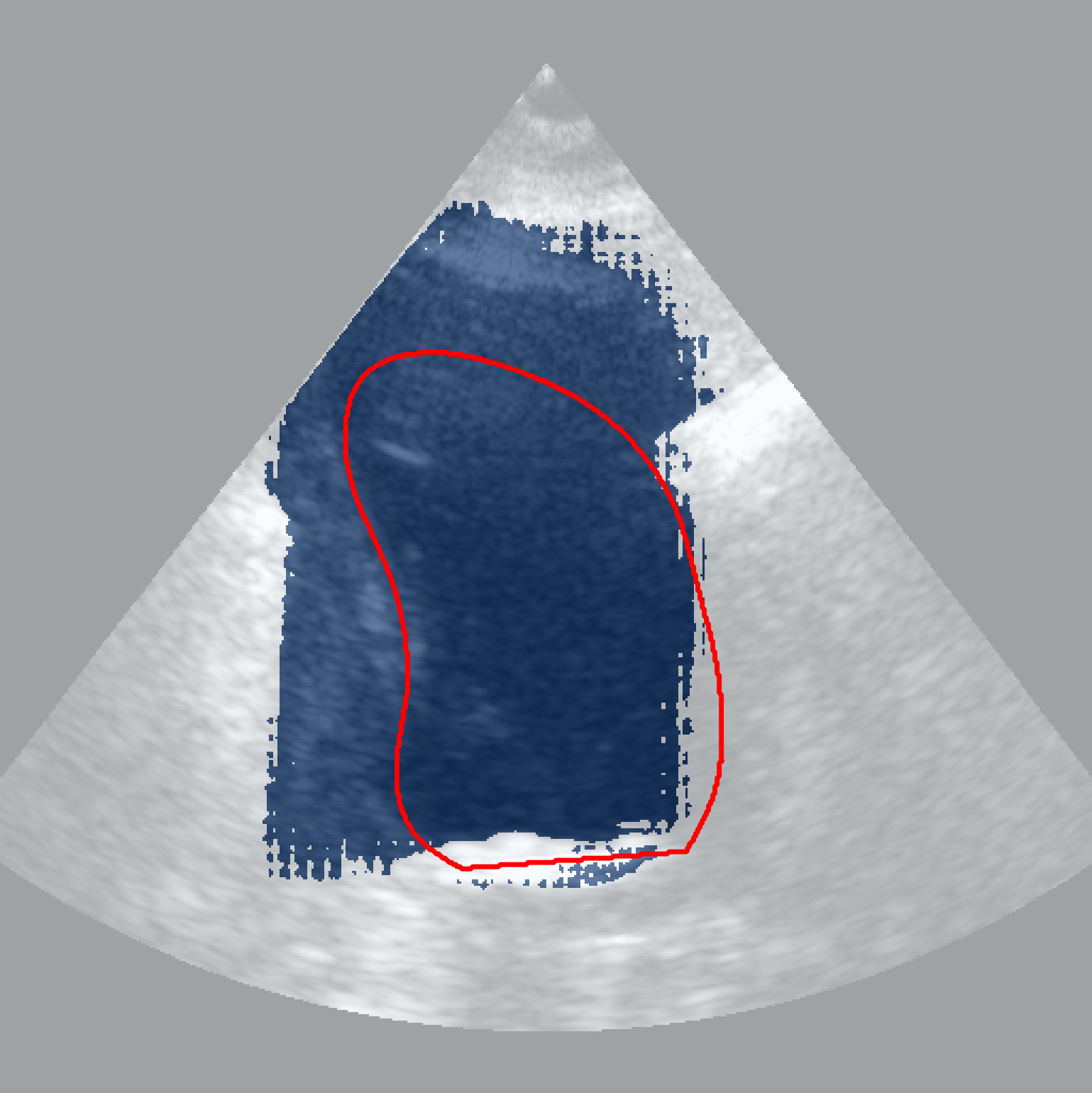} & 
    \includegraphics[width=.135\linewidth]{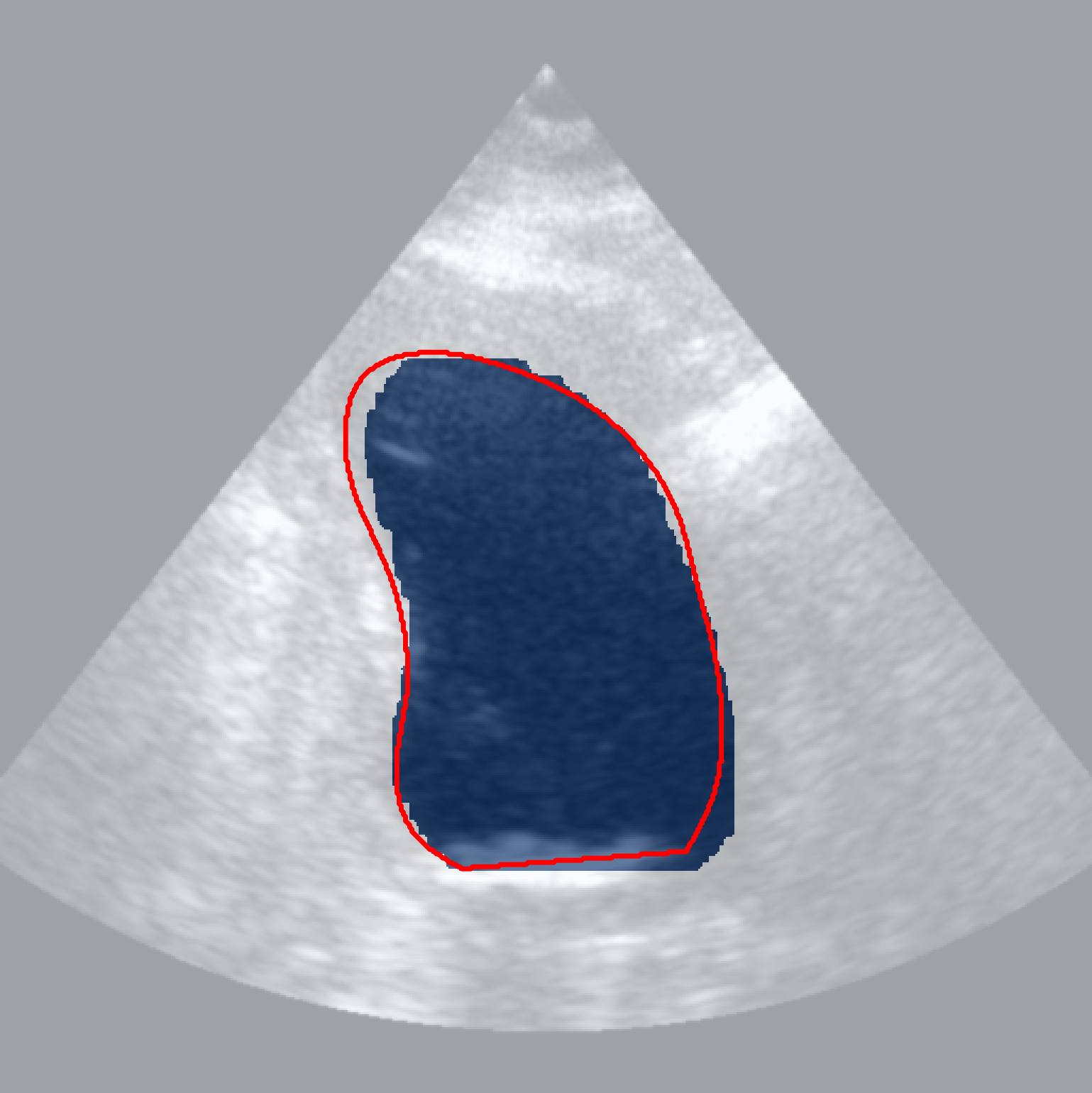} & 
    \includegraphics[width=.135\linewidth]{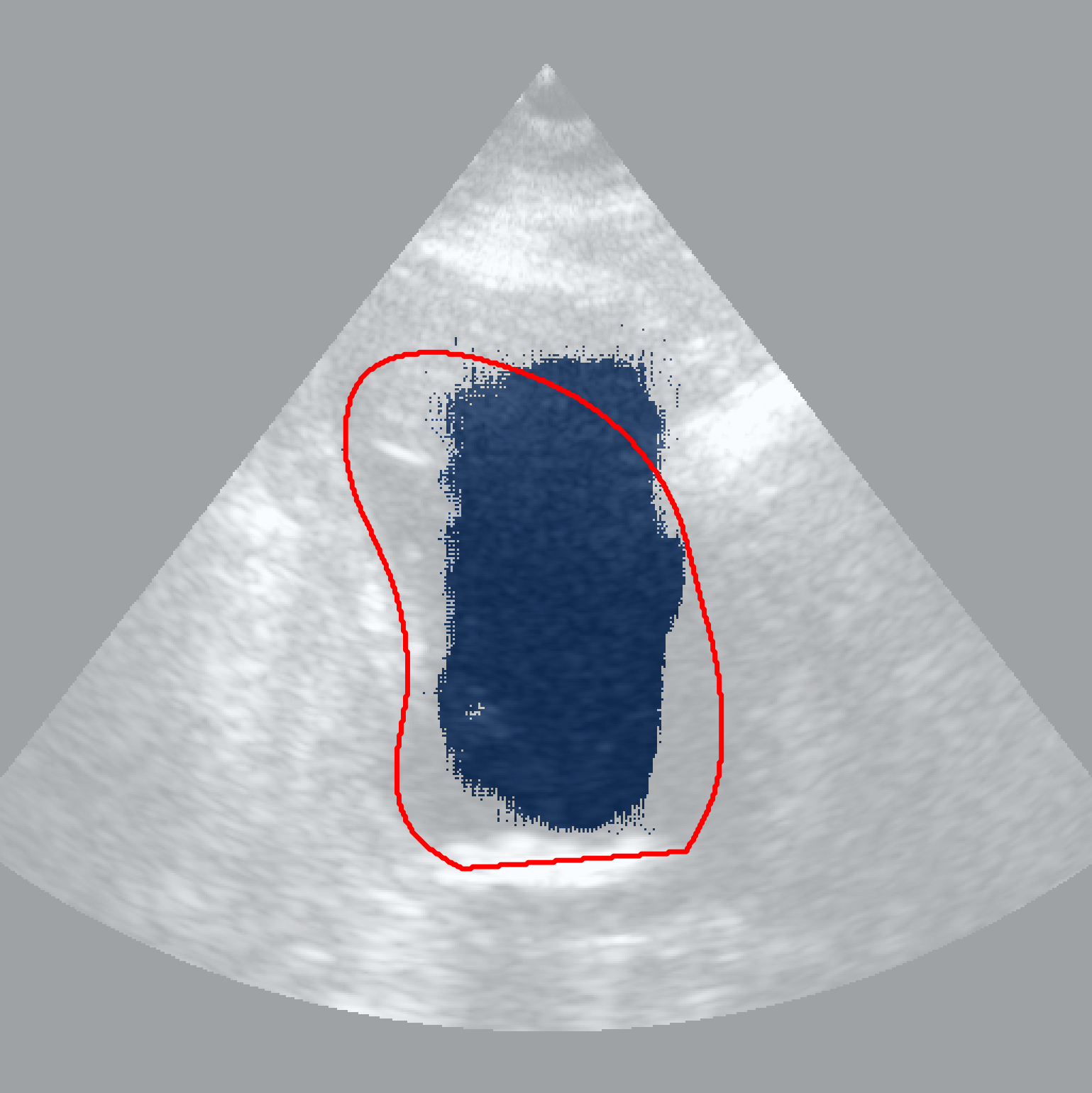} & 
    \includegraphics[width=.135\linewidth]{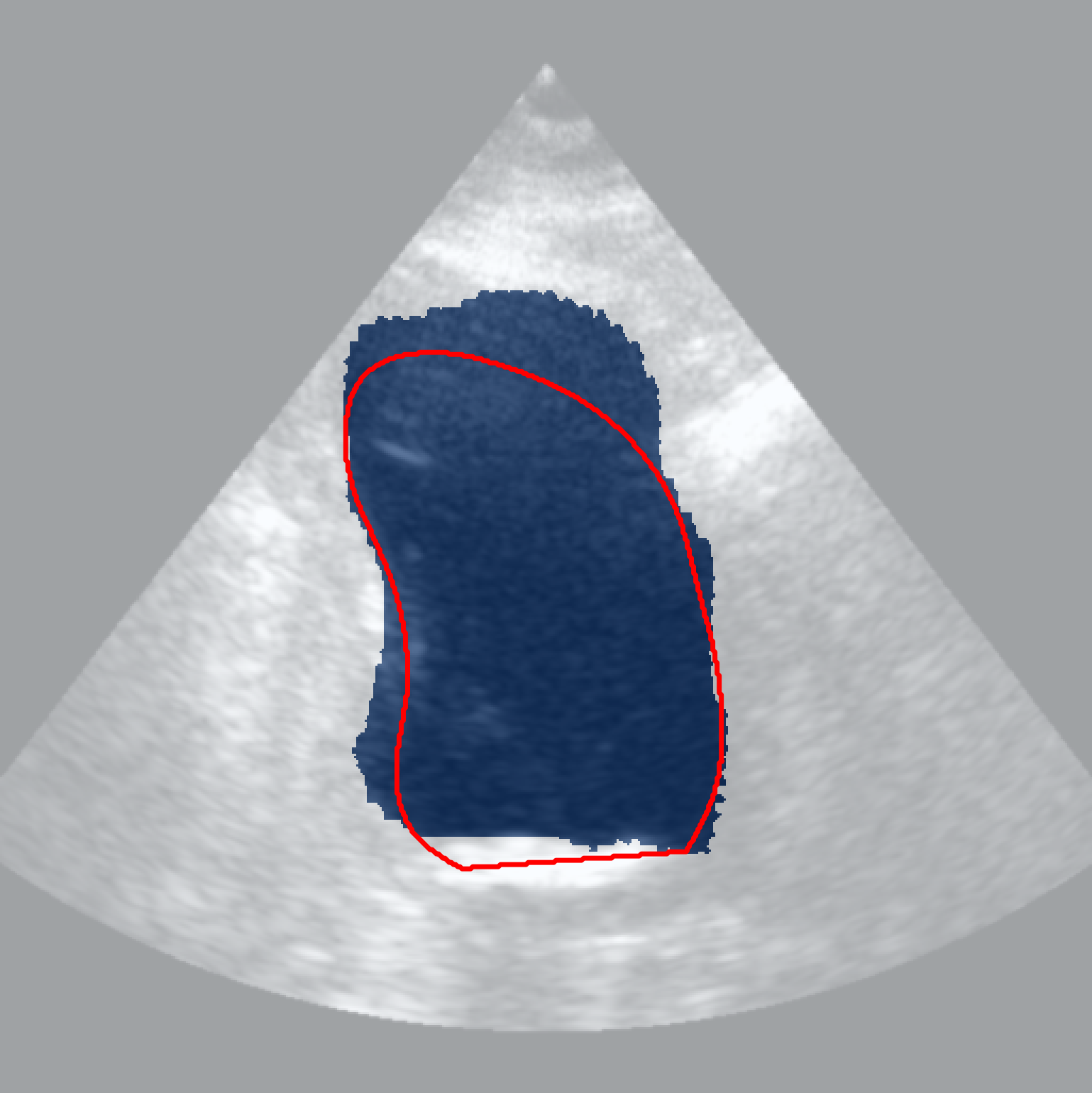} \\

    \rotatebox{90}{\hspace{1em}\revision{ACDC-LV}} & 
     \includegraphics[width=.135\linewidth]{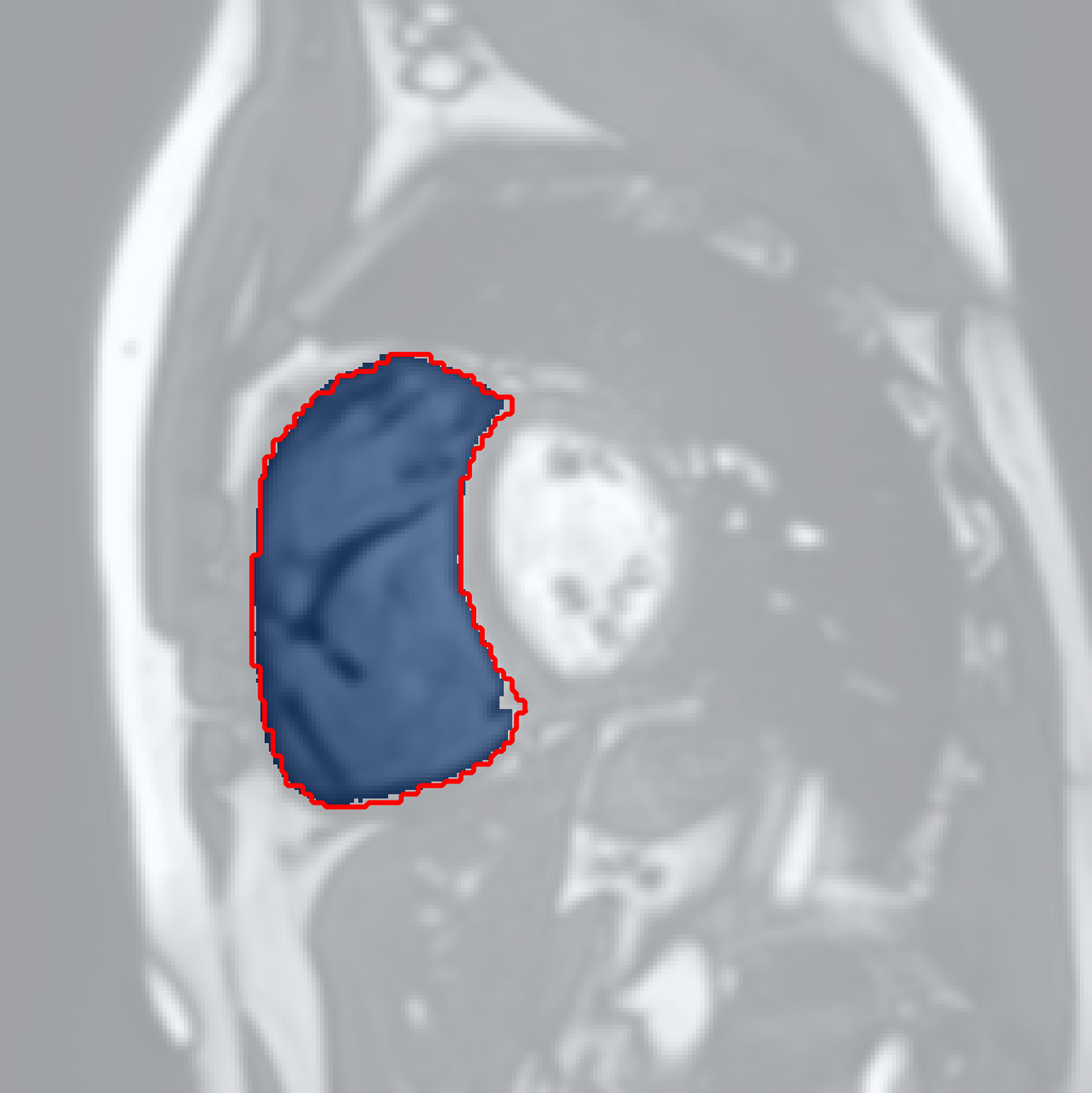}&
    \includegraphics[width=.135\linewidth]{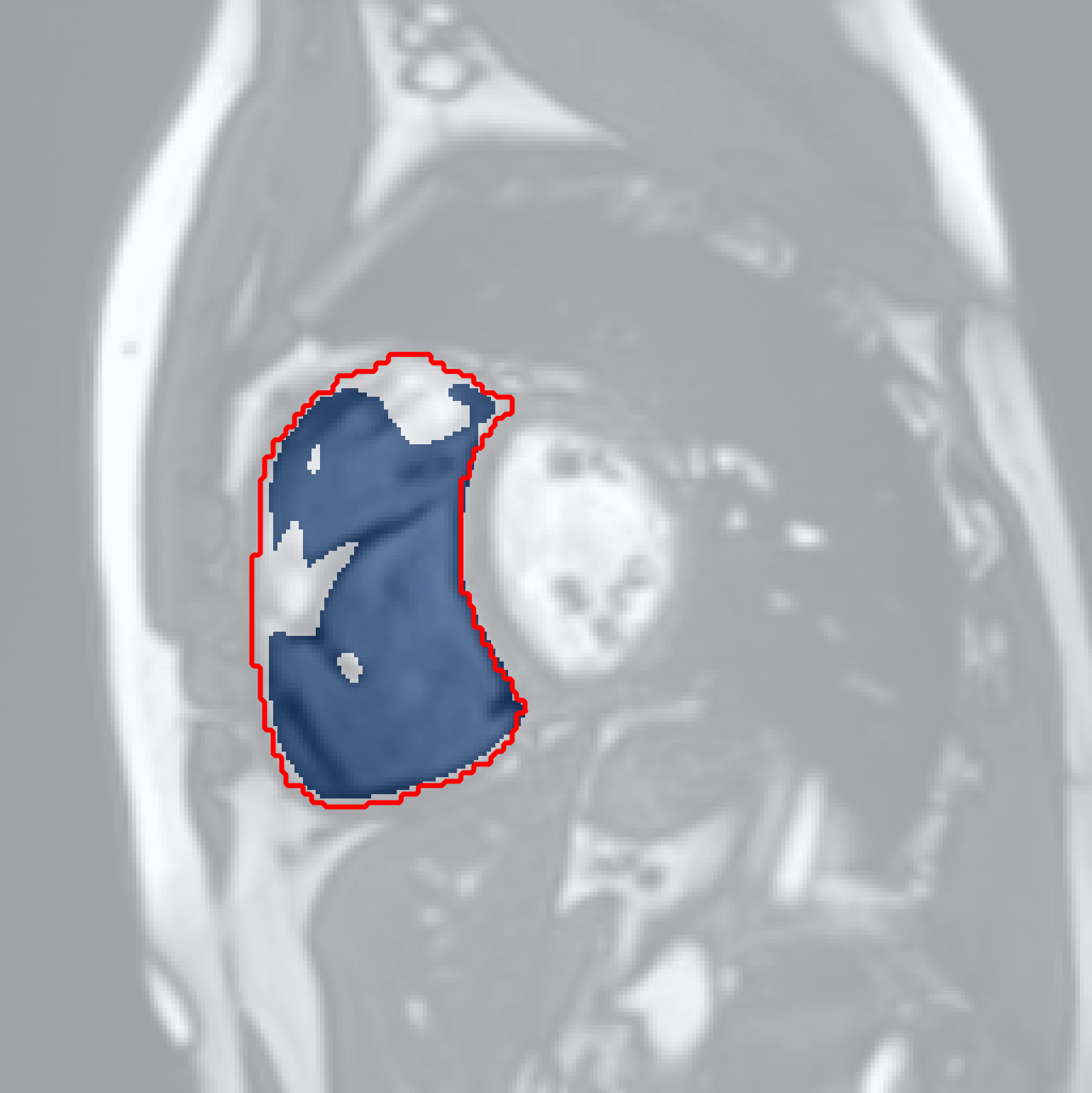} &
    \includegraphics[width=.135\linewidth]{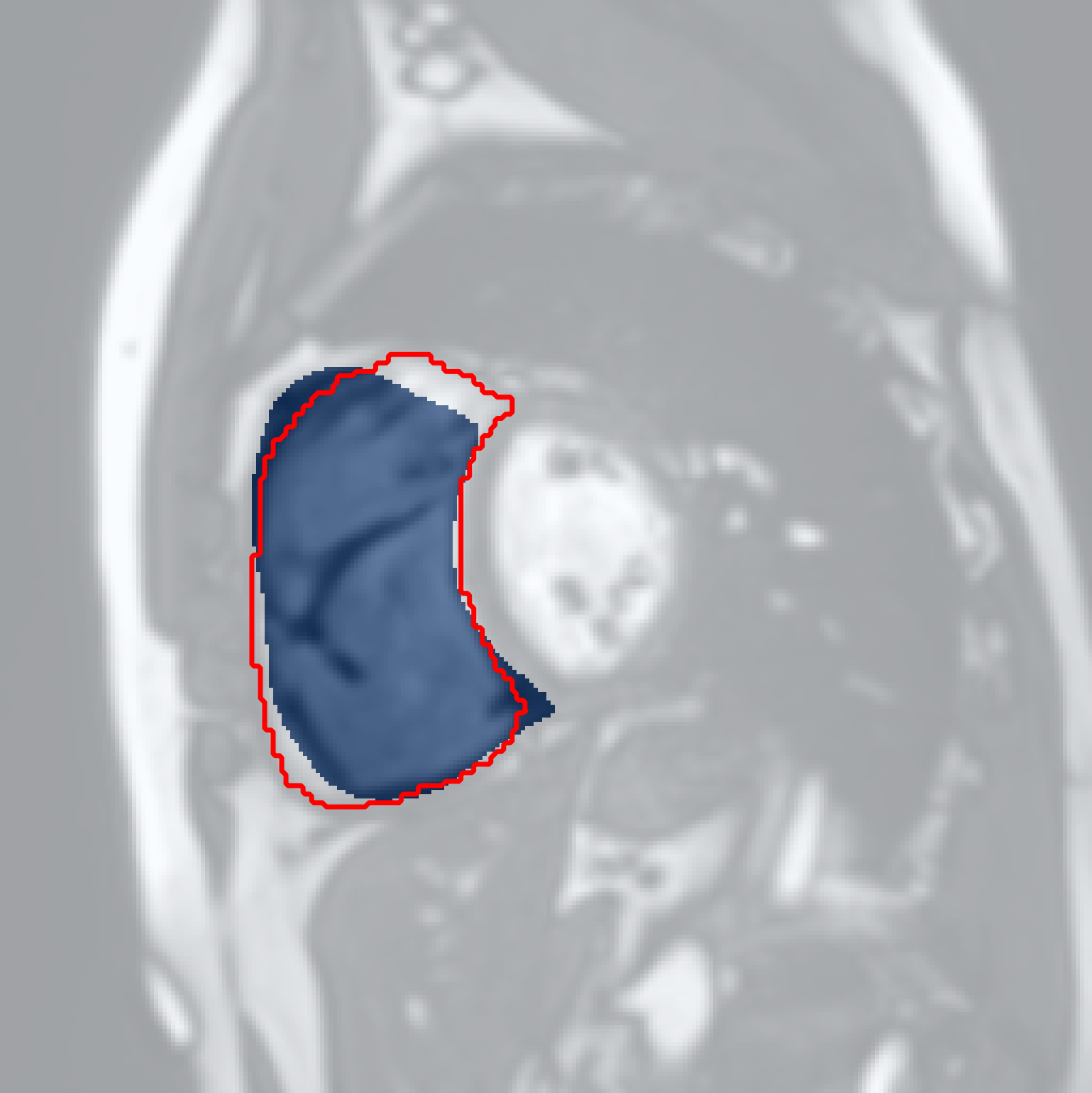} &
    \includegraphics[width=.135\linewidth]{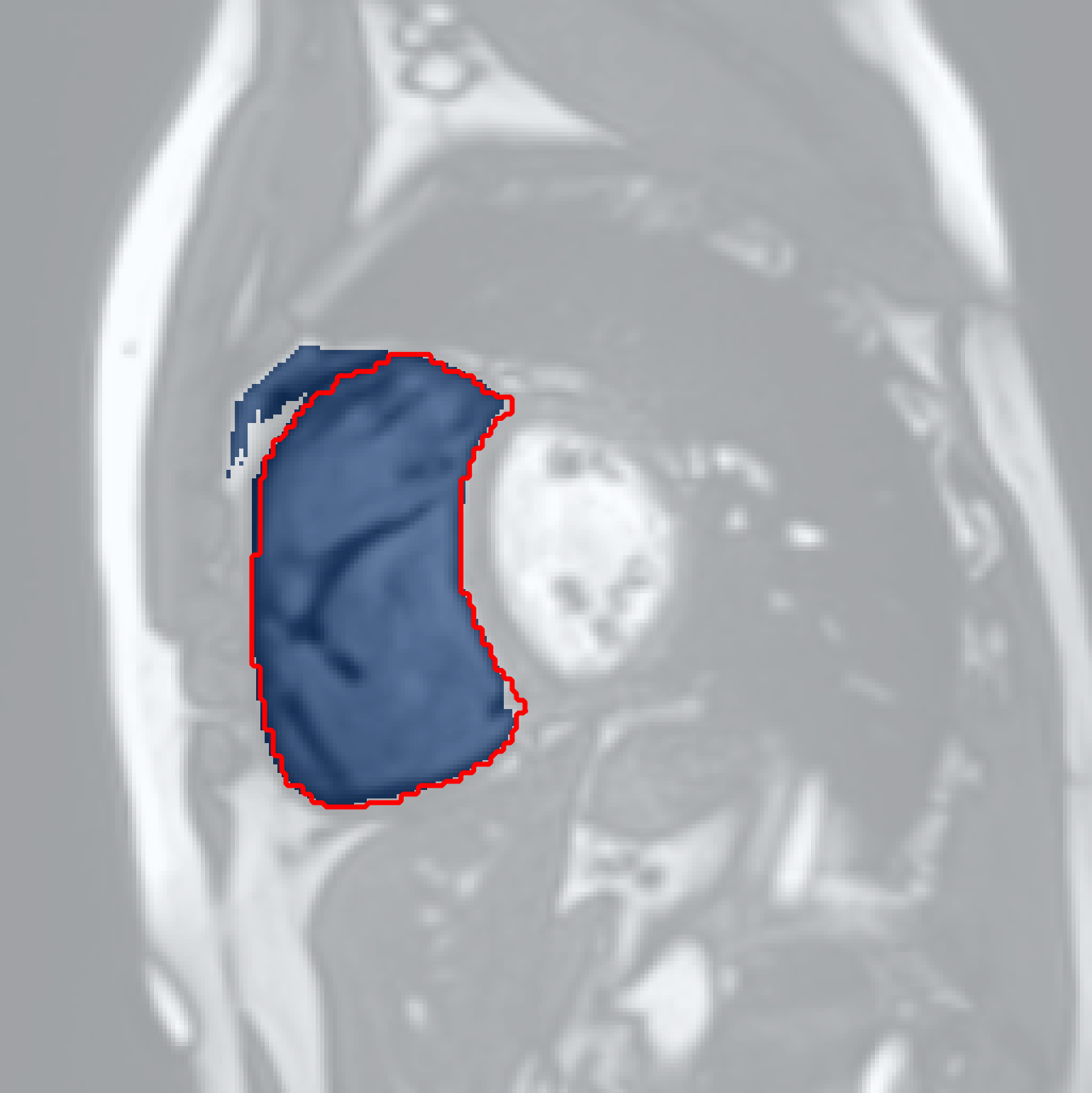} & 
    \includegraphics[width=.135\linewidth]{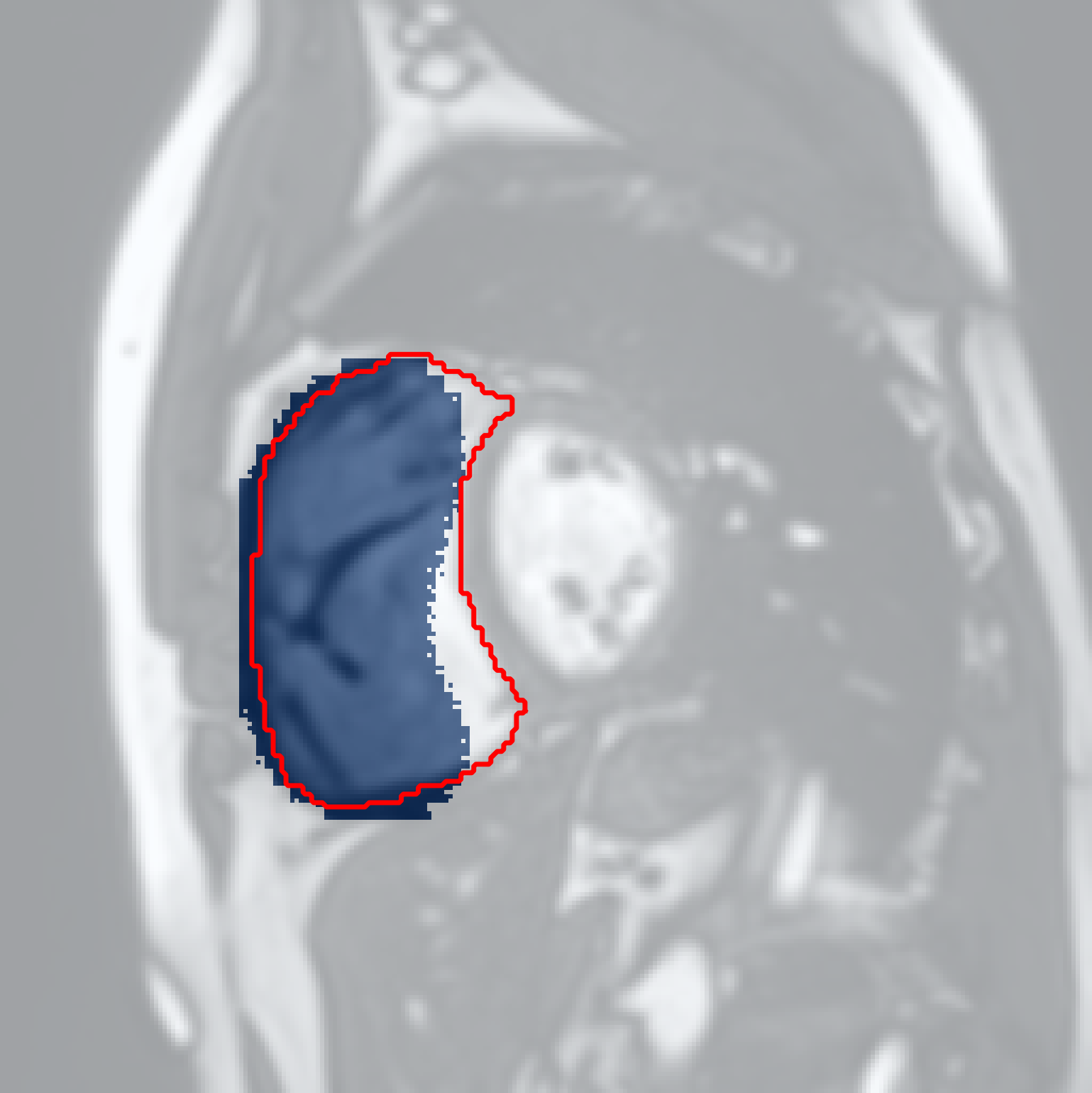} & 
    \includegraphics[width=.135\linewidth]{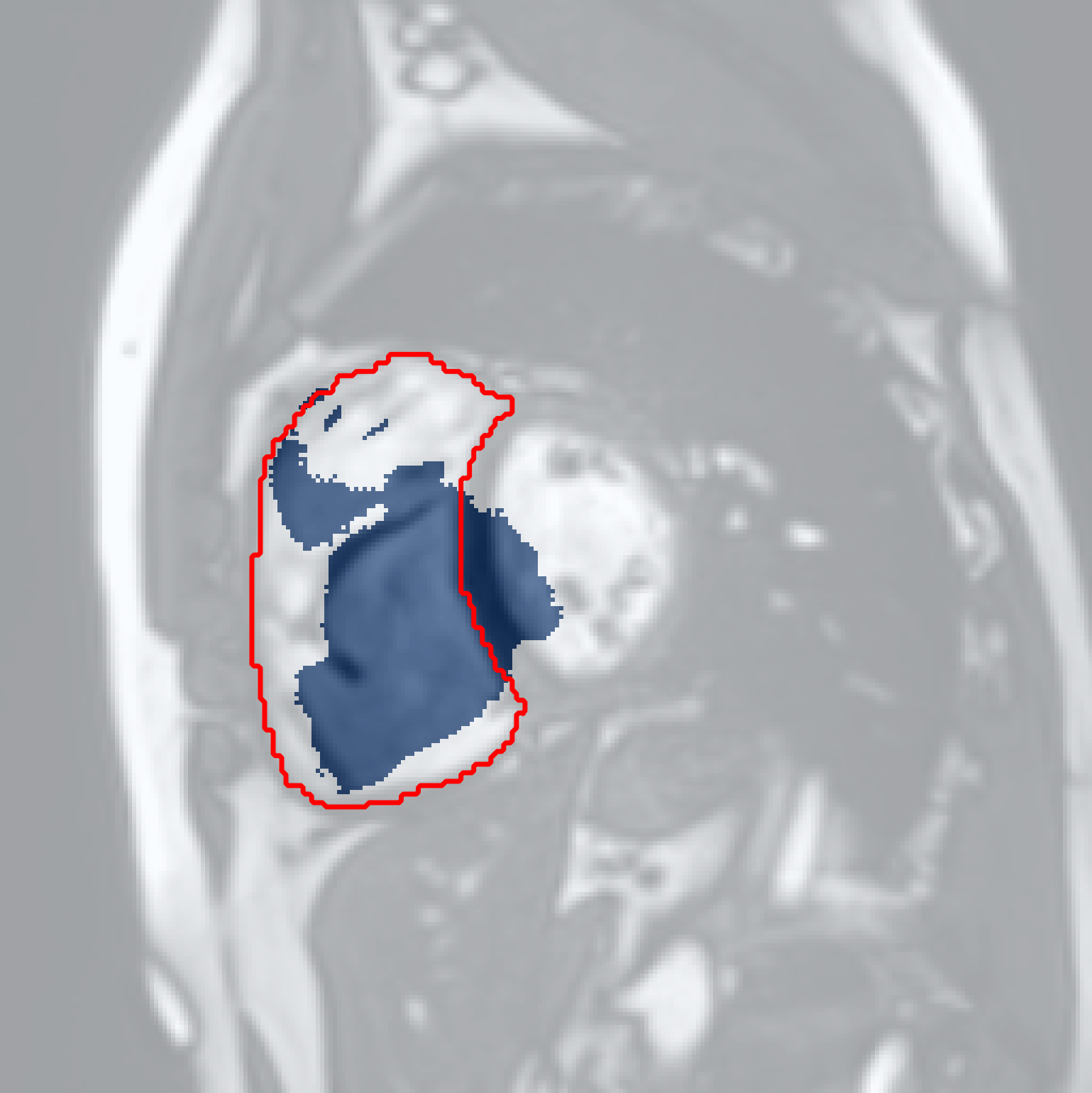} & 
    \includegraphics[width=.135\linewidth]{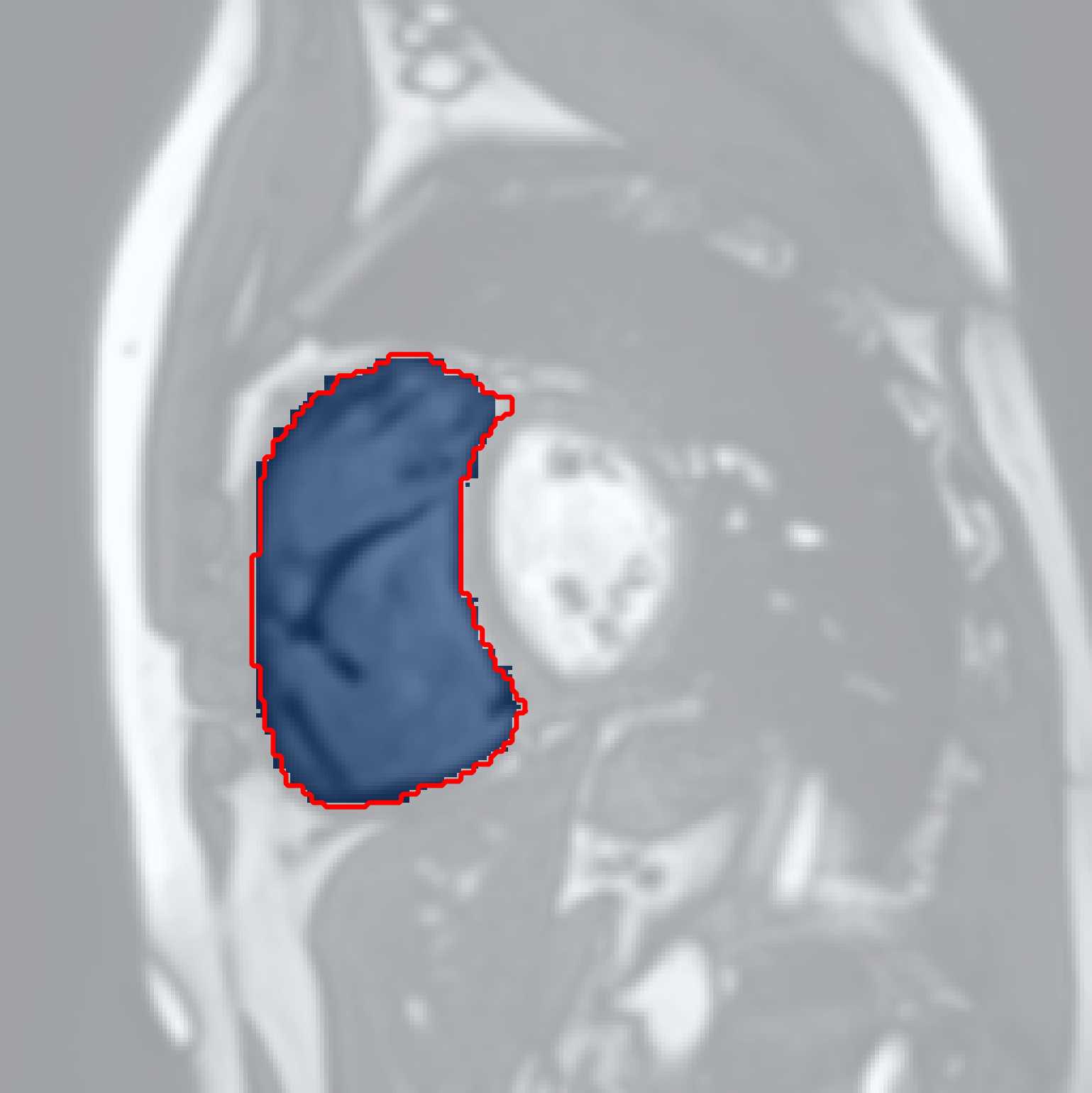} \\

    \rotatebox{90}{\hspace{1em}\revision{MSD-Spleen}} & 
    \includegraphics[width=.135\linewidth]{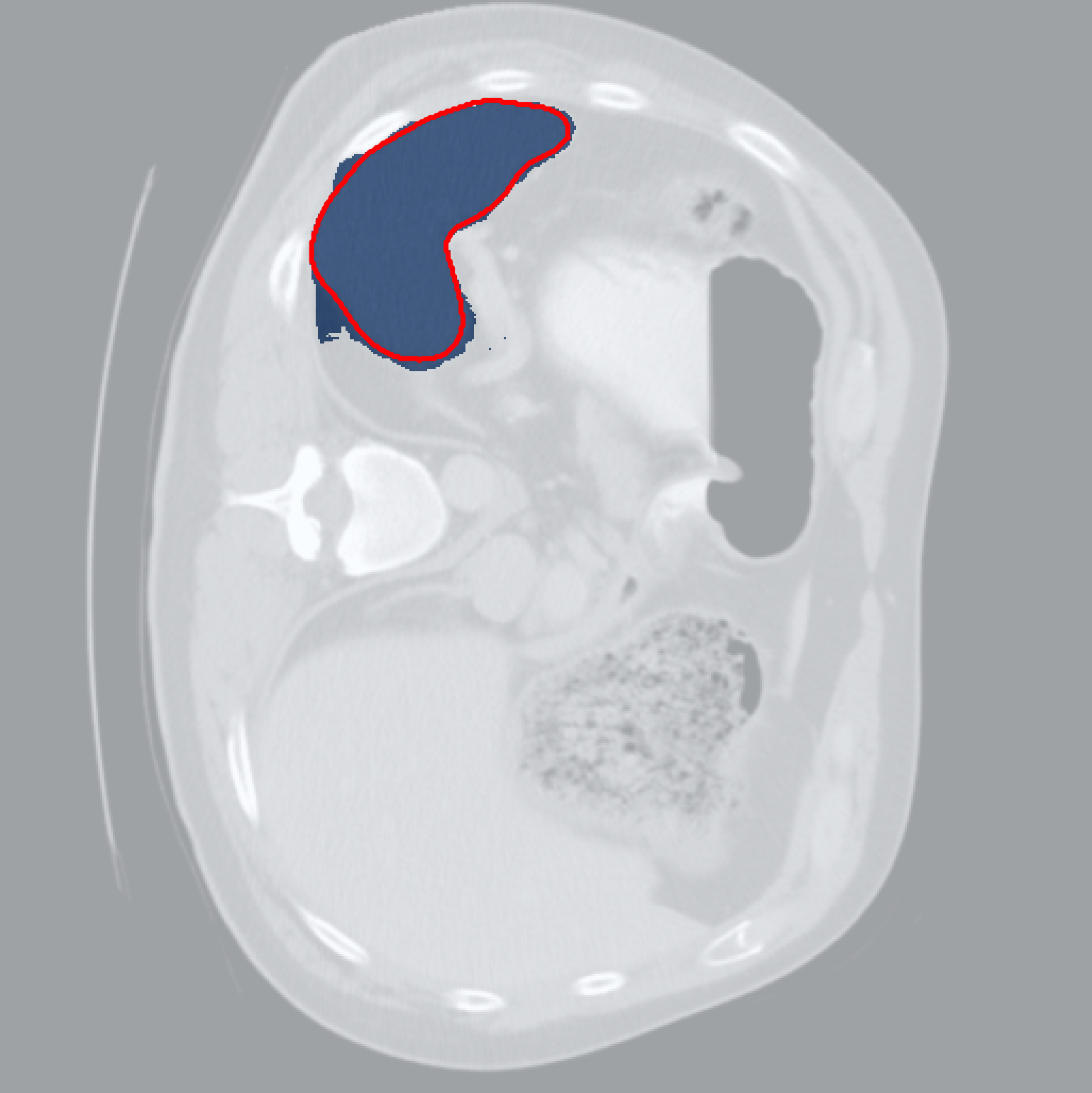} &
    \includegraphics[width=.135\linewidth]{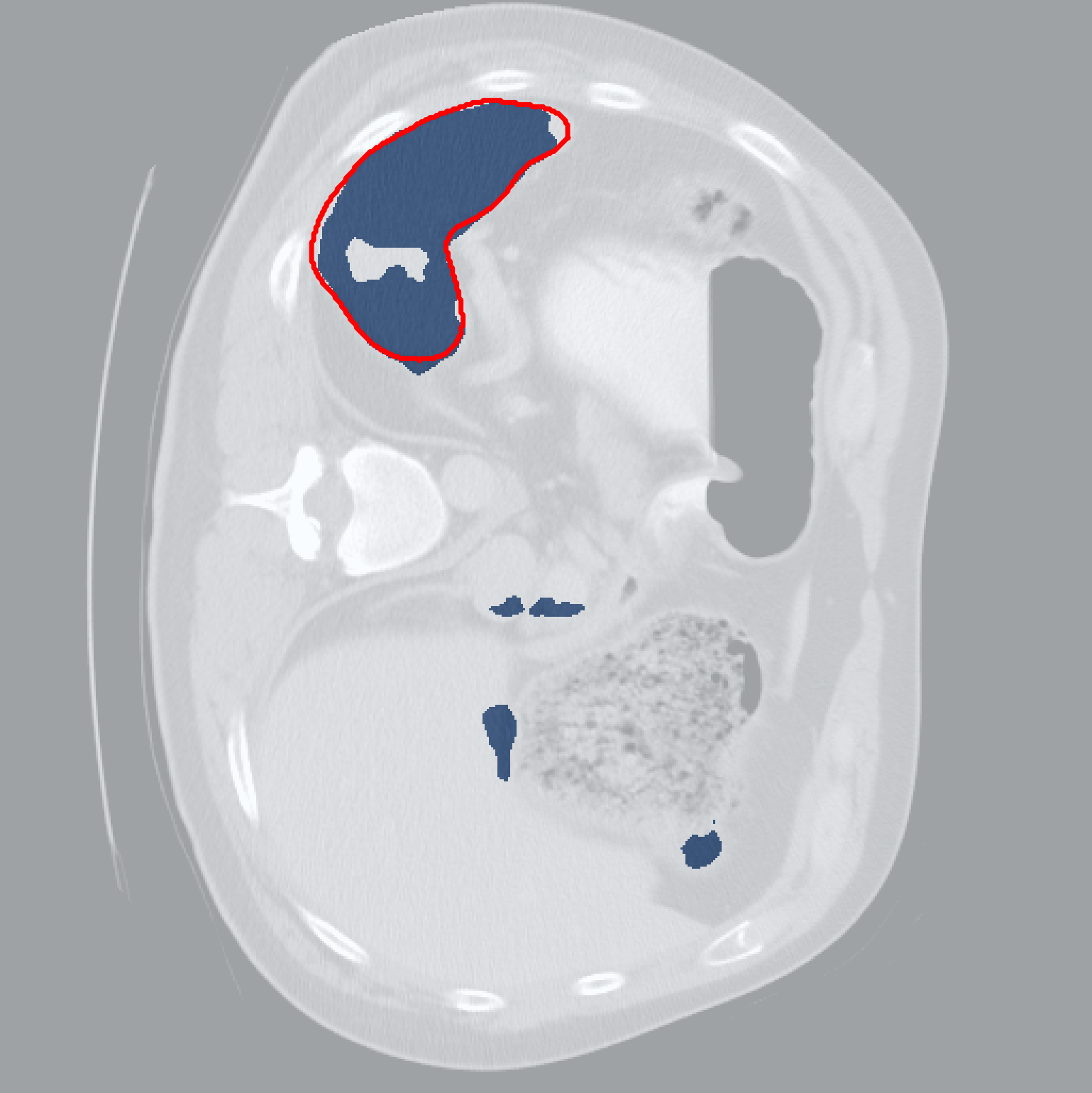} &
    \includegraphics[width=.135\linewidth]{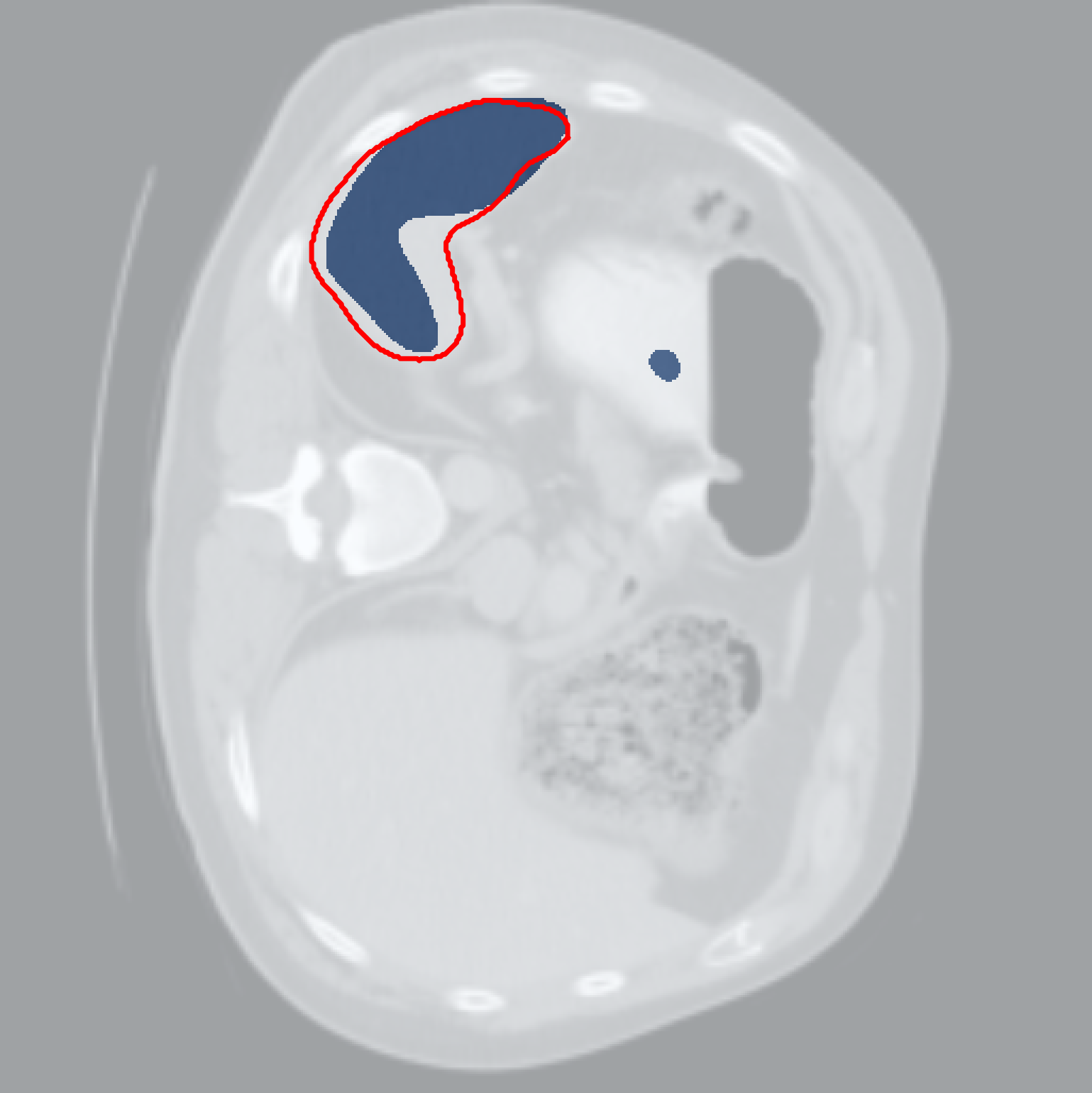} &
    \includegraphics[width=.135\linewidth]{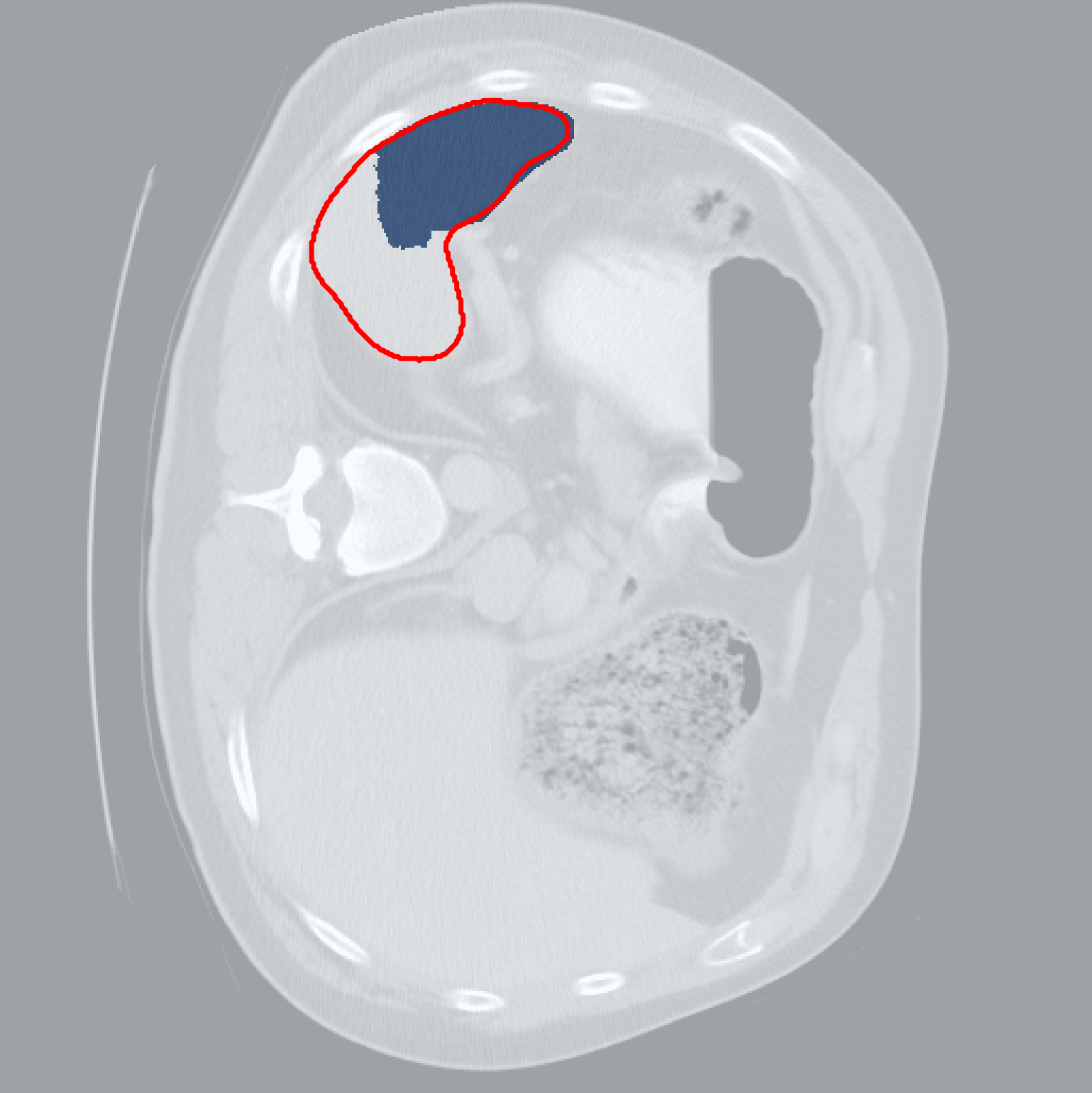} & 
    \includegraphics[width=.135\linewidth]{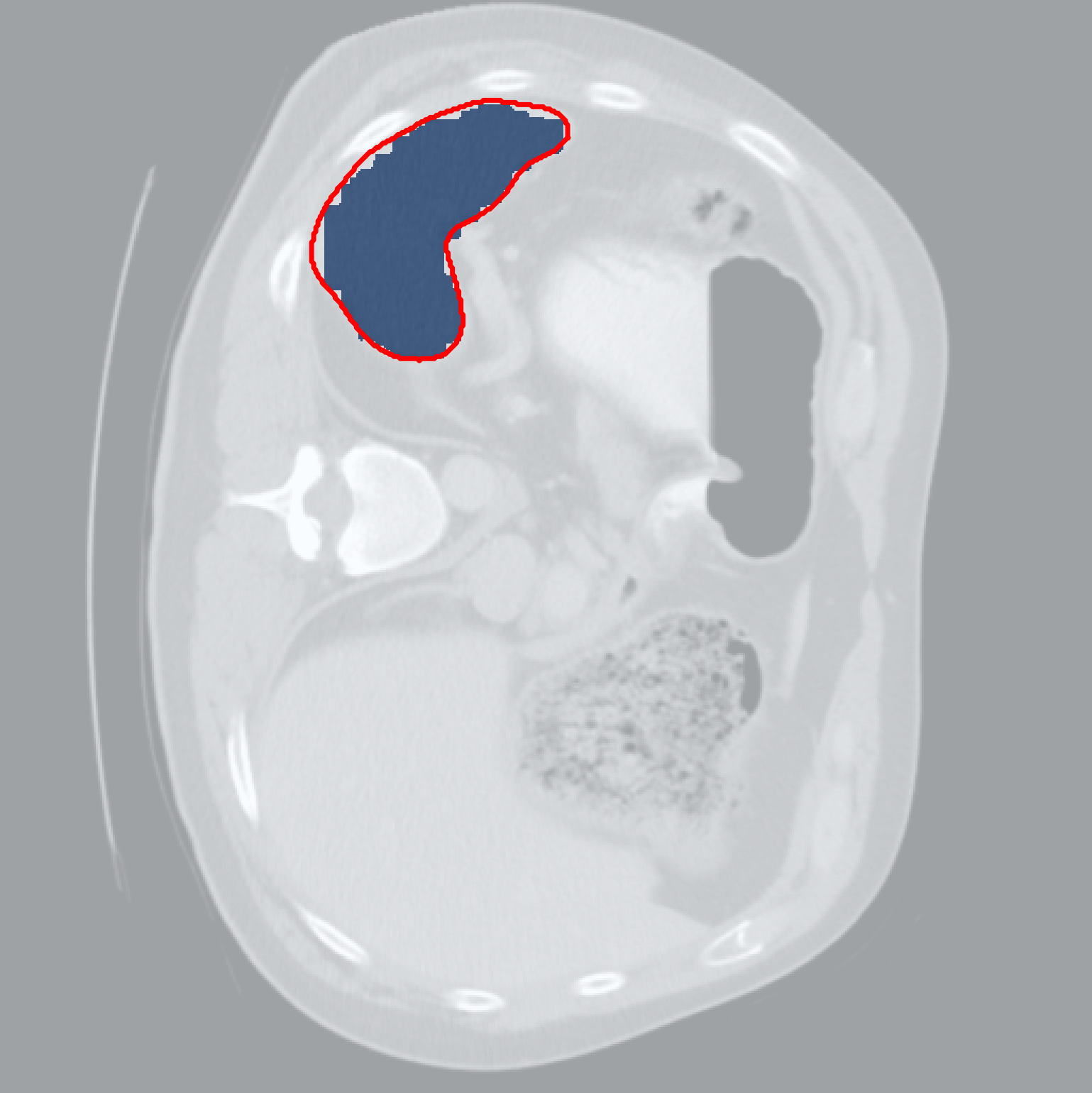} & 
    \includegraphics[width=.135\linewidth]{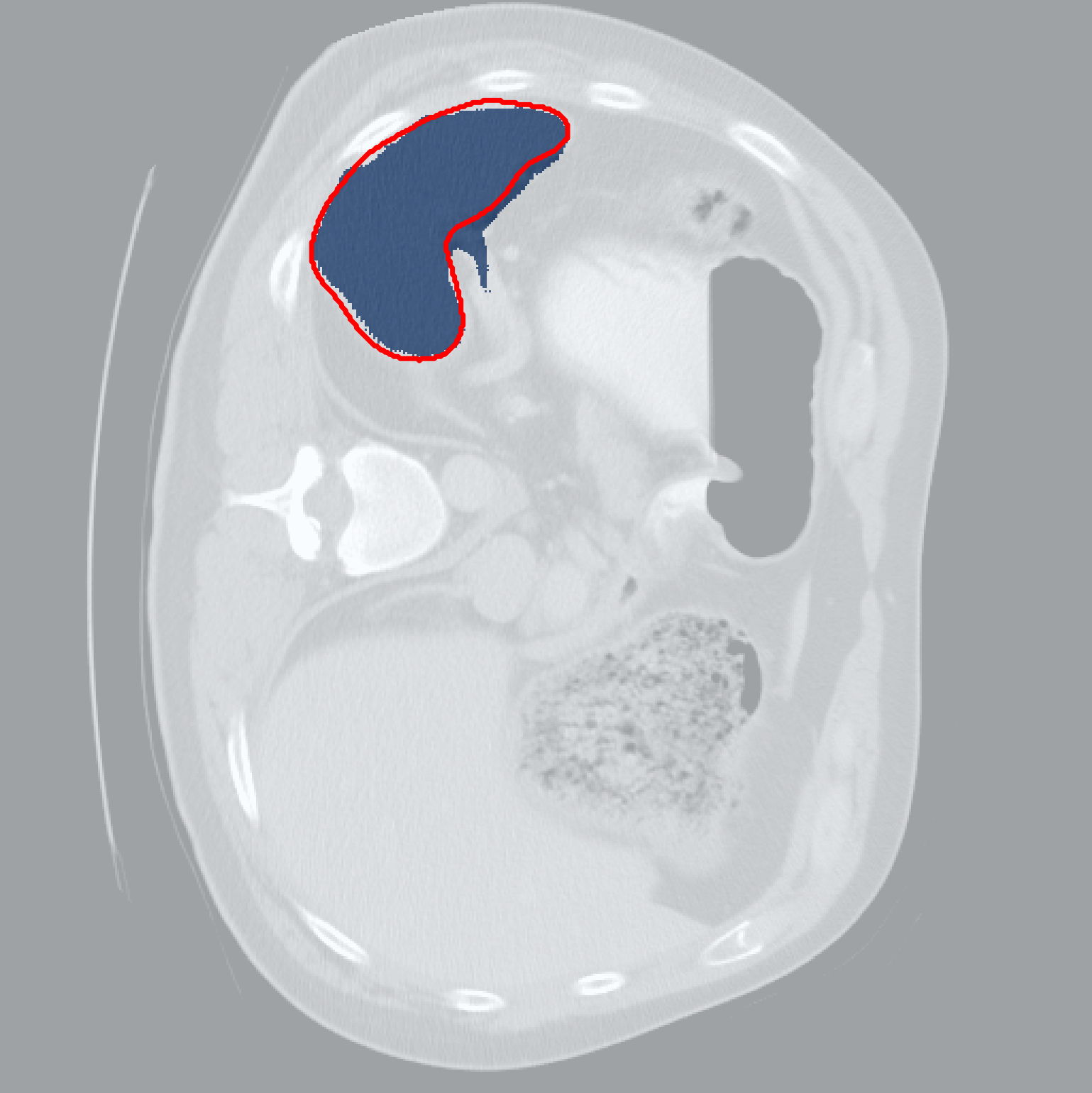} & 
    \includegraphics[width=.135\linewidth]{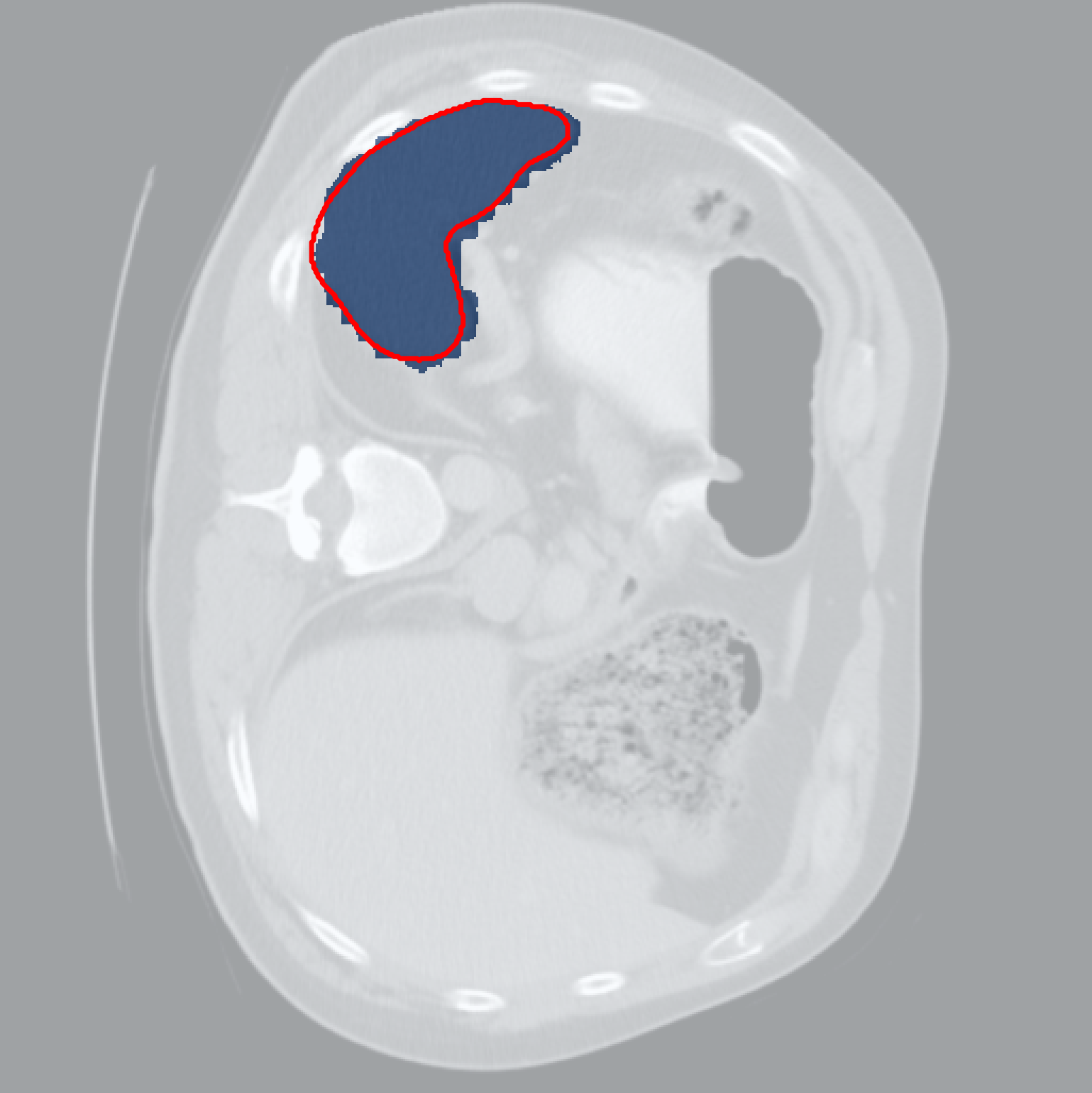} \\

    & Interactive & \multicolumn{4}{c}{\rule[2pt]{3.2cm}{.5pt} Fully superv.\rule[2pt]{3.2cm}{.5pt}} & \multicolumn{2}{c}{\rule[2pt]{1cm}{.5pt} Weakly superv.\rule[2pt]{1cm}{.5pt}}\\
     & & \multicolumn{4}{c}{GT mask} & \multicolumn{2}{c}{\centering{\textbf{Bounding box}}}\\[8pt]

        & 
      \makecell[c]{(a) SAM} &
      \makecell[c]{(b) UNet} & 
      \makecell[c]{(c) TransUNet} & 
      \makecell[c]{(d) Self-prompt.} & 
      \makecell[c]{(e) AutoSAM} & 
      \makecell[c]{(f) Kervadec et al.} & 
      \makecell[c]{(g) \textbf{Ours}}  \\
      
        & 
      \makecell[l]{\hspace{3.2em}\cite{kirillov_segment_2023}} &
      \makecell[l]{\hspace{3.2em}\cite{ronneberger_u-net_2015}} & 
      \makecell[l]{\hspace{2.3em}\cite{chen_transunet_2024}} & 
      \makecell[l]{\hspace{2em}\cite{wu_self-prompting_2023}} & 
      \makecell[l]{\hspace{2.2em}\cite{shaharabany_autosam_2023}} & 
      \makecell[l]{\hspace{1.9em}\cite{kervadec_bounding_2020} } & 
      \makecell[l]{}  
\end{tabular}
\caption{Predicted segmentations on test samples of the HC18, CAMUS, \revision{ACDC-LV and MSD-Spleen} datasets. From left to right, (a) SAM prompted with a tight box based on the ground truth, (b-e) UNet, TransUNet, Self-prompting and AutoSAM, all trained with ground-truth masks, and (f) residual UNet trained with tight box constraints and (g) our prompt learning method trained with bounding box annotations. All automatic methods are given for the 20-shot setting. Ground-truth annotation is drawn in red, with the predicted segmentation mask overlayed in blue. In most cases, our weakly supervised approach is able to produce better segmentations than methods that require full annotation masks during training.
}
    \label{fig:visual_results_20samples}
    \vspace*{-1em}
\end{figure*} 

We evaluate the efficacy of our proposed optimization scheme for prompt automation by comparing it to specialized models (trained exclusively for the target task) and existing SAM-based adaptations which require ground-truth labels, as well as a weakly supervised approach based on bounding boxes \cite{kervadec_bounding_2020,wang_bounding_2021}. 
To validate the robustness and generalizability of our approach, we conduct experiments on datasets of different imaging modalities and anatomical structures.
We then explore the effectiveness of our method through ablation studies on \revision{the impact of trainings set size, individual loss components, backbone foundation model and label noise}.

\subsection{Datasets}\label{subsec:datasets}
To ensure a comprehensive evaluation, we tested our method on ultrasound (US), MRI and CT datasets. \revisiontwo{Five} publicly available medical imaging datasets were used: the ultrasound Head Circumference dataset (HC18) \cite{heuvel_automated_2018}, the Cardiac Acquisitions for Multi-structure Ultrasound Segmentation (CAMUS) \cite{leclerc_deep_2019}, the MRI Automated Cardiac Diagnosis Challenge (ACDC) \cite{bernard_deep_2018} and the spleen \revisiontwo{and liver CT datasets} from the Medical Segmentation Decathlon (MSD) \cite{antonelli_medical_2022}. Tasks included segmentation of the head circumference, end-diastole of the left (LV) and right (RV) ventricles, as well as spleen \revisiontwo{and liver. For uniformity across all datasets, we conducted slice-based segmentation of imaging volumes and filtered out background-only slices, keeping the largest object of size at least 10 pixels.}

For HC18, we used 507 images for training, 77 for validation and 148 for testing. For CAMUS, we focused on left ventricle (LV) segmentation and kept 350 training, 50 validation and 100 test images. For ACDC, we used 90, 10 and 50 patients (765, 78 and 470 images) for training, validation and testing. For the MSD-Spleen dataset, we used 4 patients (114 images) for validation, 10 patients (340 images) for testing and the remaining 27 patients (597 images) for training. \revisiontwo{Finally, for the MSD-Liver dataset containing 131 patients, we utilized 13 patients (1,195 images) for validation and 30 patients (3,466 images) for testing.}
Following \cite{ma_segment_2024}, our preprocessing involved clipping the intensity values of each 2D image (HC18, CAMUS) or each 3D volume (ACDC, MSD) to the 0.5th and 99.5th percentiles, followed by rescaling to the range [0, 255]. Each 3D volume from the ACDC and MSD datasets was partitioned into 2D images and resampled to a fixed resolution of 1mm x 1mm. We then center-cropped and padded each sample to size 640$\times$640 (HC18), 512$\times$512 (CAMUS and MSD-Spleen) or 256$\times$256 (ACDC \revisiontwo{and MSD-Liver}). Finally, to comply with SAM's requirements, all images were resized to a fixed dimension of 3$\times$1024$\times$1024 before being fed to the foundation model.

\subsection{Implementation Details}
Our backbone promptable foundation model used SAM ViT-H, the largest variant of SAM, which was kept frozen.
Training of the prompt module operated on 20 samples for 200 epochs using a batch size of 4. The learning rate was set to 0.0001 and reduced by a factor of 0.1 midway through the training process. Additionally, a weight decay of 0.0001 was applied.
Our training process optimized the total loss, $\loss_{\text{total}}$, composed of four distinct loss terms, each assigned a specific weight. For all experiments, we used the following weight values: $\lambda_1\!=\!1$, $\lambda_2\!=\!0.01$, $\lambda_3\!=\!0.001$, and $\lambda_4\!=\!0.001$. 
\revision{These hyperparameters were tuned by optimizing for the best Dice score on the validation set, based on bounding boxes generated from the provided weak annotation and predicted mask. To avoid selecting hyperparameters encouraging hollow segmentation masks, we required the average predicted foreground-to-bounding-box size ratio to be above $50\%$. Optimization was performed on the ACDC dataset, and the same hyperparameters were then kept fixed throughout the experiments.}
To impose a strong size constraint, we set $[\epsilon_1, \epsilon_2] = [0.7, 0.9]$ and, following \cite{kervadec_bounding_2020}, scaled the parameter $t$ of the log-barrier function $\psi_t$ by a factor $1.1$ every 5 epochs.
For the consistency loss, transformations included random flips and rotations. Due to the symmetrical nature of images in the MSD-Spleen dataset, we replaced flips with random translations and scaling to better align with the dataset's characteristics. We used the model from the final training epoch at test time. 

To minimize computational complexity and speed-up training, we did not perform data augmentation. Doing so allowed us to discard the image encoder during training by using pre-computed image embeddings, reducing the number of model parameters from 141.8M to 45.6M, with 41.6M trainable parameters.

Each experiment was repeated using three randomly selected training subsets and three different initialization seeds to ensure robustness. The results were averaged across all 9 trials. All experiments were conducted using Python 3.8.10 with PyTorch on NVIDIA RTX A6000 GPUs.

\subsection{Evaluation Protocol}

\subsubsection{Evaluation Measures}
\revision{Two evaluation measures assess the performance of our proposed approach: the Dice Similarity Coefficient (DSC) and the Average Symmetric Surface Distance (ASSD).}
The DSC is defined as follows:
\begin{equation}
    DSC = \frac{2 \cdot |A \cap B|}{|A| + |B|},
\end{equation}
where \( A \) and \( B \) represent the sets of predicted and ground truth segmentation masks. The DSC quantifies the overlap between the predicted and ground truth masks, with values ranging from 0\% (no overlap) to 100\% (perfect overlap). 

\revision{The Average Symmetric Surface Distance measures the average shortest distances between contour $C_A$ to any point on contour $C_B$, and vice-versa:
\begin{equation}
    ASSD(C_A, C_B) = \frac{\sum_{a \in C_A} d(a, C_B) + \sum_{b \in C_B} d(b, C_A)}{|C_A| + |C_B|},
\end{equation}
with $d(i, C_J) = \min_{j \in C_J} d(i, j)$. The ASSD being undefined for empty ground truths or predictions, we set in such case the distance $d(i, C_j) = \sqrt{H^2 + W^2}$, corresponding to the maximum possible distance in the image.
}

\subsubsection{Baselines and Comparative Methods}
We compare our proposed method to two specialized models: a standard UNet \cite{ronneberger_u-net_2015} and TransUNet \cite{chen_transunet_2024}. Additionally, we validate our approach against the original SAM prompted with a tight bounding box based on the ground truth mask, as well as three SAM-based automated approaches: Self-prompting \cite{wu_self-prompting_2023}, AutoSAM \cite{shaharabany_autosam_2023}, and PerSAM \cite{zhang_personalize_2024}. We also do a comparison with \revision{two weakly-supervised methods based on a residual UNet trained with bounding box annotations: the first using box-based constraints \cite{kervadec_bounding_2020} and the other based on generalized MIL and smooth maximum approximation \cite{wang_bounding_2021}.}

The UNet, TransUNet, Self-prompting, PerSAM and AutoSAM models were trained using full segmentation masks. To ensure optimal performance for the baseline models, we increased the batch size to 24 for TransUNet, following \cite{chen_transunet_2024}. The UNet and TransUNet were optimized with a standard Dice cross-entropy loss. The weakly supervised and SAM-based prompt learning baselines followed the best practices outlined in \cite{kervadec_bounding_2020,wang_bounding_2021,shaharabany_autosam_2023,wu_self-prompting_2023,zhang_personalize_2024}. 

\subsection{Quantitative and Qualitative Results}

\begin{table*}[htb!]
\centering
\caption{Model performance on test sets in terms of mean ($\pm$std) \revision{2D DSC and ASSD}, with limited training set size (20 samples except for PerSAM, which uses 1 sample). The first row gives the results of SAM when prompted with a tight bounding box. The best results for weakly supervised approaches are shown in bold while the best fully-supervised results are underlined. \revision{* indicates statistical significance with a p-value $<0.05$ for all paired permutation tests between our method and each baseline.}
}
\label{tab:results_20samples}
\setlength{\tabcolsep}{2pt}
\resizebox{\linewidth}{!}
{
\begin{tabular}{clc|cc|cc|cc|cc|cc|cc}
\toprule
\multirow{3}{*}[-2pt]{Train Label} & \multirow{3}{*}[-2pt]{Method} &  \multirow{3}{*}{\parbox{1cm}{\revision{\#Train. Params}}} & \multicolumn{4}{c}{\revision{Ultrasound (US)}} & \multicolumn{4}{c}{\revision{MRI}} & \multicolumn{4}{c}{\revision{CT}} \\
\cmidrule(l{4pt}r{4pt}){4-7}\cmidrule(l{4pt}r{4pt}){8-11}\cmidrule(l{4pt}r{4pt}){12-15} 
 &  &  & \multicolumn{2}{c}{HC18} & \multicolumn{2}{c}{CAMUS} & \multicolumn{2}{c}{ACDC-RV} & \multicolumn{2}{c}{ACDC-LV} & \multicolumn{2}{c}{MSD-Spleen} & \multicolumn{2}{c}{\revisiontwo{MSD-Liver}} \\ 
 \cmidrule(l{4pt}r{4pt}){4-5} \cmidrule(l{4pt}r{4pt}){6-7}\cmidrule(l{4pt}r{4pt}){8-9}\cmidrule(l{4pt}r{4pt}){10-11}\cmidrule(l{4pt}r{4pt}){12-15}
        &  &  & DSC ($\uparrow$) & \gray{ASSD ($\downarrow$)} & DSC ($\uparrow$) & \gray{ASSD ($\downarrow$)} & DSC ($\uparrow$) & \gray{ASSD ($\downarrow$)} & DSC ($\uparrow$) & \gray{ASSD ($\downarrow$)} & DSC ($\uparrow$) & \gray{ASSD ($\downarrow$)} & \revisiontwo{DSC ($\uparrow$)} & \graytwo{ASSD ($\downarrow$)}\\
\midrule\midrule
        - & \makecell[l]{Interactive SAM \cite{kirillov_segment_2023}} & \revision{-} & 94.18 & \gray{12.98} &  85.49  & \gray{13.24} & 90.64 & \gray{1.60} & 93.71 & \gray{1.34} & 92.82 & \gray{1.75} & \revisiontwo{93.40} & \graytwo{1.82} \\ 
\midrule\midrule
        \multirow{6}{*}{\parbox{2cm}{Fully superv.\\ (GT mask)}} & \multirow{1}{*}{UNet \cite{ronneberger_u-net_2015}} & \revision{6.8\,M} & 70.55\ppm2.35 & \gray{61.90\ppm2.74} & 72.97\ppm9.47 & \gray{27.30\ppm10.29} & 50.66\ppm4.31 & \gray{36.12\ppm5.15} & 67.67\ppm3.69 & \gray{29.31\ppm5.55} & 67.12\ppm9.49 & \gray{84.18\ppm31.07} & \revisiontwo{64.86\ppm4.78} & \graytwo{26.05\ppm5.54} \\
& \multirow{1}{*}{TransUNet \cite{chen_transunet_2024}} & \revision{105\,M} &  \underline{95.82}\ppm0.29 & \gray{\underline{8.33}\ppm0.97} &  88.64\ppm0.89 & \gray{15.55\ppm4.86} &  66.19\ppm3.70 & \gray{14.55\ppm2.07} &  83.48\ppm1.97 & \gray{9.40\ppm2.79} &  70.16\ppm2.28 & \gray{\underline{14.55}\ppm2.84} & \revisiontwo{75.67\ppm2.46} & \graytwo{11.02\ppm3.06}\\ 
   \cmidrule(l{2pt}r{2pt}){2-15}
          & \multirow{1}{*}{Self-prompting \cite{wu_self-prompting_2023}} & \revision{257} & 83.38\ppm1.18 & \gray{37.23\ppm2.73} & 74.04\ppm1.20 & \gray{25.70\ppm1.01} & 52.27\ppm3.65 & \gray{41.97\ppm4.94} &  63.67\ppm1.71 & \gray{18.47\ppm0.44} & 70.95\ppm0.96 & \gray{28.31\ppm5.17} & \revisiontwo{68.15\ppm3.55} & \graytwo{27.18\ppm3.82}\\ 
&  \multirow{1}{*}{AutoSAM \cite{shaharabany_autosam_2023}} & \revision{41.6\,M} & 92.14\ppm1.80 & \gray{16.50\ppm5.51} & \underline{88.78}\ppm1.84 & \gray{\underline{11.60}\ppm2.73} & \underline{69.01}\ppm7.02 & \gray{\underline{11.58}\ppm4.1} & \underline{87.10}\ppm1.82 & \gray{\underline{5.36}\ppm1.39} & \underline{82.30}\ppm4.01 & \gray{24.51\ppm11.13} & \revisiontwo{\underline{81.05}\ppm3.42} & \graytwo{\phantom{0}\underline{8.91}\ppm3.22}\\ 
           &  PerSAM \cite{zhang_personalize_2024} & \revision{-} & 58.98\ppm0.19 & \gray{106.44\ppm0.63} & 36.13\ppm0.00 & \gray{110.88\ppm0.01} & 27.64\ppm9.48 & \gray{57.71\ppm16.52} & 45.43\ppm5.47 & \gray{43.58\ppm3.21} & 12.84\ppm6.26 & \gray{143.20\ppm11.95} & \revisiontwo{23.40\ppm0.67} & \graytwo{65.52\ppm0.58}\\ 
&  PerSAM-f \cite{zhang_personalize_2024} & \revision{2} & 68.67\ppm4.32 & \gray{75.10\ppm2.78} & 48.84\ppm4.82 & \gray{72.36\ppm10.62} & 28.47\ppm12.42 & \gray{42.80\ppm14.26} & 61.13\ppm9.16 & \gray{21.83\ppm5.96} & 36.72\ppm17.62 & \gray{55.18\ppm21.57} & \revisiontwo{55.25\ppm6.31} & \graytwo{28.15\ppm6.82}\\ 
   \midrule \midrule
            \multirow{3}{*}{\parbox{2cm}{Weak. superv.\\ \textbf{(Bound. box)}}} & Kervadec et al. \cite{kervadec_bounding_2020} & \revision{18.7\,M} & 76.08\ppm5.42 & \gray{38.98\ppm3.53} & 79.56\ppm2.46 & \gray{15.52\ppm0.94} & 55.64\ppm4.49 & \gray{31.70\ppm7.65} & 72.43\ppm4.81 & \gray{27.84\ppm11.11} & 75.64\ppm3.82 & \gray{17.27\ppm3.69} & \revisiontwo{72.43\ppm4.87} & \graytwo{30.97\ppm14.24}\\ 
& Wang et al. \cite{wang_bounding_2021} & \revision{19.6\,M} & \revision{30.23\ppm5.65} & \gray{32.77\ppm5.27} & \revision{72.66\ppm7.52} & \gray{21.63\ppm4.19} & \revision{35.30\ppm5.13} & \gray{48.45\ppm25.83} & \revision{65.35\ppm5.66} & \gray{35.62\ppm13.63} & \revision{26.36\ppm26.46} & \gray{97.23\ppm6.53} & \revisiontwo{39.46\ppm23.67} & \graytwo{41.80\ppm19.21}\\ 
 \cmidrule(l{2pt}r{2pt}){2-15}
            \ccol & \ccol \textbf{Ours} & \revision{41.6\,M} & \ccol \textbf{92.25}\ppm0.84 & \gray{\textbf{18.75\ppm3.00}} & \ccol \textbf{84.21*}\ppm1.15 & \gray{\textbf{14.22*\ppm0.97}}  & \ccol \textbf{80.77*\ppm1.12} & \gray{\textbf{5.34*\ppm0.81}} & \ccol \textbf{89.82*\ppm1.00} & \gray{\textbf{3.40*\ppm0.69}} & \ccol \textbf{83.89*\ppm1.62} & \gray{\textbf{13.14*\ppm4.48}} & \revisiontwo{\textbf{78.47\ppm1.59}} & \graytwo{\textbf{10.37\ppm3.17}}\\ 
\bottomrule
\end{tabular}
}
\end{table*}

Our main findings, based on experiments conducted on five distinct medical imaging datasets—ultrasound (HC18 and CAMUS), MRI (ACDC), and CT (MSD-Spleen \revisiontwo{and MSD-Liver})—are summarized in Table \ref{tab:results_20samples} and visualized in Fig. \ref{fig:visual_results_20samples}.

From Table \ref{tab:results_20samples}, we observe that \revision{our prompt-learning method} outperforms by a large margin the other \revision{weakly supervised approaches \cite{kervadec_bounding_2020,wang_bounding_2021}}. Compared to fully-supervised methods, our approach outputs results that are on par with the best performing specialized method (TransUNet) and SAM-based method (AutoSAM). For \revisiontwo{half of the tasks}, our approach based on bounding boxes is even able to outperform the best fully-supervised approach requiring ground truth masks. These findings are supported visually by Fig.\ref{fig:visual_results_20samples}. Interestingly, the figure also shows that our method yields a better segmentation than interactive SAM on the CAMUS test sample, supporting our claim that our proposed approach is able to address the failed predictions of the foundation model.

\subsection{Ablation Study}

\subsubsection{Impact of Loss Components}

\begin{table*}[htb!]
\centering
\caption{\revision{Impact of each loss component on the 2D DSC ($\uparrow$). Results are reported for two settings: a highly complex setting with a small backbone and training set size (SAM ViT-b and 10 samples), and a moderately complex setting with a huge backbone and larger training set size (SAM ViT-H and 20 samples). The importance of each loss component is most apparent in the highly complex setting, where the foundation model is most likely to make errors.}}
\label{tab:ablation_loss_components}
\setlength{\tabcolsep}{4pt}
\resizebox{\textwidth}{!}{
\begin{tabular}{c cccc ccccc}
\toprule
\multirow[c]{2}{*}{\makecell[c]{\revision{Task} \\ \revision{complexity}}} & \multirow[c]{2}{*}{\makecell[c]{Pseudo-label \\loss}} & \multirow[c]{2}{*}{\makecell[c]{Size \\constraint}} & \multirow[c]{2}{*}{\makecell[c]{Emptiness \\constraint}} & \multirow[c]{2}{*}{\makecell[c]{Consistency \\loss}} & \multirow[c]{2}{*}{\makecell[c]{HC18}} &  \multirow[c]{2}{*}{\makecell[c]{CAMUS}} & \multirow[c]{2}{*}{\makecell[c]{ACDC-LV}} &  \multirow[c]{2}{*}{\makecell[c]{ACDC-RV}} & \multirow[c]{2}{*}{\makecell[c]{MSD-Spleen}}\\
\\
 \midrule
\multirow[c]{3}{*}{\revision{High}} & \checkmark &   &  &  &  38.43\ppm3.16 & 76.56\ppm5.96 & 62.87\ppm7.39 & 80.06\ppm1.94 & 81.49\ppm2.18\\
& \checkmark & \checkmark & \checkmark  &  & 77.38\ppm3.39 & 82.65\ppm1.18 & 65.46\ppm3.21 & 79.38\ppm3.87 & 81.19\ppm0.86\\
& \checkmark & \checkmark & \checkmark  &  \checkmark & \textbf{79.68}\ppm1.83 & \textbf{83.13}\ppm0.33 & \textbf{72.59}\ppm1.67 & \textbf{84.33}\ppm2.02 & \textbf{81.92}\ppm0.72\\
 \midrule
 \multirow[c]{3}{*}{\revision{Moderate}} & \checkmark &   &  &  &  \revision{90.15\ppm 1.05} & \revision{83.54\ppm1.14} & \revision{72.70\ppm4.95} & \revision{87.49\ppm0.94} & \revision{\textbf{86.19}\ppm2.40} \\
 & \checkmark & \checkmark & \checkmark  &  & \revision{90.46\ppm0.66} & \revision{83.28\ppm1.11} & \revision{70.18\ppm7.01} & \revision{87.44\ppm1.55} & \revision{84.10\ppm1.43}\\
 & \checkmark & \checkmark & \checkmark  &  \checkmark & \revision{\textbf{92.25}\ppm0.84} & \revision{\textbf{84.21}\ppm1.15} & \revision{\textbf{80.77}\ppm1.12} & \revision{\textbf{89.82}\ppm1.00} & \revision{83.89\ppm1.62}\\
\bottomrule
\end{tabular}
}
\end{table*}

In the absence of ground-truth masks, our objective function exploits the predictions of the prompted foundation model, box-based constraints, and consistency-based regularization. In this ablation study, we assessed the contribution of each loss component \revision{in both a highly complex setting (with SAM ViT-b and 10 samples) and a moderately complex
setting (with SAM ViT-H and 20 samples)}. 
\revision{Table \ref{tab:ablation_loss_components}, shows that, in highly complex settings where the foundation model is most likely to make errors even when provided with a prompt, our size and emptiness constraints are able to successfully guide the model to produce more accurate predictions. This is especially true for tasks where the pseudo-label loss alone falls short (i.e., ultrasound head segmentation). However, these constraints benefit all tested datasets. Our consistency loss can further boost performance by up to 10.9\% by regularizing the prompt module's training. In a moderately complex setting, where the foundation model is more informative and the module is trained on a bigger set, using only our pseudo-label loss already provides reasonable segmentation masks. Additional constraints and consistency loss can nonetheless further improve the performance.} \revisiontwo{Similarly, Table \ref{tab:ablation_WSL} validates the strength of our proposed weakly supervised optimization scheme. Using our proposed losses yields up to 32\% improvement compared to when using the tightness constraints of \cite{kervadec_bounding_2020} to train our prompt module.}

\begin{table}[htb!]
\centering
\caption{\revisiontwo{Impact of weakly supervised optimization scheme on our prompt learning module training. The 2D test DSC ($\uparrow$) is reported after training with 20 samples. Our box-based optimization strategy significantly outperforms existing weakly supervised loss functions.}}
\label{tab:ablation_WSL}
\setlength{\tabcolsep}{4pt}
\resizebox{0.48\textwidth}{!}{
\begin{tabular}{c ccc}
\toprule
\multirow[c]{1}{*}{\makecell[c]{\revisiontwo{Weakly Supervised Loss}}} & \multirow[c]{1}{*}{\makecell[c]{\revisiontwo{HC18}}} &  \multirow[c]{1}{*}{\makecell[c]{\revisiontwo{ACDC-LV}}} & \multirow[c]{1}{*}{\makecell[c]{\revisiontwo{MSD-Spleen}}}\\
 \midrule
\revisiontwo{Tightness constraints \cite{kervadec_bounding_2020}} & \revisiontwo{85.68\ppm2.54} & \revisiontwo{61.09\ppm83.91} & \revisiontwo{73.09\ppm1.46}\\
\revisiontwo{Ours} & \revisiontwo{\textbf{92.25}\ppm0.84} & \revisiontwo{\textbf{80.77}\ppm1.12} & \revisiontwo{\textbf{83.89\ppm1.62}}\\
\bottomrule
\end{tabular}
}
\end{table}

\subsubsection{Impact of Number of Training Samples}

We evaluated the limits of our approach by experimenting with a smaller and larger training set, respectively 10 samples and the complete training set. We focused on the HC18, ACDC-LV and MSD-Spleen datasets, each representing a different modality: ultrasound, MRI and CT. We kept the same setup as with the main experiments. However, in the full data setting, we reduced the number of epochs to 20, and given the low-intensity contrast typical of ultrasound images, we tightened the size constraint on the HC18 datasets by doubling $t$ every epoch.

\begin{figure}[h!]
\centering
\setlength{\tabcolsep}{0.5pt}
\begin{tabular}{c}
    \includegraphics[width=0.7\linewidth]{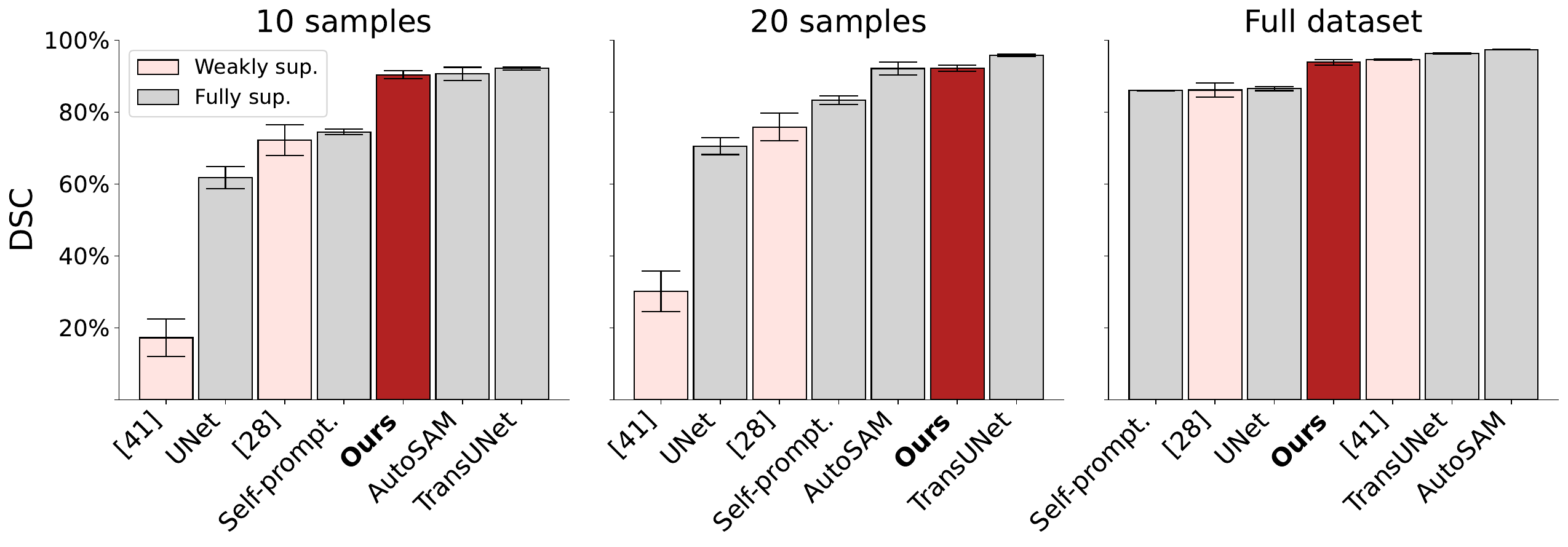}
\end{tabular}
\caption{
\revision{Results on HC18 with 10, 20 and all samples. Our weakly supervised method (dark red) consistently ranks in the top-3 approaches. In low data regimes, it outperforms other bounding box-based (pink) and fully supervised (grey) approaches by a large margin.}} 
\label{fig:results_10_20_all_samples_HC}
\end{figure}

\begin{figure}[h!]
\centering
\setlength{\tabcolsep}{0.5pt}
\begin{tabular}{c}
   \includegraphics[width=0.7\linewidth]{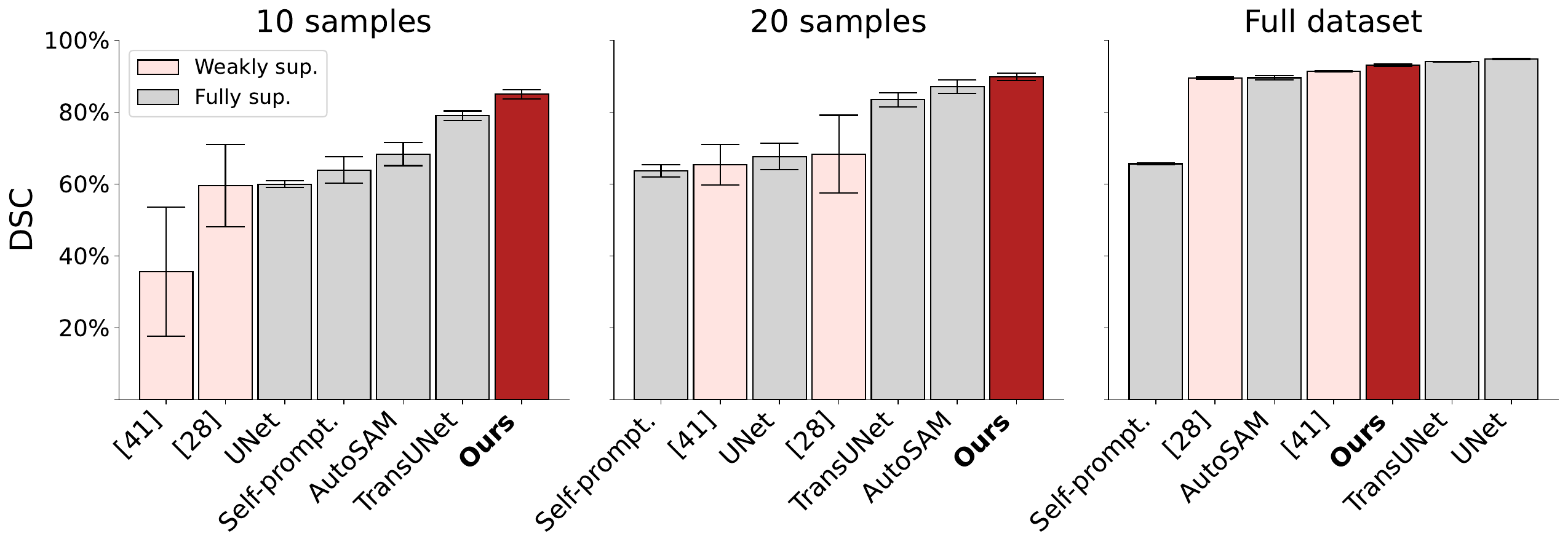}
\end{tabular}
\caption{\revision{Results on ACDC-LV for increasing training set sizes. Given only limited data, our approach (dark red) outperforms all other methods—including fully supervised methods with GT masks (grey).}} 
\label{fig:results_10_20_all_samples_ACDC1}
\end{figure}

\begin{figure}[h!]
\centering
\setlength{\tabcolsep}{0.5pt}
\begin{tabular}{c}
    \includegraphics[width=0.7\linewidth]{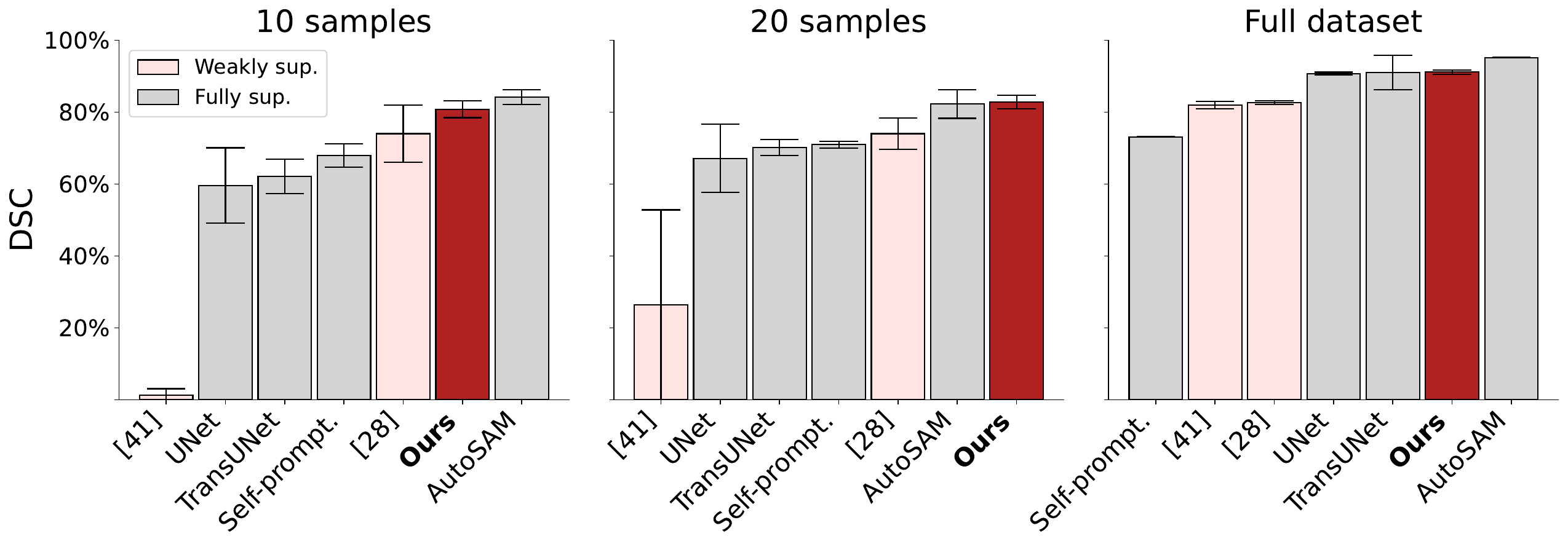} 
\end{tabular}
\caption{\revision{Results on MSD-Spleen with 10, 20 and all samples. Our method (dark red) constantly ranks top-2, surpassing all box-based (pink) and nearly all mask-based (grey) approaches.}} 
\label{fig:results_10_20_all_samples_MSDSpleen}
\end{figure}

From Fig.\ref{fig:results_10_20_all_samples_HC},  Fig.\ref{fig:results_10_20_all_samples_ACDC1} and  Fig.\ref{fig:results_10_20_all_samples_MSDSpleen}, we observe that our method always performs better than \revision{other weakly supervised approaches \cite{kervadec_bounding_2020,wang_bounding_2021}}, across different training set sizes. In the very low data setting (10 samples), our method remains competitive, outperforming \revision{by a large margin} UNet and Self-prompting, and even surpassing TransUNet on two out of three tasks. Our prompt learning approach delivers results that are comparable to—and in one case, better than—AutoSAM, without requiring GT masks. 
These results validate the robustness of our approach, in scenarios with both very limited and full training data.

\subsubsection{Impact of Foundation Model Backbone}

To assess the generalizability of our framework, we perform evaluations with different vision foundation models, specifically MedSAM \cite{ma_segment_2024}, a specialized version of SAM fine-tuned for medical imaging tasks, as well as another versions of SAM (SAM ViT-b). The results presented in Table \ref{tab:ablation_backbone} for the HC18 and ACDC datasets, none of which were used to train MedSAM, demonstrate that our method performs effectively across different foundation model backbones. Notably, using a backbone model tailored explicitly for medical image analysis improves overall performance compared to the model trained on natural images, as evidenced by the higher Dice similarity scores across all tasks. The mean Dice score increases from 79.68\% to 91.17\% with the HC18 dataset, and similar performance gains are observed for both right and left ventricle segmentation of the ACDC dataset. These results reinforce the advantage of leveraging domain-specific foundation models and confirm the robustness of our approach to different backbone models.

\begin{table}[htb!]
\centering
\caption{\revision{2D DSC ($\uparrow$) on the test set with different backbone foundation models, when trained with 10 samples.}}
\label{tab:ablation_backbone}
\setlength{\tabcolsep}{4pt}
\resizebox{0.38\textwidth}{!}{
\begin{tabular}{lccc}
\toprule
\multirow{1}{*}{Backbone} &  \multicolumn{1}{c}{HC18} &  \multicolumn{1}{c}{ACDC-RV} & \multicolumn{1}{c}{ACDC-LV}\\
\midrule
SAM ViT-b & 79.68\ppm1.83 & 72.59\ppm1.67 & 84.33\ppm2.02\\
SAM ViT-H &  90.40\ppm1.09 & 72.43\ppm3.84 & 84.97\ppm1.34\\
MedSAM &  \textbf{91.17}\ppm1.05 & \textbf{74.52}\ppm3.12 &  \textbf{86.14}\ppm1.30 \\
\bottomrule
\end{tabular}
}
\end{table}

\subsubsection{Impact of Label Noise}

\revision{Finally, we explored a challenging real-case scenario, where bounding boxes are subject to human error. We simulated label noise by randomly displacing in any direction the box boundaries by up to 1.5\%, 1.5-3\% and 3-5\% of the total number of image pixels. \revisiontwo{Visual examples of such noisy prompts are shown in Fig.\ref{fig:PromptAmbiguity}.} The results for three datasets with different tasks, modalities and image sizes are provided in Table \ref{tab:ablation_noise}. Despite an expected decrease in performance with increasing variability of the bounding box sizes, our prompt module maintains a competitive performance. For instance, given light human error (less than 1.5\% of pixel displacement), our approach only shows a decrease of 0.5-0.7\% in the Dice similarity score for HC18 and ACDC-LV.}

\begin{table}[htb!]
\centering
\caption{\revision{Mean 2D DSC ($\uparrow$) for different noise levels of the bounding box annotation (in \% of total number of image pixels), when trained with 20 samples.}}
\label{tab:ablation_noise}
\setlength{\tabcolsep}{4pt}
\resizebox{0.48\textwidth}{!}{
\begin{tabular}{lccc}
\toprule
\multirow{1}{*}{\revision{Label pixel displacement}} &  \multicolumn{1}{c}{\revision{HC18}} & \multicolumn{1}{c}{\revision{ACDC-LV}} & \multicolumn{1}{c}{\revision{MSD-Spleen}}\\
\midrule
\revision{None (tight box)} & \revision{92.25\ppm0.84} & \revision{89.82\ppm1.00} & \revision{82.82\ppm1.90}\\
\revision{$<1.5\%$} & \revision{91.75\ppm0.22} & \revision{89.20\ppm0.07	} & \revision{80.33\ppm0.77}\\
\revision{$1.5 - 3\%$} & \revision{89.87\ppm0.79} & \revision{77.45\ppm2.14} & \revision{69.89\ppm1.54}\\
\revision{$3 - 5\%$} & \revision{84.99\ppm2.42} & \revision{68.45\ppm0.83} & \revision{58.03\ppm1.36}\\
\bottomrule
\end{tabular}}
\end{table}

\section{Conclusion}
Visual foundation models have \revision{enabled significant progress} in medical image segmentation \revision{by reducing the burden of manual annotation}. \revision{Recent prompt learning strategies automate these interactive models by training auxiliary modules to generate} prompts directly from images. However, \revision{their dependence} on pixel-wise annotated datasets remains a \revision{major limitation}.  
\revision{In this work, we propose a novel framework that combines the strengths of foundation models with the cost-efficiency of weakly supervised learning.} \revision{Our approach automates and adapts foundation models through a dedicated prompt module using only bounding box annotations. The module is trained via a} multi-loss optimization \revision{scheme that integrates the segmentation} predictions \revision{from} the prompted foundation model with box-based \revision{spatial} constraints and consistency regularization. \revision{Our method not only reduces annotation costs but also improves segmentation performance compared to existing weakly supervised approaches}. \revision{Through extensive} experiments across multi-modal datasets\revision{—spanning full-data, limited-data and out-of-domain settings—we} demonstrate the generalizability and robustness of our \revision{method}. \revision{Although our current implementation focuses on SAM-based models, our proposed framework is readily} extendable to other interactive foundation models, and \revision{provides a promising direction for future work in multi-class and multi-organ segmentation tasks.}

\printbibliography

@article{antonelli_medical_2022,
	title = {The {Medical} {Segmentation} {Decathlon}},
	volume = {13},
	copyright = {2022 The Author(s)},
	issn = {2041-1723},
	url = {https://www.nature.com/articles/s41467-022-30695-9},
	doi = {10.1038/s41467-022-30695-9},
	number = {1},
	journal = {Nature Communications},
	author = {Antonelli, Michela and Reinke, Annika and Bakas, Spyridon and Farahani, Keyvan and Kopp-Schneider, Annette and Landman, Bennett A. and Litjens, Geert and Menze, Bjoern and Ronneberger, Olaf and Summers, Ronald M. and van Ginneken, Bram and Bilello, Michel and Bilic, Patrick and Christ, Patrick F. and Do, Richard K. G. and Gollub, Marc J. and Heckers, Stephan H. and Huisman, Henkjan and Jarnagin, William R. and McHugo, Maureen K. and Napel, Sandy and Pernicka, Jennifer S. Golia and Rhode, Kawal and Tobon-Gomez, Catalina and Vorontsov, Eugene and Meakin, James A. and Ourselin, Sebastien and Wiesenfarth, Manuel and Arbeláez, Pablo and Bae, Byeonguk and Chen, Sihong and Daza, Laura and Feng, Jianjiang and He, Baochun and Isensee, Fabian and Ji, Yuanfeng and Jia, Fucang and Kim, Ildoo and Maier-Hein, Klaus and Merhof, Dorit and Pai, Akshay and Park, Beomhee and Perslev, Mathias and Rezaiifar, Ramin and Rippel, Oliver and Sarasua, Ignacio and Shen, Wei and Son, Jaemin and Wachinger, Christian and Wang, Liansheng and Wang, Yan and Xia, Yingda and Xu, Daguang and Xu, Zhanwei and Zheng, Yefeng and Simpson, Amber L. and Maier-Hein, Lena and Cardoso, M. Jorge},
	year = {2022},
	pages = {4128},
}

@inproceedings{avidan_visual_2022,
	address = {Cham},
	series = {Lecture {Notes} in {Computer} {Science}},
	title = {Visual {Prompt} {Tuning}},
	volume = {13693},
	url = {https://link.springer.com/10.1007/978-3-031-19827-4_41},
	urldate = {2023-09-08},
	booktitle = {ECCV},
	publisher = {Springer Nature Switzerland},
	author = {Jia, Menglin and Tang, Luming and Chen, Bor-Chun and Cardie, Claire and Belongie, Serge and Hariharan, Bharath and Lim, Ser-Nam},
	editor = {Avidan, Shai and Brostow, Gabriel and Cissé, Moustapha and Farinella, Giovanni Maria and Hassner, Tal},
	year = {2022},
	doi = {10.1007/978-3-031-19827-4_41},
	pages = {709--727},
}

@misc{ayzenberg_protosam_2024,
	title = {{ProtoSAM}: {One}-{Shot} {Medical} {Image} {Segmentation} {With} {Foundational} {Models}},
	shorttitle = {{ProtoSAM}},
	url = {http://arxiv.org/abs/2407.07042},
	publisher = {arXiv},
        number = {arXiv},
	author = {Ayzenberg, Lev and Giryes, Raja and Greenspan, Hayit},
	year = {2024}
}

@inproceedings{bearman_whats_2016,
	address = {Cham},
	title = {What’s the {Point}: {Semantic} {Segmentation} with {Point} {Supervision}},
	isbn = {978-3-319-46478-7},
	shorttitle = {What’s the {Point}},
	doi = {10.1007/978-3-319-46478-7_34},
	booktitle = {{ECCV}},
	publisher = {Springer International Publishing},
	author = {Bearman, Amy and Russakovsky, Olga and Ferrari, Vittorio and Fei-Fei, Li},
	editor = {Leibe, Bastian and Matas, Jiri and Sebe, Nicu and Welling, Max},
	year = {2016},
	pages = {549--565},
}

@article{bernard_deep_2018,
	title = {Deep {Learning} {Techniques} for {Automatic} {MRI} {Cardiac} {Multi}-{Structures} {Segmentation} and {Diagnosis}: {Is} the {Problem} {Solved}?},
	volume = {37},
	issn = {1558-254X},
	shorttitle = {Deep {Learning} {Techniques} for {Automatic} {MRI} {Cardiac} {Multi}-{Structures} {Segmentation} and {Diagnosis}},
	doi = {10.1109/TMI.2018.2837502},
	number = {11},
	journal = {IEEE Trans. Med. Imaging},
	author = {Bernard, O. and Lalande, A. and Zotti, C. and Cervenansky, F. and Yang, X. and Heng, P.-A. and Cetin, I. and Lekadir, K. and Camara, O. and Ballester, M. A. Gonzalez and Sanroma, G. and Napel, S. and Petersen, S. and Tziritas, G. and Grinias, E. and Khened, M. and Kollerathu, V. A. and Krishnamurthi, G. and Rohé, M.-M. and Pennec, X. and Sermesant, M. and Isensee, F. and Jäger, P. and Maier-Hein, K. H. and Full, P. M. and Wolf, I. and Engelhardt, S. and Baumgartner, C. F. and Koch, L. M. and Wolterink, J. M. and Išgum, I. and Jang, Y. and Hong, Y. and Patravali, J. and Jain, S. and Humbert, O. and Jodoin, P.-M.},
	month = nov,
	year = {2018},
	pages = {2514--2525},
}

@article{boykov2001fast,
  title={Fast approximate energy minimization via graph cuts},
  author={Boykov, Yuri and Veksler, Olga and Zabih, Ramin},
  journal={IEEE Trans. Pattern Anal. Mach. Intell.},
  volume={23},
  number={11},
  pages={1222--1239},
  year={2001},
  publisher={IEEE}
}

@inproceedings{chao_hardnet_2019,
	address = {Seoul, Korea (South)},
	title = {{HarDNet}: {A} {Low} {Memory} {Traffic} {Network}},
	copyright = {https://ieeexplore.ieee.org/Xplorehelp/downloads/license-information/IEEE.html},
	isbn = {978-1-72814-803-8},
	shorttitle = {{HarDNet}},
	url = {https://ieeexplore.ieee.org/document/9010717/},
	doi = {10.1109/ICCV.2019.00365},
	urldate = {2024-05-28},
	booktitle = {ICCV},
	publisher = {IEEE},
	author = {Chao, Ping and Kao, Chao-Yang and Ruan, Yushan and Huang, Chien-Hsiang and Lin, Youn-Long},
	month = oct,
	year = {2019},
	pages = {3551--3560},
}

@article{chen_transunet_2024,
	title = {{TransUNet}: {Rethinking} the {U}-{Net} architecture design for medical image segmentation through the lens of transformers},
	volume = {97},
	issn = {1361-8415},
	shorttitle = {{TransUNet}},
	url = {https://www.sciencedirect.com/science/article/pii/S1361841524002056},
	doi = {10.1016/j.media.2024.103280},
	journal = {Med. Image Anal.},
	author = {Chen, Jieneng and Mei, Jieru and Li, Xianhang and Lu, Yongyi and Yu, Qihang and Wei, Qingyue and Luo, Xiangde and Xie, Yutong and Adeli, Ehsan and Wang, Yan and Lungren, Matthew P. and Zhang, Shaoting and Xing, Lei and Lu, Le and Yuille, Alan and Zhou, Yuyin},
	year = {2024},
	pages = {103280},
}

@article{chen_rsprompter_2024,
	title = {{RSPrompter}: {Learning} to {Prompt} for {Remote} {Sensing} {Instance} {Segmentation} {Based} on {Visual} {Foundation} {Model}},
	volume = {62},
	copyright = {https://ieeexplore.ieee.org/Xplorehelp/downloads/license-information/IEEE.html},
	issn = {0196-2892, 1558-0644},
	shorttitle = {{RSPrompter}},
	url = {https://ieeexplore.ieee.org/document/10409216/},
	doi = {10.1109/TGRS.2024.3356074},
	urldate = {2024-06-19},
	journal = {IEEE Trans. Geosci. Remote Sens.},
	author = {Chen, Keyan and Liu, Chenyang and Chen, Hao and Zhang, Haotian and Li, Wenyuan and Zou, Zhengxia and Shi, Zhenwei},
	year = {2024},
	pages = {1--17},
}

@misc{cheng_sam-med2d_2023,
	title = {{SAM}-{Med2D}},
	url = {http://arxiv.org/abs/2308.16184},
	doi = {10.48550/arXiv.2308.16184},
	urldate = {2023-09-18},
	publisher = {arXiv},
        number = {arXiv},
	author = {Cheng, Junlong and Ye, Jin and Deng, Zhongying and Chen, Jianpin and Li, Tianbin and Wang, Haoyu and Su, Yanzhou and Huang, Ziyan and Chen, Jilong and Jiang, Lei and Sun, Hui and He, Junjun and Zhang, Shaoting and Zhu, Min and Qiao, Yu},
	month = aug,
	year = {2023},
}

@inproceedings{dai_boxsup_2015,
	title = {{BoxSup}: {Exploiting} {Bounding} {Boxes} to {Supervise} {Convolutional} {Networks} for {Semantic} {Segmentation}},
	shorttitle = {{BoxSup}},
	url = {https://ieeexplore.ieee.org/document/7410548},
	doi = {10.1109/ICCV.2015.191},
	urldate = {2024-06-30},
	booktitle = {ICCV},
	author = {Dai, Jifeng and He, Kaiming and Sun, Jian},
	month = dec,
	year = {2015},
	pages = {1635--1643},
}

@inproceedings{dosovitskiy_image_2021,
	title = {An {Image} is {Worth} 16x16 {Words}: {Transformers} for {Image} {Recognition} at {Scale}},
	booktitle = {ICLR},
	author = {Dosovitskiy, Alexey and Beyer, Lucas and Kolesnikov, Alexander and Weissenborn, Dirk and Zhai, Xiaohua and Unterthiner, Thomas and Dehghani, Mostafa and Minderer, Matthias and Heigold, Georg and Gelly, Sylvain and Uszkoreit, Jakob and Houlsby, Neil},
	year = {2021},
}

@inproceedings{gaillochet_automating_2024,
	address = {Cham},
	title = {Automating {MedSAM} by {Learning} {Prompts} with {Weak} {Few}-{Shot} {Supervision}},
	isbn = {978-3-031-73471-7},
	doi = {10.1007/978-3-031-73471-7_7},
	booktitle = {MedAGI},
	publisher = {Springer Nature Switzerland},
	author = {Gaillochet, Mélanie and Desrosiers, Christian and Lombaert, Hervé},
	editor = {Deng, Zhongying and Shen, Yiqing and Kim, Hyunwoo J. and Jeong, Won-Ki and Aviles-Rivero, Angelica I. and He, Junjun and Zhang, Shaoting},
	year = {2024},
	pages = {61--70},
}

@misc{gu_how_2024,
	title = {How to build the best medical image segmentation algorithm using foundation models: a comprehensive empirical study with {Segment} {Anything} {Model}},
	shorttitle = {How to build the best medical image segmentation algorithm using foundation models},
	url = {http://arxiv.org/abs/2404.09957},
	urldate = {2024-05-16},
	publisher = {arXiv},
        number = {arXiv},
	author = {Gu, Hanxue and Dong, Haoyu and Yang, Jichen and Mazurowski, Maciej A.},
	month = may,
	year = {2024},
}

@inproceedings{hatamizadeh_unetr_2022,
	title = {{UNETR}: {Transformers} for {3D} {Medical} {Image} {Segmentation}},
	shorttitle = {{UNETR}},
	doi = {10.1109/WACV51458.2022.00181},
	booktitle = {WACV},
	author = {Hatamizadeh, Ali and Tang, Yucheng and Nath, Vishwesh and Yang, Dong and Myronenko, Andriy and Landman, Bennett and Roth, Holger R. and Xu, Daguang},
	month = jan,
	year = {2022},
	pages = {1748--1758},
}

@article{heuvel_automated_2018,
	title = {Automated measurement of fetal head circumference using {2D} ultrasound images},
	volume = {13},
	issn = {1932-6203},
	url = {https://journals.plos.org/plosone/article?id=10.1371/journal.pone.0200412},
	doi = {10.1371/journal.pone.0200412},
	number = {8},
	urldate = {2024-03-06},
	journal = {Plos One},
	author = {van den Heuvel, Thomas L. A. and de Bruijn, Dagmar and de Korte, Chris L. and van Ginneken, Bram},
	year = {2018},
	pages = {e0200412},
}

@article{huang_segment_2024,
	title = {Segment anything model for medical images?},
	volume = {92},
	issn = {1361-8415},
	url = {https://www.sciencedirect.com/science/article/pii/S1361841523003213},
	doi = {10.1016/j.media.2023.103061},
	urldate = {2024-02-24},
	journal = {Med. Image Anal.},
	author = {Huang, Yuhao and Yang, Xin and Liu, Lian and Zhou, Han and Chang, Ao and Zhou, Xinrui and Chen, Rusi and Yu, Junxuan and Chen, Jiongquan and Chen, Chaoyu and Liu, Sijing and Chi, Haozhe and Hu, Xindi and Yue, Kejuan and Li, Lei and Grau, Vicente and Fan, Deng-Ping and Dong, Fajin and Ni, Dong},
	month = feb,
	year = {2024},
	pages = {103061},
}

@inproceedings{hsu_weakly_2019,
	title = {Weakly {Supervised} {Instance} {Segmentation} using the {Bounding} {Box} {Tightness} {Prior}},
	volume = {32},
	url = {https://papers.nips.cc/paper_files/paper/2019/hash/e6e713296627dff6475085cc6a224464-Abstract.html},
	urldate = {2024-06-30},
	booktitle = {NeurIPS},
	publisher = {Curran Associates, Inc.},
	author = {Hsu, Cheng-Chun and Hsu, Kuang-Jui and Tsai, Chung-Chi and Lin, Yen-Yu and Chuang, Yung-Yu},
	year = {2019},
}

@article{isensee_nnu-net_2021,
	title = {{nnU}-{Net}: a self-configuring method for deep learning-based biomedical image segmentation},
	volume = {18},
	copyright = {2020 The Author(s), under exclusive licence to Springer Nature America, Inc.},
	issn = {1548-7105},
	shorttitle = {{nnU}-{Net}},
	url = {https://www.nature.com/articles/s41592-020-01008-z},
	doi = {10.1038/s41592-020-01008-z},
	number = {2},
	urldate = {2024-03-18},
	journal = {Nature Methods},
	author = {Isensee, Fabian and Jaeger, Paul F. and Kohl, Simon A. A. and Petersen, Jens and Maier-Hein, Klaus H.},
	month = feb,
	year = {2021},
	pages = {203--211},
}

@article{jia_constrained_2017,
	title = {Constrained {Deep} {Weak} {Supervision} for {Histopathology} {Image} {Segmentation}},
	volume = {36},
	number = {11},
	urldate = {2024-06-30},
	journal = {IEEE Trans. Med. Imaging},
	author = {Jia, Zhipeng and Huang, Xingyi and Chang, Eric I-Chao and Xu, Yan},
	year = {2017},
	pages = {2376--2388},
}

@inproceedings{kervadec_bounding_2020,
	title = {Bounding boxes for weakly supervised segmentation: {Global} constraints get close to full supervision},
	shorttitle = {Bounding boxes for weakly supervised segmentation},
	url = {https://proceedings.mlr.press/v121/kervadec20a.html},
	urldate = {2023-11-16},
	booktitle = {MIDL},
	publisher = {PMLR},
	author = {Kervadec, Hoel and Dolz, Jose and Wang, Shanshan and Granger, Eric and Ayed, Ismail Ben},
	month = sep,
	year = {2020},
	pages = {365--381},
}

@inproceedings{kervadec2022constrained,
  title={Constrained deep networks: Lagrangian optimization via log-barrier extensions},
  author={Kervadec, Hoel and Dolz, Jose and Yuan, Jing and Desrosiers, Christian and Granger, Eric and Ayed, Ismail Ben},
  booktitle={EUSIPCO},
  pages={962--966},
  year={2022},
  organization={IEEE}
}

@inproceedings{khoreva_simple_2017,
	address = {Honolulu, HI},
	title = {Simple {Does} {It}: {Weakly} {Supervised} {Instance} and {Semantic} {Segmentation}},
	isbn = {978-1-5386-0457-1},
	shorttitle = {Simple {Does} {It}},
	url = {http://ieeexplore.ieee.org/document/8099664/},
	doi = {10.1109/CVPR.2017.181},
	urldate = {2024-06-30},
	booktitle = {CVPR},
	publisher = {IEEE},
	author = {Khoreva, Anna and Benenson, Rodrigo and Hosang, Jan and Hein, Matthias and Schiele, Bernt},
	month = jul,
	year = {2017},
	pages = {1665--1674},
}

@misc{kim_customizing_2024,
	title = {Customizing {Segmentation} {Foundation} {Model} via {Prompt} {Learning} for {Instance} {Segmentation}},
	url = {arXiv},
	urldate = {2024-05-29},
	publisher = {arXiv},
        number = {arXiv},
	author = {Kim, Hyung-Il and Yun, Kimin and Yun, Jun-Seok and Bae, Yuseok},
	month = mar,
	year = {2024},
}

@inproceedings{kirillov_segment_2023,
	address = {Paris, France},
	title = {Segment {Anything}},
	url = {https://ieeexplore.ieee.org/document/10378323/},
	doi = {10.1109/ICCV51070.2023.00371},
	booktitle = {ICCV},
	publisher = {IEEE},
	author = {Kirillov, Alexander and Mintun, Eric and Ravi, Nikhila and Mao, Hanzi and Rolland, Chloe and Gustafson, Laura and Xiao, Tete and Whitehead, Spencer and Berg, Alexander C. and Lo, Wan-Yen and Dollár, Piotr and Girshick, Ross},
	year = {2023},
	pages = {3992--4003},
}

@inproceedings{kulharia_box2seg_2020,
	title = {{Box2Seg}: {Attention} {Weighted} {Loss} and {Discriminative} {Feature} {Learning} for {Weakly} {Supervised} {Segmentation}},
	isbn = {978-3-030-58583-9},
	shorttitle = {{Box2Seg}},
	doi = {10.1007/978-3-030-58583-9_18},
	booktitle = {ECCV},
	author = {Kulharia, Viveka and Chandra, Siddhartha and Agrawal, Amit and Torr, Philip and Tyagi, Ambrish},
	year = {2020},
	pages = {290--308},
}

@article{leclerc_deep_2019,
	title = {Deep {Learning} for {Segmentation} {Using} an {Open} {Large}-{Scale} {Dataset} in {2D} {Echocardiography}},
	volume = {38},
	issn = {1558-254X},
	url = {https://ieeexplore.ieee.org/document/8649738},
	doi = {10.1109/TMI.2019.2900516},
	number = {9},
	urldate = {2024-02-24},
	journal = {IEEE Trans. Med. Imaging},
	author = {Leclerc, Sarah and Smistad, Erik and Pedrosa, João and Østvik, Andreas and Cervenansky, Frederic and Espinosa, Florian and Espeland, Torvald and Berg, Erik Andreas Rye and Jodoin, Pierre-Marc and Grenier, Thomas and Lartizien, Carole and D’hooge, Jan and Lovstakken, Lasse and Bernard, Olivier},
	month = sep,
	year = {2019},
	pages = {2198--2210},
}

@misc{li_autoprosam_2024,
	title = {{AutoProSAM}: {Automated} {Prompting} {SAM} for {3D} {Multi}-{Organ} {Segmentation}},
	shorttitle = {{AutoProSAM}},
	url = {arXiv},
	publisher = {arXiv},
        number = {arXiv},
	author = {Li, Chengyin and Khanduri, Prashant and Qiang, Yao and Sultan, Rafi Ibn and Chetty, Indrin and Zhu, Dongxiao},
	year = {2024}
}

@inproceedings{lin_scribblesup_2016,
	address = {Las Vegas, NV, USA},
	title = {{ScribbleSup}: {Scribble}-{Supervised} {Convolutional} {Networks} for {Semantic} {Segmentation}},
	isbn = {978-1-4673-8851-1},
	shorttitle = {{ScribbleSup}},
	url = {http://ieeexplore.ieee.org/document/7780713/},
	doi = {10.1109/CVPR.2016.344},
	urldate = {2024-06-30},
	booktitle = {CVPR},
	publisher = {IEEE},
	author = {Lin, Di and Dai, Jifeng and Jia, Jiaya and He, Kaiming and Sun, Jian},
	month = jun,
	year = {2016},
	pages = {3159--3167},
}

@article{litjens_survey_2017,
	title = {A survey on deep learning in medical image analysis},
	volume = {42},
	issn = {1361-8423},
	doi = {10.1016/j.media.2017.07.005},
	journal = {Med. Image Anal.},
	author = {Litjens, Geert and Kooi, Thijs and Bejnordi, Babak Ehteshami and Setio, Arnaud Arindra Adiyoso and Ciompi, Francesco and Ghafoorian, Mohsen and van der Laak, Jeroen A. W. M. and van Ginneken, Bram and Sánchez, Clara I.},
	month = dec,
	year = {2017},
	pmid = {28778026},
	pages = {60--88},
}

@article{ma_segment_2024,
	title = {Segment anything in medical images},
	volume = {15},
	copyright = {2024 The Author(s)},
	issn = {2041-1723},
	url = {https://www.nature.com/articles/s41467-024-44824-z},
	doi = {10.1038/s41467-024-44824-z},
	number = {1},
	urldate = {2024-02-22},
	journal = {Nature Communications},
	author = {Ma, Jun and He, Yuting and Li, Feifei and Han, Lin and You, Chenyu and Wang, Bo},
	month = jan,
	year = {2024},
	pages = {654},
}

@article{mazurowski_segment_2023,
	title = {Segment anything model for medical image analysis: {An} experimental study},
	volume = {89},
	issn = {1361-8415},
	shorttitle = {Segment anything model for medical image analysis},
	url = {https://www.sciencedirect.com/science/article/pii/S1361841523001780},
	doi = {10.1016/j.media.2023.102918},
	urldate = {2023-09-07},
	journal = {Med. Image Anal.},
	author = {Mazurowski, Maciej A. and Dong, Haoyu and Gu, Hanxue and Yang, Jichen and Konz, Nicholas and Zhang, Yixin},
	month = oct,
	year = {2023},
	pages = {102918},
}

@inproceedings{pathak_constrained_2015,
	address = {Santiago, Chile},
	title = {Constrained {Convolutional} {Neural} {Networks} for {Weakly} {Supervised} {Segmentation}},
	isbn = {978-1-4673-8391-2},
	url = {http://ieeexplore.ieee.org/document/7410566/},
	doi = {10.1109/ICCV.2015.209},
	urldate = {2024-06-30},
	booktitle = {ICCV},
	publisher = {IEEE},
	author = {Pathak, Deepak and Krahenbuhl, Philipp and Darrell, Trevor},
	month = dec,
	year = {2015},
	pages = {1796--1804},
}

@article{rajchl_deepcut_2017,
	title = {{DeepCut}: {Object} {Segmentation} {From} {Bounding} {Box} {Annotations} {Using} {Convolutional} {Neural} {Networks}},
	volume = {36},
	issn = {1558-254X},
	shorttitle = {{DeepCut}},
	url = {https://ieeexplore.ieee.org/document/7739993},
	doi = {10.1109/TMI.2016.2621185},
	number = {2},
	urldate = {2024-03-20},
	journal = {IEEE Trans. Med. Imaging},
	author = {Rajchl, Martin and Lee, Matthew C. H. and Oktay, Ozan and Kamnitsas, Konstantinos and Passerat-Palmbach, Jonathan and Bai, Wenjia and Damodaram, Mellisa and Rutherford, Mary A. and Hajnal, Joseph V. and Kainz, Bernhard and Rueckert, Daniel},
	month = feb,
	year = {2017},
	pages = {674--683},
}

@inproceedings{ronneberger_u-net_2015,
	address = {Cham},
	series = {Lecture {Notes} in {Computer} {Science}},
	title = {U-{Net}: {Convolutional} {Networks} for {Biomedical} {Image} {Segmentation}},
	isbn = {978-3-319-24574-4},
	shorttitle = {U-{Net}},
	doi = {10.1007/978-3-319-24574-4_28},
	booktitle = {MICCAI},
	publisher = {Springer International Publishing},
	author = {Ronneberger, Olaf and Fischer, Philipp and Brox, Thomas},
	editor = {Navab, Nassir and Hornegger, Joachim and Wells, William M. and Frangi, Alejandro F.},
	year = {2015},
	pages = {234--241},
}

@article{rother_grabcut_2004,
	title = {"{GrabCut}": interactive foreground extraction using iterated graph cuts},
	volume = {23},
	issn = {0730-0301},
	shorttitle = {"{GrabCut}"},
	url = {https://doi.org/10.1145/1015706.1015720},
	doi = {10.1145/1015706.1015720},
	number = {3},
	urldate = {2024-06-25},
	journal = {ACM Trans. Graph.},
	author = {Rother, Carsten and Kolmogorov, Vladimir and Blake, Andrew},
	year = {2004},
	pages = {309--314},
}

@inproceedings{shaharabany_autosam_2023,
	title = {{AutoSAM}: {Adapting} {SAM} to {Medical} {Images} by {Overloading} the {Prompt} {Encoder}},
	booktitle = {BMVC},
	author = {Shaharabany, Tal},
	year = {2023},
}

@inproceedings{song_box-driven_2019,
	title = {Box-{Driven} {Class}-{Wise} {Region} {Masking} and {Filling} {Rate} {Guided} {Loss} for {Weakly} {Supervised} {Semantic} {Segmentation}},
	doi = {10.1109/CVPR.2019.00325},
	booktitle = {CVPR},
	author = {Song, Chunfeng and Huang, Yan and Ouyang, Wanli and Wang, Liang},
	year = {2019},
	pages = {3131--3140},
}

@inproceedings{vaswani_attention_2017,
	title = {Attention is {All} you {Need}},
	volume = {30},
	url = {https://proceedings.neurips.cc/paper_files/paper/2017/hash/3f5ee243547dee91fbd053c1c4a845aa-Abstract.html},
	urldate = {2023-05-09},
	booktitle = {NeurIPS},
	publisher = {Curran Associates, Inc.},
	author = {Vaswani, Ashish and Shazeer, Noam and Parmar, Niki and Uszkoreit, Jakob and Jones, Llion and Gomez, Aidan N and Kaiser, Łukasz and Polosukhin, Illia},
	year = {2017},
}

@inproceedings{wang_bounding_2021,
	address = {Cham},
	series = {Lecture {Notes} in {Computer} {Science}},
	title = {Bounding {Box} {Tightness} {Prior} for {Weakly} {Supervised} {Image} {Segmentation}},
	isbn = {978-3-030-87196-3},
	doi = {10.1007/978-3-030-87196-3_49},
	booktitle = {MICCAI},
	publisher = {Springer International Publishing},
	author = {Wang, Juan and Xia, Bin},
	editor = {de Bruijne, Marleen and Cattin, Philippe C. and Cotin, Stéphane and Padoy, Nicolas and Speidel, Stefanie and Zheng, Yefeng and Essert, Caroline},
	year = {2021},
	pages = {526--536},
}

@article{wang_review_2023,
	title = {Review of large vision models and visual prompt engineering},
	volume = {1},
	issn = {2950-1628},
	url = {https://www.sciencedirect.com/science/article/pii/S2950162823000474},
	doi = {10.1016/j.metrad.2023.100047},
	number = {3},
	urldate = {2024-05-31},
	journal = {Meta-Radiology},
	author = {Wang, Jiaqi and Liu, Zhengliang and Zhao, Lin and Wu, Zihao and Ma, Chong and Yu, Sigang and Dai, Haixing and Yang, Qiushi and Liu, Yiheng and Zhang, Songyao and Shi, Enze and Pan, Yi and Zhang, Tuo and Zhu, Dajiang and Li, Xiang and Jiang, Xi and Ge, Bao and Yuan, Yixuan and Shen, Dinggang and Liu, Tianming and Zhang, Shu},
	month = nov,
	year = {2023},
	pages = {100047},
}

@article{wu_MedicalSAMAdapter_2025,
  title = {Medical {{SAM}} Adapter: {{Adapting}} Segment Anything Model for Medical Image Segmentation},
  shorttitle = {Medical {{SAM}} Adapter},
  author = {Wu, Junde and Wang, Ziyue and Hong, Mingxuan and Ji, Wei and Fu, Huazhu and Xu, Yanwu and Xu, Min and Jin, Yueming},
  year = {2025},
  month = may,
  journal = {Medical Image Analysis},
  volume = {102},
  pages = {103547},
  issn = {1361-8415},
  doi = {10.1016/j.media.2025.103547},
  urldate = {2025-04-02}
}

@inproceedings{wu_self-prompting_2023,
	address = {Cham},
	series = {Lecture {Notes} in {Computer} {Science}},
	title = {Self-prompting {Large} {Vision} {Models} for {Few}-{Shot} {Medical} {Image} {Segmentation}},
	isbn = {978-3-031-45857-6},
	doi = {10.1007/978-3-031-45857-6_16},
	booktitle = {DART},
	publisher = {Springer Nature Switzerland},
	author = {Wu, Qi and Zhang, Yuyao and Elbatel, Marawan},
	editor = {Koch, Lisa and Cardoso, M. Jorge and Ferrante, Enzo and Kamnitsas, Konstantinos and Islam, Mobarakol and Jiang, Meirui and Rieke, Nicola and Tsaftaris, Sotirios A. and Yang, Dong},
	year = {2023},
	pages = {156--167},
}

@inproceedings{yue_surgicalsam_2024,
	title = {{SurgicalSAM}: {Efficient} {Class} {Promptable} {Surgical} {Instrument} {Segmentation}},
	volume = {38},
	copyright = {Copyright (c) 2024 Association for the Advancement of Artificial Intelligence},
	shorttitle = {{SurgicalSAM}},
	url = {https://ojs.aaai.org/index.php/AAAI/article/view/28514},
	doi = {10.1609/aaai.v38i7.28514},
	booktitle = {AAAI},
	author = {Yue, Wenxi and Zhang, Jing and Hu, Kun and Xia, Yong and Luo, Jiebo and Wang, Zhiyong},
	year = {2024},
	note = {Number: 7},
	pages = {6890--6898},
}

@inproceedings{zhang_personalize_2024,
	title = {Personalize {Segment} {Anything} {Model} with {One} {Shot}},
	url = {http://arxiv.org/abs/2305.03048},
	urldate = {2023-10-20},
	booktitle = {ICLR},
	publisher = {arXiv},
        number = {arXiv},
	author = {Zhang, Renrui and Jiang, Zhengkai and Guo, Ziyu and Yan, Shilin and Pan, Junting and Ma, Xianzheng and Dong, Hao and Gao, Peng and Li, Hongsheng},
	year = {2024},
}

@inproceedings{zhang_segment_2023,
	address = {Istanbul, Turkiye},
	title = {Segment {Anything} {Model} ({SAM}) for {Medical} {Image} {Segmentation}: {A} {Preliminary} {Review}},
	booktitle = {{BIBM}},
	publisher = {IEEE},
	author = {Zhang, Leying and Deng, Xiaokang and Lu, Yu},
	year = {2023},
	pages = {4187--4194},
}

@inproceedings{zou_segment_2023,
	title = {Segment {Everything} {Everywhere} {All} at {Once}},
	volume = {36},
	url = {https://proceedings.neurips.cc/paper_files/paper/2023/hash/3ef61f7e4afacf9a2c5b71c726172b86-Abstract-Conference.html},
	urldate = {2024-05-29},
	booktitle = {NeurIPS},
	author = {Zou, Xueyan and Yang, Jianwei and Zhang, Hao and Li, Feng and Li, Linjie and Wang, Jianfeng and Wang, Lijuan and Gao, Jianfeng and Lee, Yong Jae},
	month = dec,
	year = {2023},
	pages = {19769--19782},
}

\end{document}